\documentclass{vgtc}                          
\usepackage{mathptmx}                  

\usepackage{amsmath}
\usepackage{amssymb}
\usepackage{comment}

\usepackage{enumitem}

\usepackage[ruled,linesnumbered]{algorithm2e}

\usepackage{xcolor}
\definecolor{mycolor}{RGB}{128, 0, 64}
\usepackage{helvet}
\newenvironment{helveticasection}
  {\sffamily}  %
  {}           %

\newcommand{\up}[0]{{\textcolor{teal}{{$\blacktriangle$}}}}
\newcommand{\down}[0]{{\textcolor{red}{{$\blacktriangledown$}}}}

\newcommand{\revise}[1]{{\textcolor{black}{#1}}}

\ifpdf
  \pdfoutput=1\relax                   
  \usepackage{graphicx}                
  \DeclareGraphicsExtensions{.pdf,.png,.jpg,.jpeg} 
\else
  \usepackage{graphicx}                
  \DeclareGraphicsExtensions{.eps}     
\fi%

\graphicspath{{figures/}{pictures/}{images/}{./}} 

\usepackage{microtype}                 
\PassOptionsToPackage{warn}{textcomp}  
\usepackage{textcomp}                  
\usepackage{mathptmx}                  
\usepackage{times}                     
\usepackage{cite}                      
\usepackage{tabu}                      
\usepackage{booktabs}                  

\onlineid{0}

\vgtccategory{Research}

\vgtcinsertpkg

\title{iGAiVA: Integrated Generative AI and Visual Analytics in a Machine Learning Workflow for Text Classification}

\author{Yuanzhe Jin\thanks{e-mail: yuanzhe.jin@eng.ox.ac.uk}\\ %
        \scriptsize University of Oxford %
\and Adri\'{a}n Carrasco-Revilla\thanks{e-mail: adrian.carrasco@inetum.com}\\ %
     \scriptsize Inetum Spain %
\and  Min Chen\thanks{e-mail: min.chen@eng.ox.ac.uk}\\ %
    \scriptsize University of Oxford}

\abstract{%
In developing machine learning (ML) models for text classification, one common challenge is that the collected data is often not ideally distributed, especially when new classes are introduced in response to changes of data and tasks. In this paper, we present a solution for using visual analytics (VA) to guide the generation of synthetic data using large language models. As VA enables model developers to identify data-related deficiency, data synthesis can be targeted to address such deficiency. We discuss different types of data deficiency, describe different VA techniques for supporting their identification, and demonstrate the effectiveness of targeted data synthesis in improving model accuracy. In addition, we present a software tool, iGAiVA, which maps four groups of ML tasks into four VA views, integrating generative AI and VA into an ML workflow for developing and improving text classification models.
} 

\keywords{Machine learning, text classification, visual analytics, generative AI, large language models, LLM, synthetic data, tag cloud treemap, radial basis function, four-view design, VIS4ML}

\begin{document}



\firstsection{Introduction\label{sec:Intro}}

\maketitle


Text analysis and visualization have been studied extensively in the field of visualization and visual analytics (VIS for short). VIS4ML~\cite{Sacha:2018:TVCG}, i.e., using VIS to support machine learning (ML) workflows, has also been a research topic attracting much attention in the past decade. 
While generative AI has become a disruptive technology with global impact, developing ML models for analyzing texts in many contexts is still a non-trivial undertaking.
The performance of ML models is commonly impaired by the limitation of data collection, because (i) the information space (i.e., all possible variations) of texts is huge and adequate sampling would demand both time and resources, (ii) often the amount of text data available to an ML workflow in an application is limited and the distribution of the sampled data is stewed, or (iii) organizational and societal changes sometimes undermine existing ML models while their retraining is hindered by limited availability of new data.

This work is concerned with using VIS to support an ML4ML workflow in the domain of text analysis. In particular, we use large language models (LLMs) to address the limited availability of real-world data in training ML models for analyzing text data automatically and dynamically in computerized ticketing systems. Because the search space for selecting text examples for LLMs is huge, we use VIS to enable ML developers to target the uses of LLMs at the gaps and shortcomings in the training data. In other words, through visual analytics, ML developers can use their knowledge and reasoning effectively to reduce the search space significantly. The contributions of this work include:
\begin{itemize}
    \item Proposing a novel VIS4ML approach in which VIS techniques are used to guide the processes of data synthesis using LLMs.
    \vspace{-1.5mm}
    \item Developing a software tool, \textbf{iGAiVA}, which integrates generative AI and visual analytics into a unified ML workflow.%
    \vspace{-1.5mm}
    \item Demonstrating the effectiveness of targeted data synthesis in improving ML model accuracy through visual analytics.
\end{itemize}
\section{Related Work}
\label{sec:RelatedWork}
In this section, we review the previous research in VIS for text analysis, ML for text analysis, and VIS for supporting ML workflows. 

\subsection{VIS in Text Analysis}
Alharbi et al.~\cite{Alharbi:2018:CGVC, Alharbi:2019:Com} conducted a survey on text visualization, categorizing existing surveys into five groups and offering recommendations to researchers.
Liu et al.~\cite{Liu:2018:TVCG} conducted a task-driven survey, categorizing text visualization techniques according to user tasks, such as topic understanding, sentiment analysis, and discourse analysis.
Fischer et al.~\cite{Fischer:2022:Book} focused their survey on the analysis of communication data, highlighting current practices, challenges, and emerging research directions in the area.

There are several commonly-seen plots for text visualization. TagCloud~\cite{Kaser:2007:arXiv} and Word Cloud~\cite{Cui:2010:PacificVis} summarizes the statistics of word usage using different character sizes. Texts are often combined with trees and graphs to depict the relationships among words and phrases (e.g., \cite{Le:2014:ICML, Gansner:2012:ISGD}).
There are also several visual designs for social media data, such as maps for topics~\cite{Park:2003:xml} and flows for changing texts~\cite{Cui:2011:TVCG, Liu:2012:TIS, Wu:2017:TVCG}.

Since texts are themselves ``raw visual representations'', almost all text visualization plots feature processed data resulting from statistics and algorithms. Much of VIS research in text analysis focused on techniques and tools where statistics, algorithms, visualization, and interaction were integrated. In recent years, ML models (i.e., ML-learned algorithms) play more noticeable roles in text analysis. 
Luo et al.~\cite{Luo:2010:VIS} developed a tool (EventRiver) where event-based text analysis and visualization were integrated to reveal the events that inspired some texts and stories. 
Park et al.~\cite{Park:2012:SIGCHI} introduced a temporal model for representing narrative clinical text in chronological order.
Brehmer et al.~\cite{Brehmer:2014:VIS} designed a system (Overview) for investigative journalists to perform document clustering, visualization, and labeling.
Liu et al.~\cite{Liu:2015:TVCG} combined a Bayesian network model with flow-based text visualization to analyze real-time text streams.
Duo and Liu~\cite{Dou:2016:TVCG} combined topic modeling with interactive visualization to analyze temporal text data.
Wu et al.~\cite{Wu:2017:TVCG} combined a self-organizing map (SOM) with glyph-based timeline visualization to explore social media data.
Abdul-Rahman et al.\cite{AbdulRahman:2017:CGF} developed a VIS tool (ViTA) for text similarity analysis, which allowed users to define text comparison functions using block diagrams. 

El-Assady et al. used an integrated approach to analyze conversational text~\cite{El:2018:CGF, El:2019:ACL}, explore topic space views~\cite{El:2016:CGF}, and examine named entity relationships~\cite{El:2017:CGF}.
Schorr et al.~\cite{Schorr:2020:vis} developed a web-based tool for detecting repetitive patterns in a text corpus.
Knittel et al.~\cite{Knittel:2021:TVCG} proposed to use a parallel clustering approach to handle high-volume text data dynamically.
Sevastjanova et al.~\cite{Sevastjanova:2021:VIS} developed a system to enable personalized insight reports in visual discourse analysis.
Fischer et al.~\cite{Fischer:2021:CGF} described a technique for communication analysis through interactive modeling of dynamic data.

\subsection{ML in Text Analysis and Generation}
ML models have become an indispensable part of various workflows for processing texts. Many text analysis tasks fall into the category of text classification \cite{Aravindan:2023:SMART}, which includes topic modeling (e.g., \cite{Qiang:2020:short}) and sentiment analysis (e.g., \cite{Raghunathan:2023:challenges, Li:2024:explanation}).
Other text processing tasks include relationship extraction (e.g., \cite{Wang:2024:Frontiers}), machine translation (e.g., \cite{La:2023:ISIDA}), text summarization (e.g.,\cite{El:2021:ESA}), text generation (e.g., \cite{Crothers:2023:machine}), spelling and grammar checking (e.g.,\cite{Etoori:2018:ACL}), and so on.

The introduction of pre-trained models like BERT~\cite{Devlin:2019:bert} enhanced text analysis capabilities.
The emergence of LLMs, e.g., GPT-3~\cite{Dale:2021:gpt}, has introduced important new resources to text analysis and visualization. Schetinger et al. discussed some challenges associated with these models~\cite{Schetinger:2023:CGF}, while El-Assady et al.~\cite{El:2017:TVCG} and Sevastjanova et al.~\cite{Sevastjanova:2022:TVCG, Sevastjanova:2023:VDS} explored potential opportunities.
In our work, we use LLMs to generate synthetic training data to address the challenges in collecting labelled data in some real-world applications.

\subsection{VIS for Supporting ML Workflows}
\noindent\textbf{VIS4ML in General.}
Tam et al.~\cite{Tam:2016:TVCG} analyzed humans' role in ML workflows and estimated their contributions quantitatively using information theory. 
Sacha et al.~\cite{Sacha:2018:TVCG} outlined a VIS4ML ontology, pointing out that many processes in ML workflows can benefit from VIS.
Wang et al.~\cite{Wang:2020:TVCG} introduced HypoML, a VIS-enabled method for testing different hypotheses about whether ML models have learned particular features.
\revise{Zhang et al. \cite{Zhang:2022:TVCG} and Chen et al. \cite{Chen:2020:TVCG} used VIS for quality analysis of   
the data used in training and testing.}%
 
VIS has been utilized to support a variety of ML workflows featuring techniques such as
CNN \cite{Wang:2020:TVCG,Borowski:2020:ICLR},
RNN \cite{Shen:2020:PVIS,Wang:2021:TVCG},
decision trees \cite{Liu:2017:TVCG,Sarailidis:2023:CG},
random forest \cite{Chatzimparmpas:2023:InfoVIS,Lau:2014:journal},
reinforcement learning \cite{Shin:2024:BE,Ze:2023:RAL, Metz:2023:CGF},
ensemble learning \cite{Ye:2022:TVCG, Ferano:2023:journal},
and so on.
These VIS4ML workflows have been used to develop ML models for different applications, e.g.,
music analysis \cite{Ye:2022:TVCG, Ferano:2023:journal},
system control \cite{Ze:2023:RAL, Shin:2024:BE},
weather prediction \cite{Shen:2020:PVIS, Lau:2014:journal},
image classification \cite{Wang:2020:TVCG,Borowski:2020:ICLR}, and so on.

\revise{\emph{Data argumentation} (e.g., \cite{Bayer:2022:ACM}) and \emph{data imputation} (e.g., \cite{Richter:2024:Access}) both refer to techniques for generating synthetic data based on statistical characteristics of known data. The former usually focuses more on synthetic data used for training ML models and the latter focuses on filling in the blanks of missing data. VIS has been used in supporting both argumentation (e.g., Lee et al. \cite{Lee:2023:IICAIET}) and data imputation (Sarma et al.~\cite{Sarma:2022:TVCG}, Song et al.~\cite{Song:2021:VIS}, B{\"o}gl et al. \cite{Bogl:2015:VAST}). To generate synthetic text data, LLMs provide a more powerful tool than traditional techniques, because of more statistical attributes and more reliable measures that are computed from very large datasets, and because of other models for text synthesis, spelling and grammar checking and correction, style mutation, etc.      
}

In this work, we use VIS to support example selection for \revise{LLM-based data argumentation}, which in turn supports the development of text classifiers. In other words, it is VIS4ML4ML.

\vspace{2mm}
\noindent\textbf{VIS4ML in Text Analysis.}
Workflows for developing ML models for text analysis often include different processes where human knowledge is used to assist in the development.
For example,
Heimerl et al.~\cite{Heimerl:2012:VIS} proposed VIS methods for training and evaluating text classifiers.
Sperrle et al.~\cite{Sperrle:2019:VAST} developed a tool called VIANA for assisting annotation of argumentation with the aid of interactive visualization.
El-Assady et al. proposed VIS techniques for topic model optimization, through visualizing parameters~\cite{El:2017:TVCG}, progressive learning parameters~\cite{El-Men:2017:TVCG}, and manipulating speculative execution~\cite{El-Men:2018:TVCG}.
El-Assady et al.~\cite{El-Men:2019:TVCG} developed a VIS4ML tool, with which ML developers can explore semantic concept spaces to guide topic model refinement.
Sperrle et al.~\cite{Sperrle:2021:CGF} developed a VIS tool for refining
topic models with the aid of adaptive guidance generated using mixed-initiative.
Wang et al.~\cite{Wang:2021:VIS} developed M2Lens, a VIS tool for better understanding and diagnosing multi-modality models for sentiment analysis.

VIS can also provide means to enhance the explainability and interpretability of ML models.
Liu et al.~\cite{Liu:2017:Journal} highlighted how visual analytics enhances the interpretation of ML models in text analysis.
Spinner et al.~\cite{Spinner:2019:TVCG} developed explAIner, a VIS framework for interactive and explainable ML.
Sevastjanova et al.~\cite{Sevastjanova:2021:ACL} used VIS to aid the explanation of language model contextualization.
Sevastjanova et al.~\cite{Sevastjanova:2021:TIIS} proposed a gamified approach for explaining language phenomena through interactive labeling.
Li et al.~\cite{Li:2022:TVCG} developed a VIS tool (DeepNLPVis) to facilitate a better understanding of NLP models for text classification.

\begin{figure}[htbp]
  \centering
  \includegraphics[width=\linewidth]{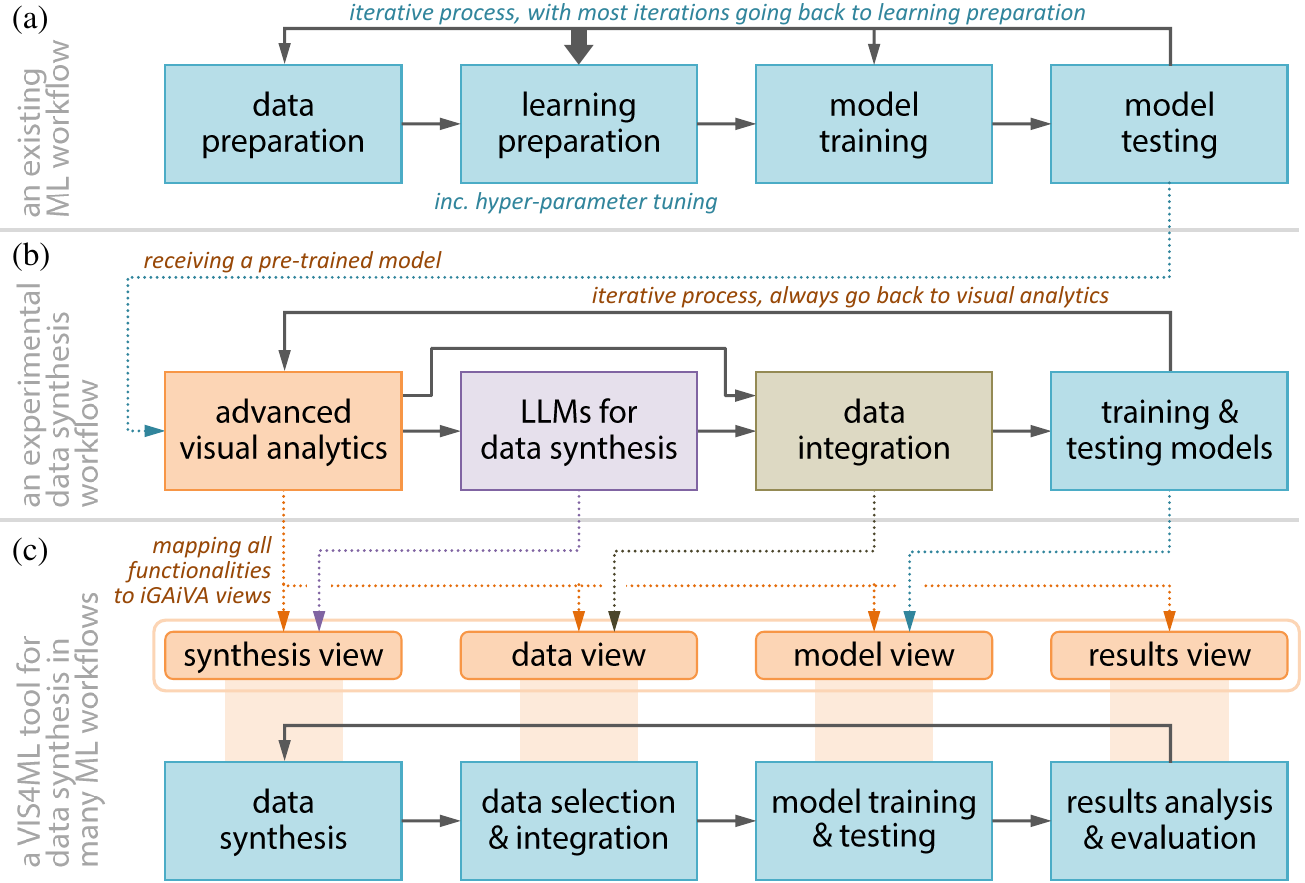}
  \caption{The evolution from a conventional ML workflow to an experimental workflow involving the uses of VIS techniques and LLMs for data synthesis, and then to an iterative workflow supported by a VIS4ML tool where VIS and LLMs techniques are integrated.}
\label{fig:Workflows}
\vspace{-4mm}
\end{figure}

\section{Background, Motivation, and Process Overview}
\label{sec:Overview}
Inetum (https://www.inetum.com/en/Inetum) is an agile IT services company that provides digital services and solutions, and a global group that helps companies and institutions to get the most out of digital flow. The FabLab team in Inetum (Spain) has been delivering ML models for classifying text messages received from computerized ticketing systems, facilitating automated task distribution and rapid ticket responses. For different ticketing systems (e.g., finance, IT, estate, etc.), the messages can be different (e.g., in terms of contexts, lengths, keywords, acronyms, etc.). Different hosts of ticketing systems may have different organizational structures for handling these tickets, leading to different classification schemes. The FabLab team has found that developing models for individual ticketing systems can deliver better solutions to the customers.

In conventional ML workflows, as shown in Fig. \ref{fig:Workflows}(a), ML developers perform numerous iterations from data preparation to learning preparation, training, and testing \cite{Hohman:2020:CHI}. As demonstrated by many existing VIS4ML techniques (Section \ref{sec:RelatedWork}), VIS techniques can help reduce the number of iterations and speed up the human-centric processes in these iterations. When the first author started his 2-month placement in Inetum, we quickly identified the first requirement based on our observation of the existing ML workflows and our knowledge and experience of VIS4ML, i.e.,
\textbf{R1: Using more VIS techniques to help identify possible causes of errors}.

During the first month of the placement, we introduced several VIS techniques for identifying possible causes of errors (\textbf{R1}). In particular, we noticed two types of causes: for some classes, (a) there may not be enough training data, and (b) the training data may be of a skewed distribution.
Similar to many industrial ML problems, it is not always feasible to collect a sufficient amount of real-world data, especially when some organizational changes trigger semantic modifications to the classification scheme (e.g., adding a new class, merging two classes, or changing some class definitions). In order not for the ML workflow to wait for the relatively slow process to collect a large amount of new data, we identified the second requirement, i.e.,
\textbf{R2: Using large language models (LLMs) to generate synthetic data for training and using VIS techniques to guide the data synthesis process}.

During the second month of the placement, we carried out experiments to address \textbf{R2}. We received positive results indicating that (i) synthetic data could improve ML models, and (ii) VIS techniques could help target data synthesis to the possible causes of errors and evaluate the effectiveness of each data synthesis action. These results encouraged us to consider a substantial industrial requirement. Because the FabLab team in Inetum (Spain) has to develop different ML models for different ticketing systems, ideally there is a software tool that can support multiple ML workflows. This led to the third requirement, i.e.,
\textbf{R3: Designing and developing a VIS4ML tool for supporting ML workflows involving data synthesis}.

The three main requirements for this work were identified in an agile manner, which is consistent with the nested model approach \cite{Henriques:2014:nested}. During the 2-month placement, we formulated a VIS4ML workflow as shown in Fig. \ref{fig:Workflows}(b). We will detail these VIS techniques in Section \ref{sec:VIS}, and VIS-guided data synthesis in Section \ref{sec:LLM}. Following the placement, we designed and prototyped a VIS4ML tool, called iGAiVA (\emph{integrated Generative AI and Visual Analytics}), for enabling VIS-guided data synthesis in iterative ML workflows as illustrated in Fig. \ref{fig:Workflows}(c).
We will detail the design of iGAiVA and its prototype in Section \ref{sec:iGAiVA}. There were two further placements, where the system was evaluated and improved in an agile manner. We will report this process in Section \ref{sec:Conclusions}.
\begin{table}[t]
\centering
\caption{As an example, a real-world dataset was collected from a computerized ticketing system. 39,100 labeled messages are divided into 15 classes, and the column ``Size'' indicates the number of messages. 80\% of the data was used for training, and 20\% for testing. The column ``Recall'' shows the results of testing the best CatBoost model developed with the traditional ML workflow in Fig. \ref{fig:Workflows}(a).}
\label{tab:Dataset}
\begin{tabular}{rlrr}
\textbf{T[i]} & \textbf{Topic Name} & \textbf{Size} & \textbf{Recall}\\
\hline
T1 & IT support and assistance & 8529 &  0.976\\
T2 & Account activation and access issues & 11350 & 0.943\\
T3 & Password and device security & 4719 & 0.892\\
T4 & Printer issues and troubleshooting & 1387 & 0.769\\
T5 & HP Dock connectivity issues & 2755 & 0.732\\
T6 & Employee documentation and errors & 1888 & 0.528\\
T7 & Access and login issues & 1963 & 0.714\\
T8 & Opening and managing files/devices & 1028 & 0.508\\
T9 & Mobile email and VPN setup & 1466 & 0.626\\
T10 & IT support and communication & 1699 & 0.427\\
T11 & Error handling in RPG programming & 471 & 0.926\\
T12 & Email security and attachments & 358 & 0.375\\
T13 & Humanitarian aid for Ukraine & 180 & 0.178\\
T14 & Internet connectivity issues in offices & 764 & 0.526\\
T15 & Improving integration with Infojobs & 543 & 0.376\\
\hline
   & \textbf{All data objects} & 39100 & 0.821\\
\hline
\end{tabular}
\vspace{-4mm}
\end{table}

\begin{figure}[t]
  \centering
  \includegraphics[width=84mm]{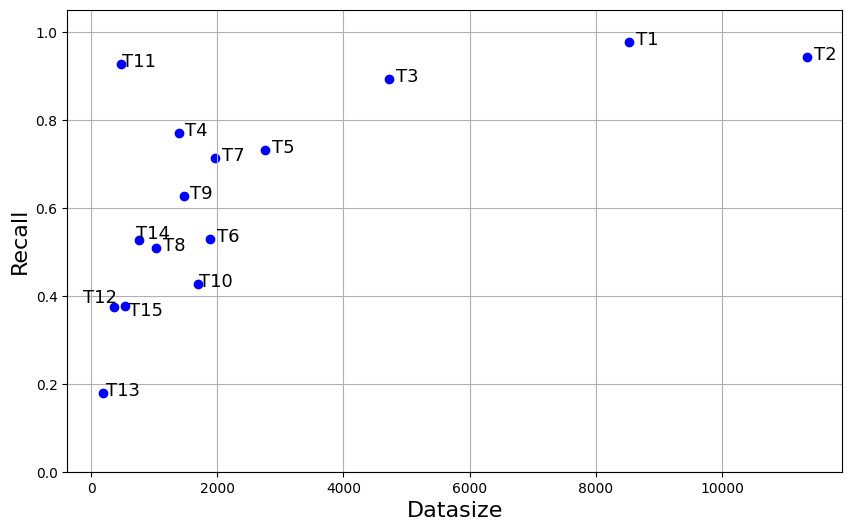}\\[-2mm]
  \caption{Source data size vs. Recall}
\label{fig:InitialScatterPlot}
\vspace{-4mm}
\end{figure}

\section{Visual Analytics for Problem Identification}
\label{sec:VIS}
While the technical work reported in this paper is applicable to most computerized ticketing systems, as an example, we focus on one specific system in this paper. This ticketing system deals with IT-related problems. As illustrated in Table \ref{tab:Dataset}, according to the organizational structure, incoming messages are to be classified into 15 task categories and assigned to the relevant IT specialists.

Before advanced VIS techniques were deployed, the team of ML developers worked with the traditional ML workflow as shown in Fig. \ref{fig:Workflows}(a) for many months in conjunction with a dataset consisting of 39,100 labeled data objects (messages). The team used two ML methods, namely gradient-boosted decision trees (using CatBoost) and convolutional neural network (CNN, using TensorFlow), to train the models in conjunction with 80\% randomly data objects, and tested with the remaining 20\% data objects. A large number of iterations were focused on hyper-parameters of the training process. The recall column in Table \ref{tab:Dataset} shows the results of testing the best CatBoost model obtained using the traditional ML workflow. Note that each class $T_i$ has $p_i$ labelled positive messages and $n_i = \text{Total} - p_i$ labelled negative messages. Hence, $n_i \gg p_i$, especially for some small classes. In this work, we focus on recall that is not affected by $n_i \gg p_i$, though we also used other performance metrics (e.g., accuracy, F1-score, and so on).     

As shown in Fig. \ref{fig:InitialScatterPlot}, one of our initial visualizations revealed some noticeable correlation between the number of data objects in a class and the recall of the class. We suspected that there were some data sampling issues, and therefore focused our attention on data distribution. While visualizing summary statistics (e.g., Fig. \ref{fig:InitialScatterPlot}) can suggest aspects to be investigated, it is necessary to conduct a more detailed investigation into the potential causes of the high error rates in some classes.

As there are $\geq 180$ data objects in each class, visualization alone could not prioritize what to visualize, and interactively reading every message is excessively costly in terms of time and cognitive effort. Meanwhile, texts are rich in semantics and their meanings are context-sensitive, commonly-available algorithms cannot encode an adequate amount of human knowledge for dealing with messages in a specific ticketing system. This is exactly the scenario where statistics, algorithms, visualization, and interaction cannot work alone effectively but have to be co-deployed in a workflow \cite{Chen:2011:Computer}, which is the principal motivation of visual analytics (VA).

For the requirement \textbf{R1} discussed in Section \ref{sec:Overview}, we use four types of VA techniques for organizing data objects using analytical algorithms and observing different patterns visually. From the visual patterns, we hypothesize potential causes of errors and use statistics to evaluate such hypotheses. In the following four subsections, we describe how each type of VA enables pattern discovery and hypothesis generation. 

\begin{figure}[t]
  \centering
  \includegraphics[width=\linewidth]{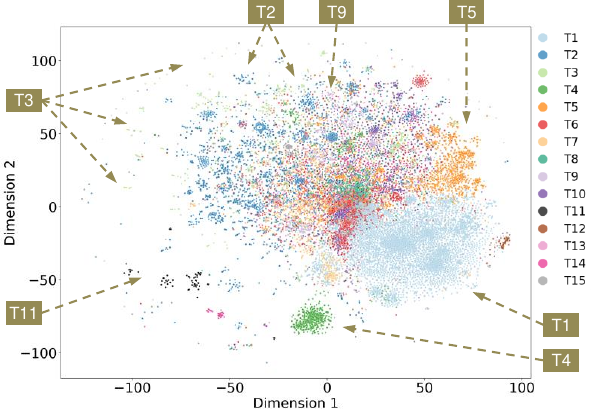}
  \caption{Visual patterns in a t-SNE scatter plot can offer some hints about the data-related causes of accurate or erroneous classification.}
\label{fig:t-SNE}
\vspace{-4mm}
\end{figure}
\begin{figure*}[t]
    \centering
    \begin{tabular}{@{}c@{\hspace{2mm}}c@{}c@{}c@{}c@{}}
    \includegraphics[height=45mm]{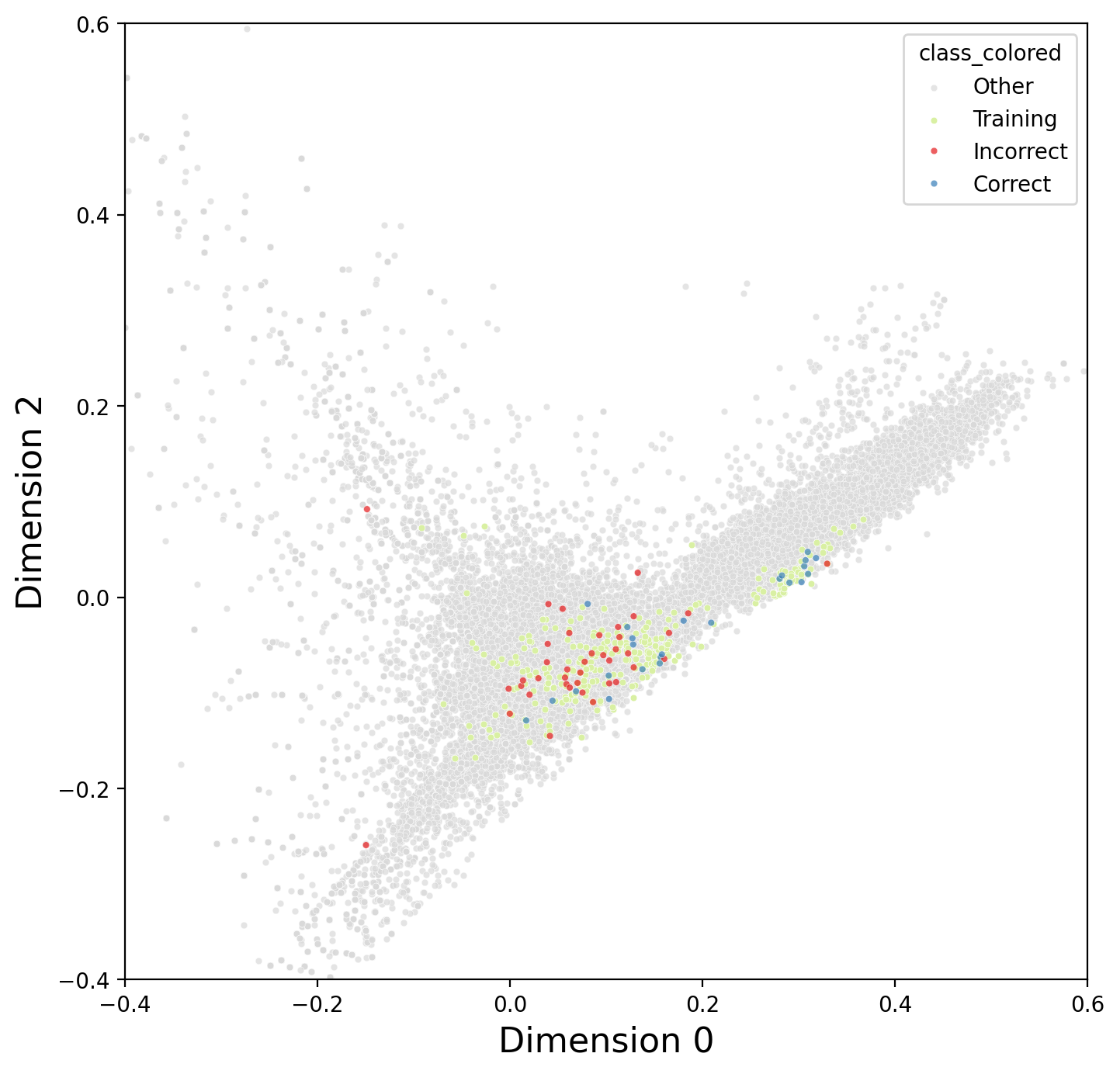} &
    \includegraphics[height=45mm]{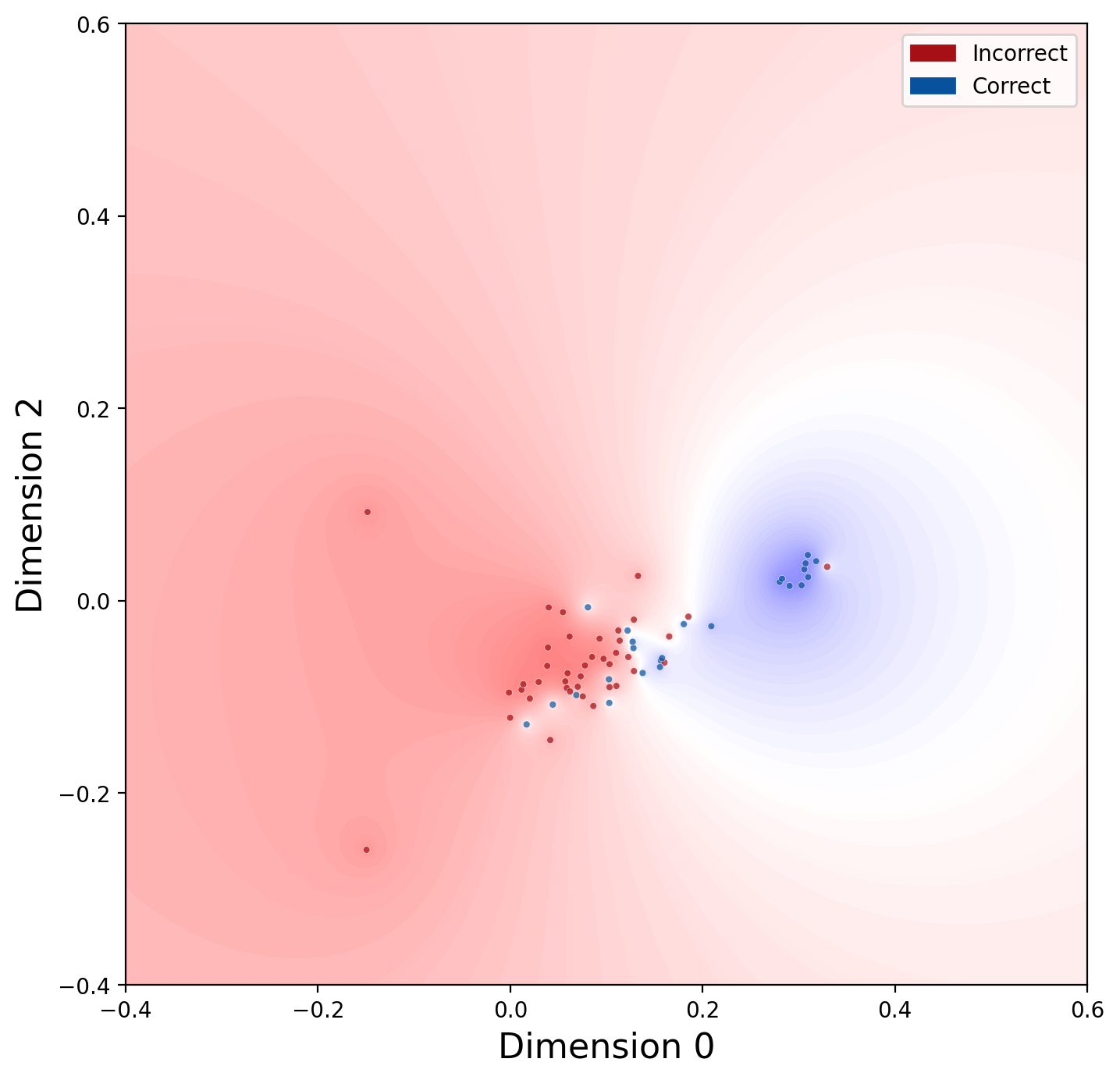} &
    \includegraphics[height=45mm]{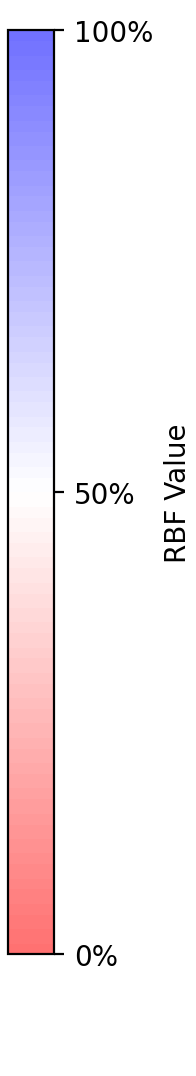} &
    \includegraphics[height=18mm]{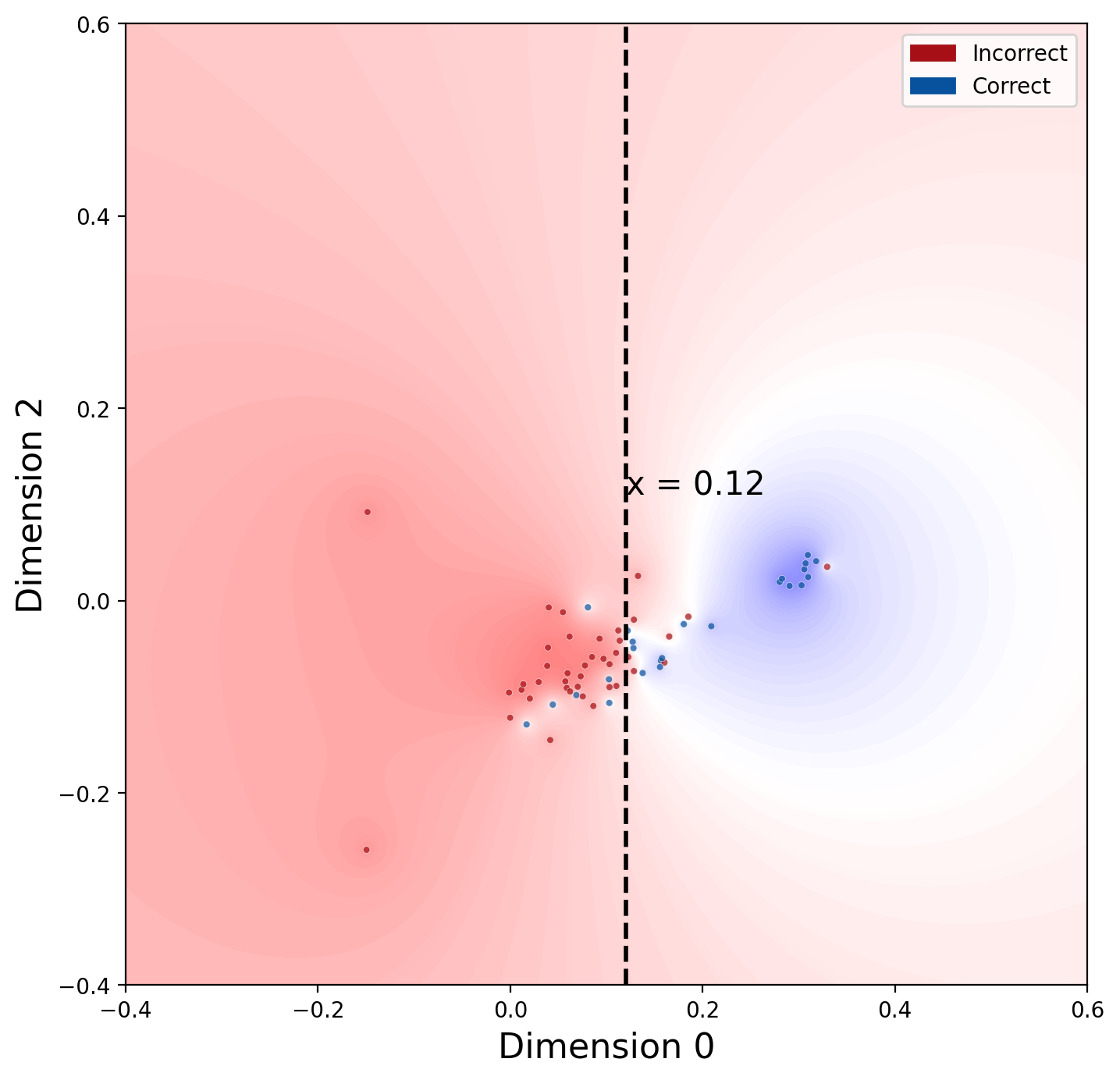} &
    \includegraphics[height=45mm]{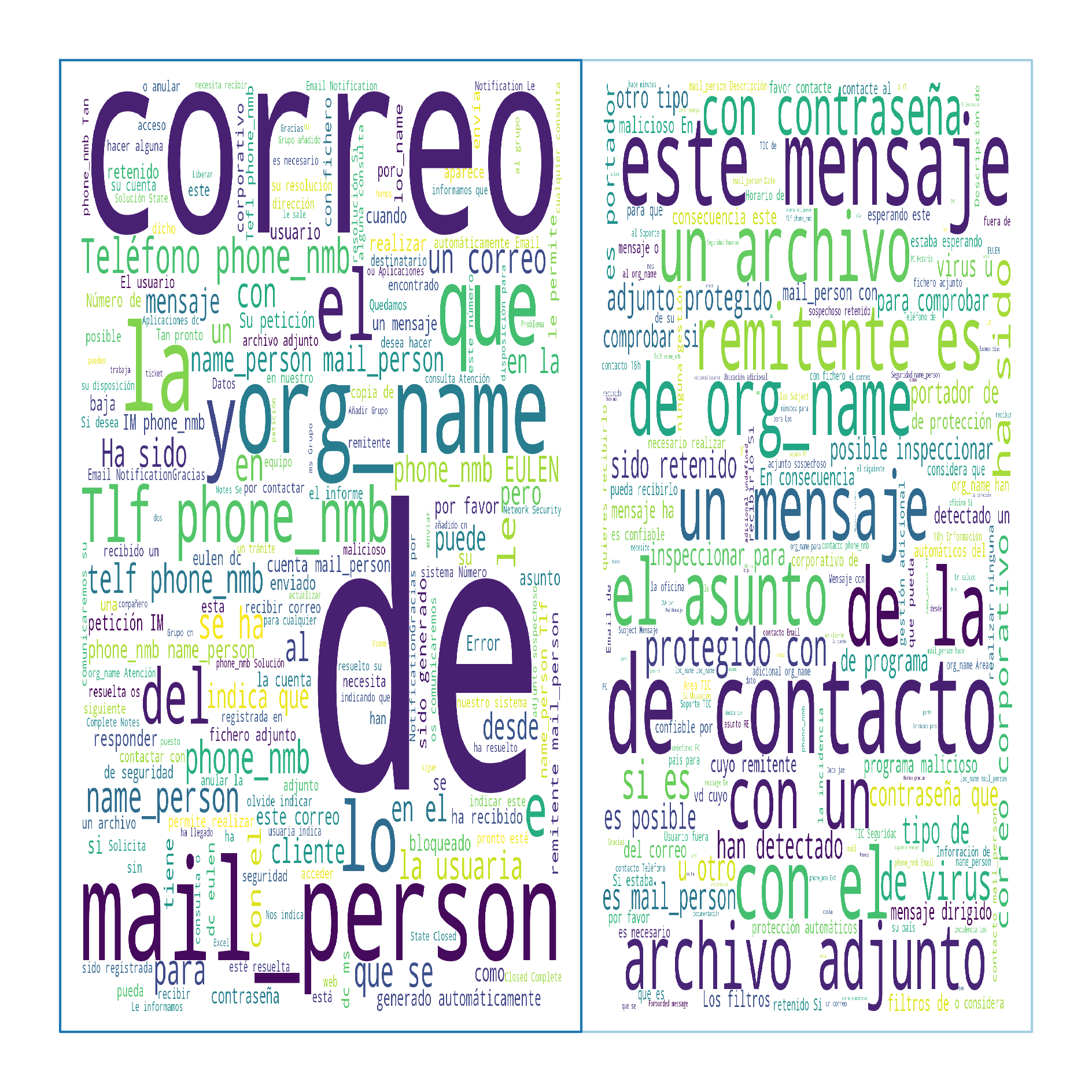}\\
    \multicolumn{5}{p{0.95\linewidth}}{(a) Class T12 PCA dimensions 0 (7.98\%) and 2 (4.06\%), from left to right: PCA scatter plot, RBF heatmap, division line, tag-treemap} \\[2mm]
    \includegraphics[height=45mm]{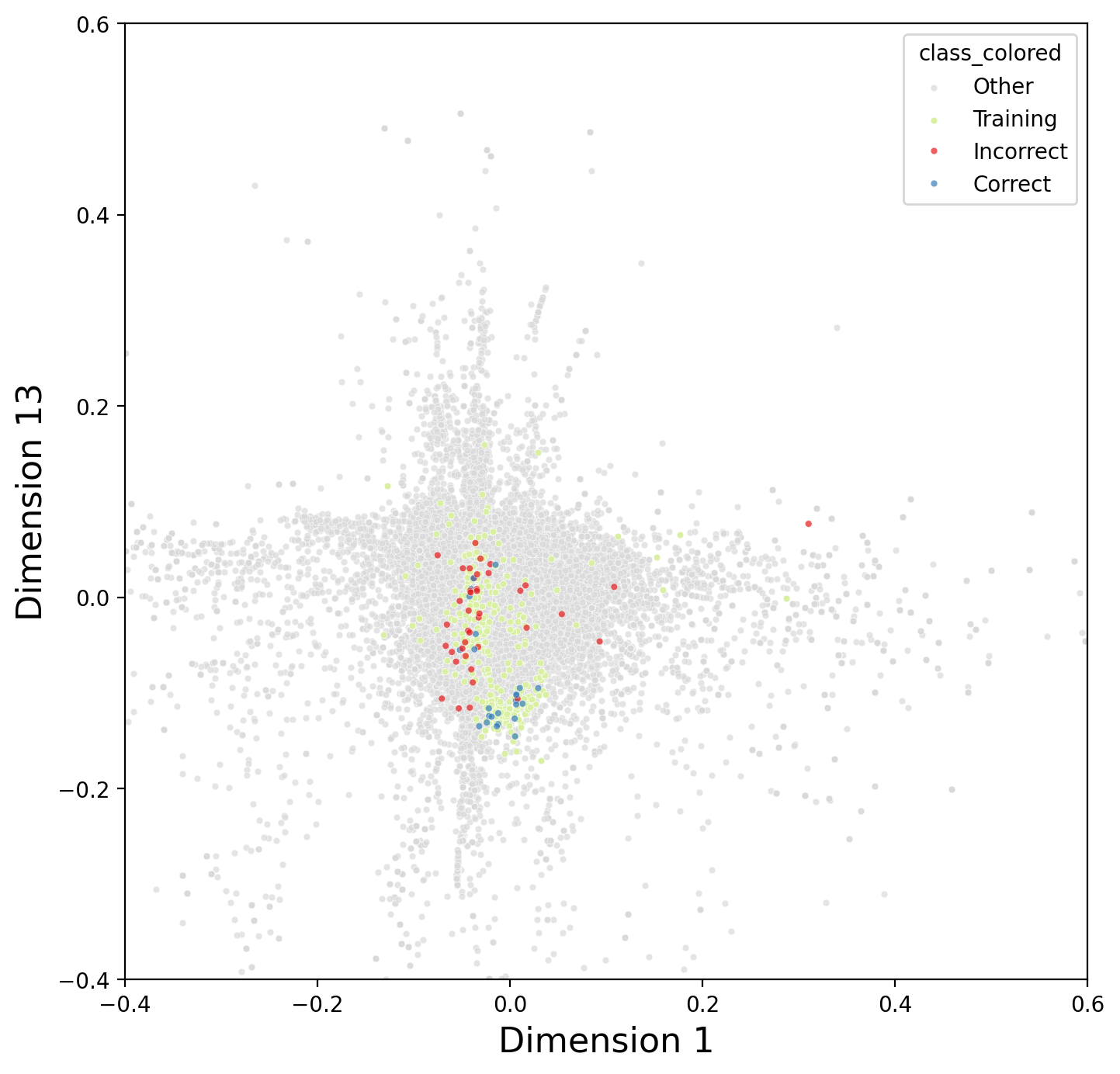} &
    \includegraphics[height=45mm]{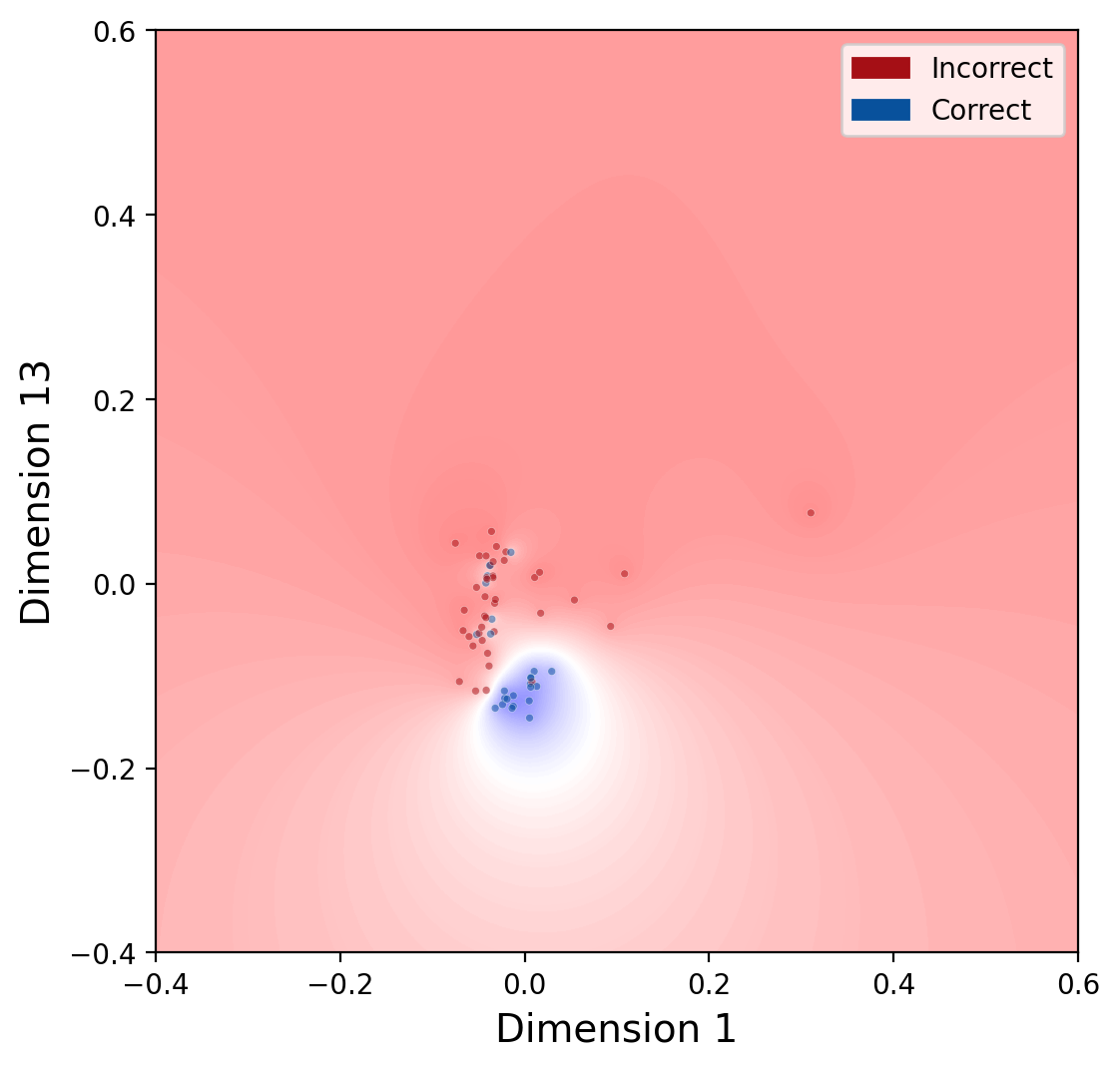} &
    \includegraphics[height=45mm]{figures/RBF_legend.png} &
    \includegraphics[height=18mm]{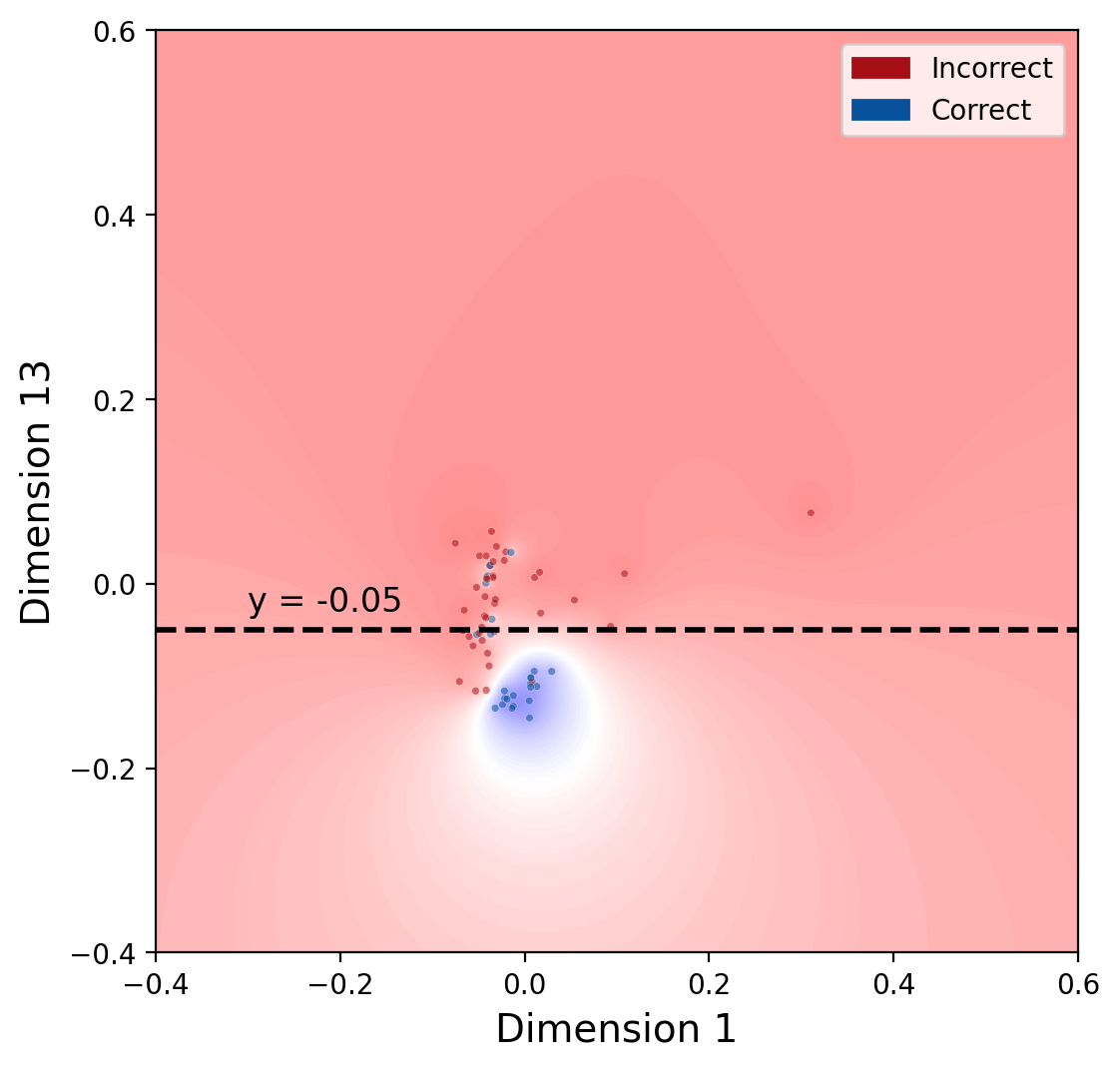} &
    \includegraphics[height=45mm]{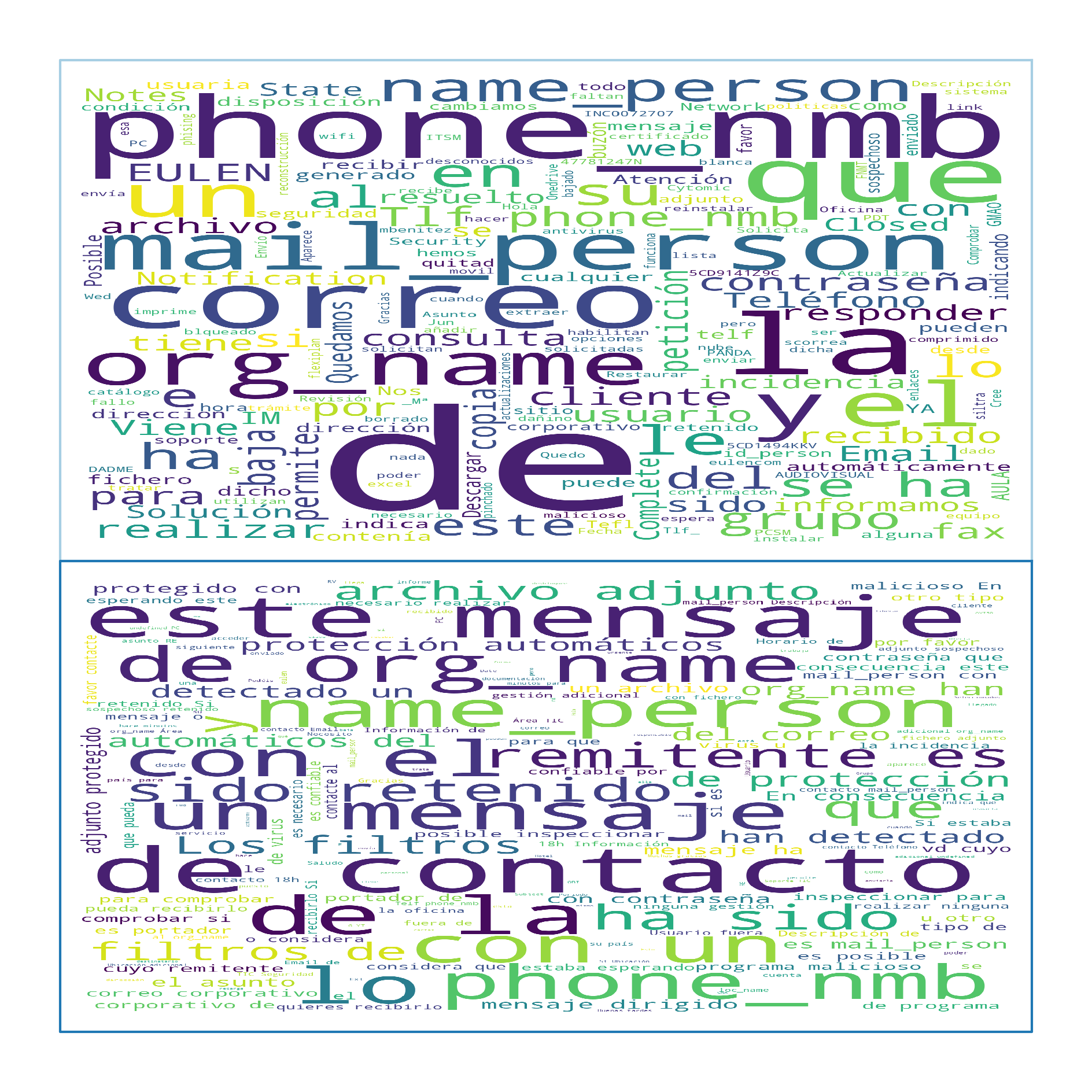}\\
    \multicolumn{5}{p{0.95\linewidth}}{(b) Class T12 PCA dimensions 1 (5.61\%) and 13 (0.74\%), from left to right: PCA scatter plot, RBF heatmap, division line, tag-treemap}
    \end{tabular} 
    \caption{Two examples of detailed visual analysis for investigating class T12. The two PCA scatter plots on the left show that Dimension 0 in (a) and Dimension 13 in (b) can separate the data objects into two regions, and data objects in one region have higher recall, while the overall class recall is only 37.5\%. Each RBF plot in the second column makes the boundary between the high-recall and low-recall regions clearer, enabling the selection of a division line to study the summary statistics of the messages in the two regions using a tag-treemap on the right.}
    \label{fig:PCA-RBF-TTM}
    \vspace{-4mm}
\end{figure*}

\subsection{t-SNE$\!$ (t-distributed$\!$ Stochastic$\!$ Neighbor$\!$ Embedding)}
t-SNE is an unsupervised non-linear dimensionality reduction that constructs a $k$-D data model based on a $n$-D dataset (typically $n \gg k$). When $k = 2$, the data model is commonly visualized using a scatter plot. Fig. \ref{fig:t-SNE} shows the application of t-SNE to the aforementioned dataset captured from an IT ticketing system. From the figure, we can observe two groups (or kinds) of patterns and each group has three subgroups:
\begin{itemize}
    \item[\textbf{P1:}] Data objects in a class (\textbf{P1a}) are clumped together, e.g., classes T1, T4; (\textbf{P1b}) form many small clusters, e.g., T2; or (\textbf{P1c}) are very much scattered around, e.g., T3.%
    \vspace{-1mm}
    \item[\textbf{P2:}] Data objects in a class are (\textbf{P2a}) relatively isolated from other classes, e.g., T11; (\textbf{P2b}) partly isolated and partly mingled, e.g., T5; (\textbf{P2c}) mostly mingled with other classes, e.g., T9.
\end{itemize}

\noindent In a t-SNE scatter plot, when data objects in a class are clumped together or isolated from other classes, it indicates that the two feature dimensions formulated by the t-SNE dimensionality reduction algorithm can be used to distinguish this class from others relatively easily. Likely a reasonably good ML model can learn similar features and hence classify this class accurately. When data objects in a class are mingled with others or scattered around, it indicates that the two feature dimensions used for plotting the data objects are not ideal for distinguishing this class from others. It suggests that an ML model may have some difficulties in classifying this class. However, since the ML model will likely learn many other features, some of which could still help achieve good performance. For example, in Fig. \ref{fig:t-SNE}, the visual patterns for classes T2 and T3 are not ideal, but their classification results are fairly good (Table \ref{tab:Dataset}). The ML model must have learned some useful features that are not characterized by the t-SNE plot.

\subsection{PCA (Principal Component Analysis)}
In order to visualize the data distribution of each class in relation to a good number of features, we use PCA to extract $K$ features (we set $K=20$ for the data in this paper). As shown in Fig. \ref{fig:PCA-RBF-TTM}, we use a scatter plot to visualize the data points in each class with the $x$ and $y$ axes corresponding to two dimensions $d_i, d_j$ such that $d_i, d_j \in [0, K-1]$, and $d_i \neq d_j$. In particular, we color the data objects that we correctly classified in blue and otherwise in red. As the visual context, data objects in other classes are shown in grey.

We start with viewing the scatter plots for those classes with relatively low recall (Table \ref{tab:Dataset}). For each of such classes, we first examine $K-1$ scatter plots with consecutive dimensions $(d_0, d_1)$, $(d_2, d_3), \ldots, (d_{18}, d_{19})$, ensuring that all dimensions are observed. When we identify more than one interesting dimension (e.g., $d_0$ and $d_6$), we also examine additional scatter plots (e.g., $(d_0, d_6)$). When we examine these scatter plots, we look for the following patterns:
\begin{itemize}
    \item[\textbf{P3:}] An area has more red dots than blue dots.
    \vspace{-1mm}
    \item[\textbf{P4:}] An area has only a few dots, some of which are red.
    \vspace{-1mm}
    \item[\textbf{P5:}] An area has red dots scattered among grey dots.
\end{itemize}

\noindent In Fig. \ref{fig:PCA-RBF-TTM}, there are two examples of PCA scatter plots, which show patterns of \textbf{P3} and \textbf{P5}. It focuses on class T12, within PCA dimensions 0 and 2 in (a) and dimensions 1 and 13 in (b). The data objects used for training are shown in yellow-green, testing data objects are shown in blue (if correct) or red (if incorrect), and the data objects in other classes are shown in grey to provide a holistic context. In (a), the colored dots on the left are largely red (incorrect), scattered among grey dots. In (b), the colored dots in the upper part exhibit a similar pattern.

In the context of the two feature dimensions plotted, \textbf{P3} indicates that a group of misclassified messages share similar features. This leads to hypothesis \textbf{H1:} \emph{Is it possible that the ML model relies heavily on features similar to these two PCA features?} 

\textbf{P4} indicates that these a few messages in the same class may differ from others in the same class (in terms of the two dimensions concerned). If this pattern is repeated for most dimensions, especially among the top dimensions (the first a few ones) corresponding to the principal components that encode more information about the data, this leads to hypothesis \textbf{H2:} \emph{Is it possible that the ML model has not learned suitable features for classifying these messages?}    

\textbf{P5} indicates that some messages in the class may appear quite similar to those in other classes. Similar to \textbf{P4}, if the pattern is repeated among most dimensions, including the top dimensions, this leads to hypothesis \textbf{H3:} \emph{Is it possible that the ML model relies on these features for classifying messages in some other classes but they may not be suitable for classifying messages in this class?} 

ML model developers usually have some knowledge and skills to investigate the hypotheses \textbf{H1}, \textbf{H2}, and \textbf{H3} further numerically.

\subsection{RBF (Radial Basis Function)}
From the PCA scatter plots in Fig. \ref{fig:PCA-RBF-TTM}, one cannot always judge easily whether there are more red or blue dots if the numbers of correct and incorrect results are not significantly different, because counting demands costly cognitive effort. Furthermore, ML developers are often curious about areas between sampled data objects in a class, since PCA assumes that each feature dimension is continuous, and potentially there are other messages that could have feature values between the known values of two sampled messages. We therefore use a radial basis function (RBF) to estimate the recall error rate across the 2D feature space depicted in a 2D PCA scatter plot. In Fig, \ref{fig:PCA-RBF-TTM}, the second column shows two heatmaps after applying an RBF to the two PCA scatter plots on the left. We superimpose the testing data objects, as dots, on the top of the RBF heatmap. This allows viewers to judge if a pixel in the heatmap is further away or close to the tested data objects, providing intuitive and implicit uncertainty information about the colors in the RBF heatmap. From an RBF plot, one can observe the following patterns:
\begin{itemize}
    \item[\textbf{P6:}] An area that has a high or low recall error rate (when there are many data objects nearby).
    \vspace{-1mm}
    \item[\textbf{P7:}] An area that may \textbf{potentially} have a high or low recall error rate (when not many data objects are nearby).
\end{itemize}

\noindent When one is viewing an RBF heatmap, usually one cannot help hypothesize \textbf{H4:} \emph{Will there be dots between the two known dots?} Technically, the hypothesis is: \emph{Are there messages whose feature values will fall between the values of known messages?}

\subsection{Tag-Treemap: Combining Tag Cloud and Treemap}
In a PCA scatter plot, when two dots are closely located, it indicates that in terms of the two principal components (feature dimensions) depicted by the scatter plot, the two text messages (data objects) are similar. However, viewers may not be able to interpret what the similarity means in terms of the texts concerned, e.g., uses of words, sentence structure, and so on. While one can always read the actual messages to consolidate the interpretation of distances in a PCA scatter plot, this can be time-consuming. We thus combine two visual representations, tag cloud and treemap (referred to as \emph{tag-treemap}), to depict $k$ tag clouds associated with $k$ subsets of texts that belong to the same parent set.

For example, we can define all testing data objects in a class as a parent set. We can consider two subsets, for the data objects classified correctly and those incorrectly. We can also divide testing data objects based on their features. In Fig. \ref{fig:PCA-RBF-TTM}, the two RBF heatmaps can be divided into two regions, and we can enrich the relatively abstract patterns in the PCA scatter plots and RBF heatmaps by juxtaposing the tag clouds for these regions as shown on the right of Fig. \ref{fig:PCA-RBF-TTM}. Each of these two tag-treemaps reveals the characteristic difference between the two regions. In (a), the tag clouds on the left and right convey the keyword statistics of 38 and 26 text messages respectively. We can observe the right (i.e., the more accurate part) has a much evener distribution of keywords than the left. Likely the ML model has learned some features similar to PCA Dimension 0 and may treat these two parts differently. In (b), the tag clouds above (33 messages) and below (31 messages) show similar divergence between the two parts, though the divergence is not as substantial as (a). We also use similar tag-treemaps to check whether the testing data has the same keyword statistics as the training data and all data objects in the class. In this case, the checking confirms the similarity. If there were noticeable divergences, the ML developers could resample the testing and training data to ensure that the two datasets could represent the whole class.

Using tag-treemaps, one can observe: 
\begin{itemize}
    \item[\textbf{P8:}] The summary statistical patterns about the keywords in two or more groups of data objects.
\end{itemize}

\noindent For example, one can compare (\textbf{P8a}) a group of closely clustered data objects with other data objects in the same class, (\textbf{P8b}) red dots vs. blue dots, (\textbf{P8c}) different classes, and so on.  

In the next section, we will discuss that the hypotheses \textbf{H1}$\sim$\textbf{H4} lead to the approach of data synthesis.

\begin{figure*}[t]
    \centering
    \begin{tabular}{@{}cc@{}cc@{}}
        \includegraphics[height=48mm]{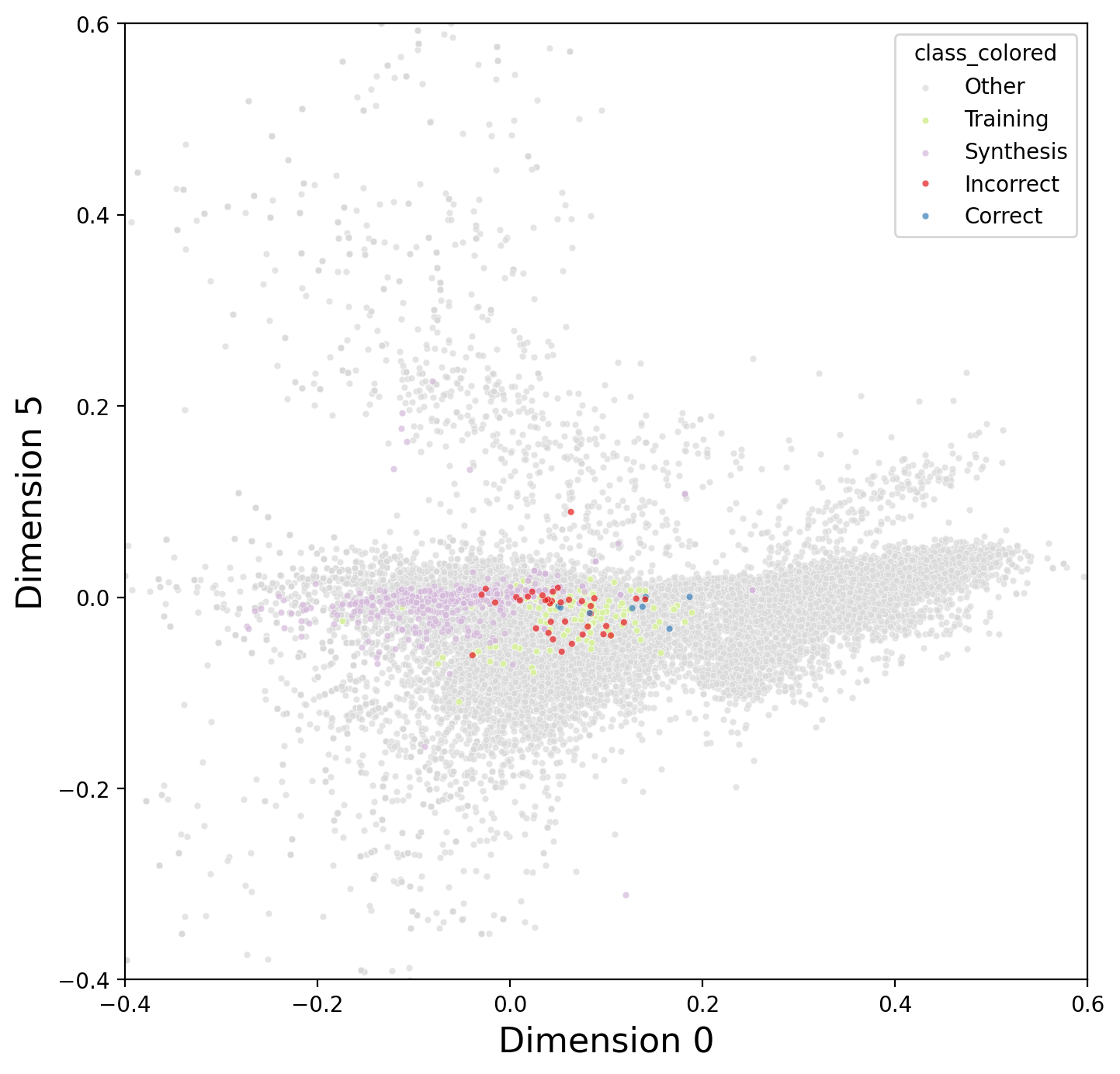} &
        \includegraphics[height=48mm]{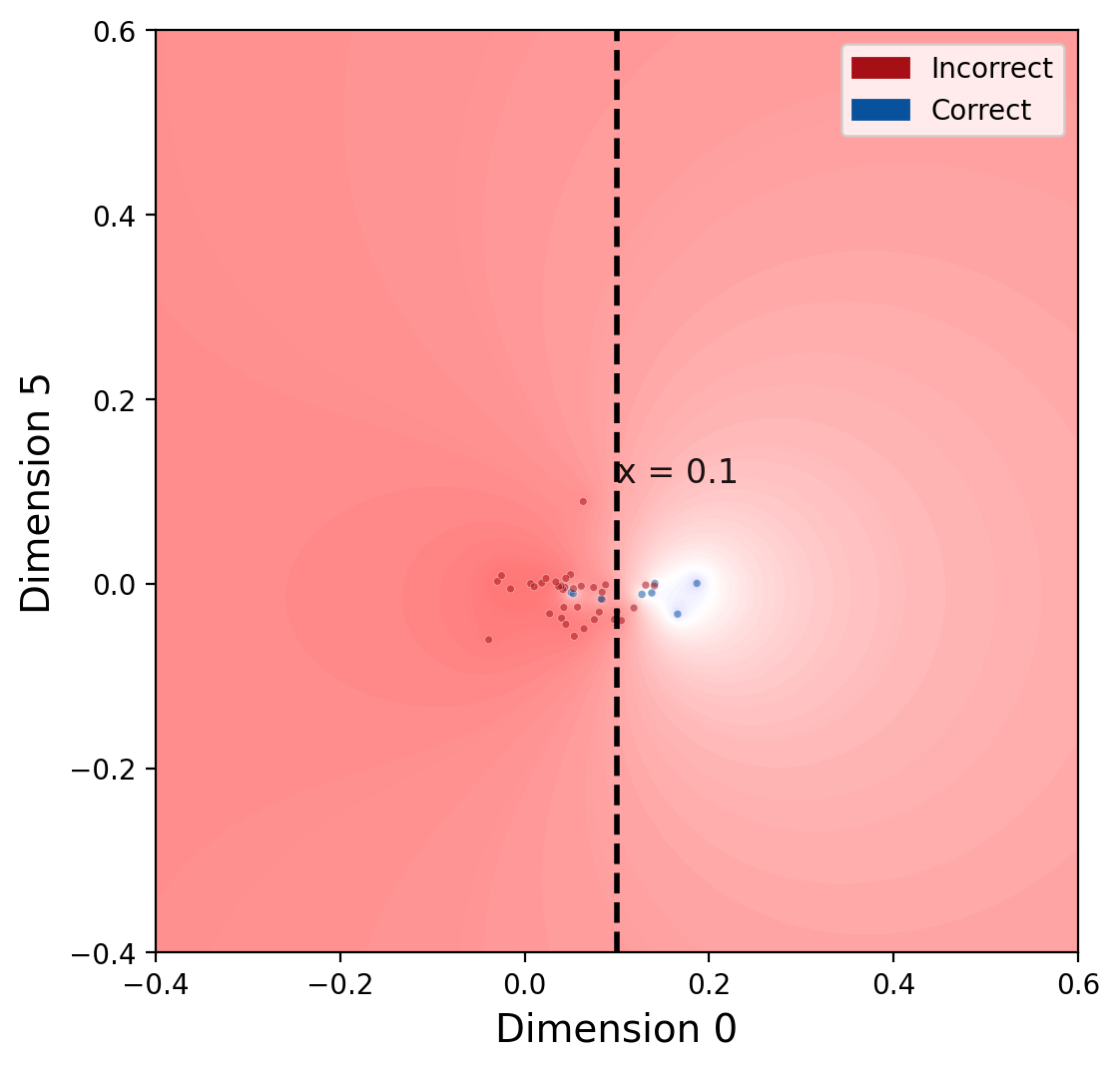} &
        \includegraphics[height=48mm]{figures/RBF_legend.png} &
        \includegraphics[height=48mm]{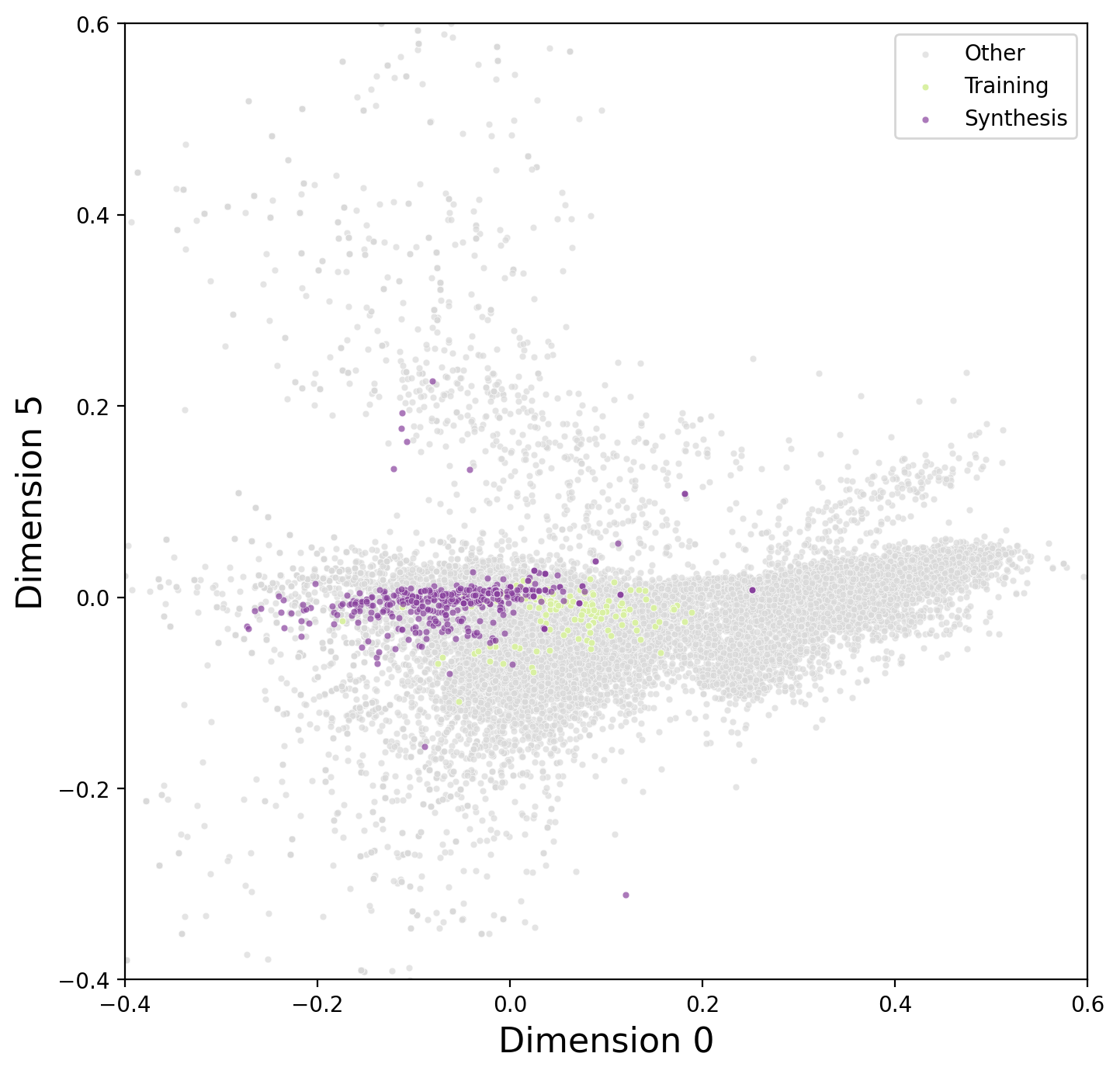} \\
        (a) correctness distribution (before) &
        \multicolumn{2}{c}{(b) two area divided by dimension 0} & 
        (c) distribution of synthesized data \\[2mm]
        \includegraphics[height=48mm]{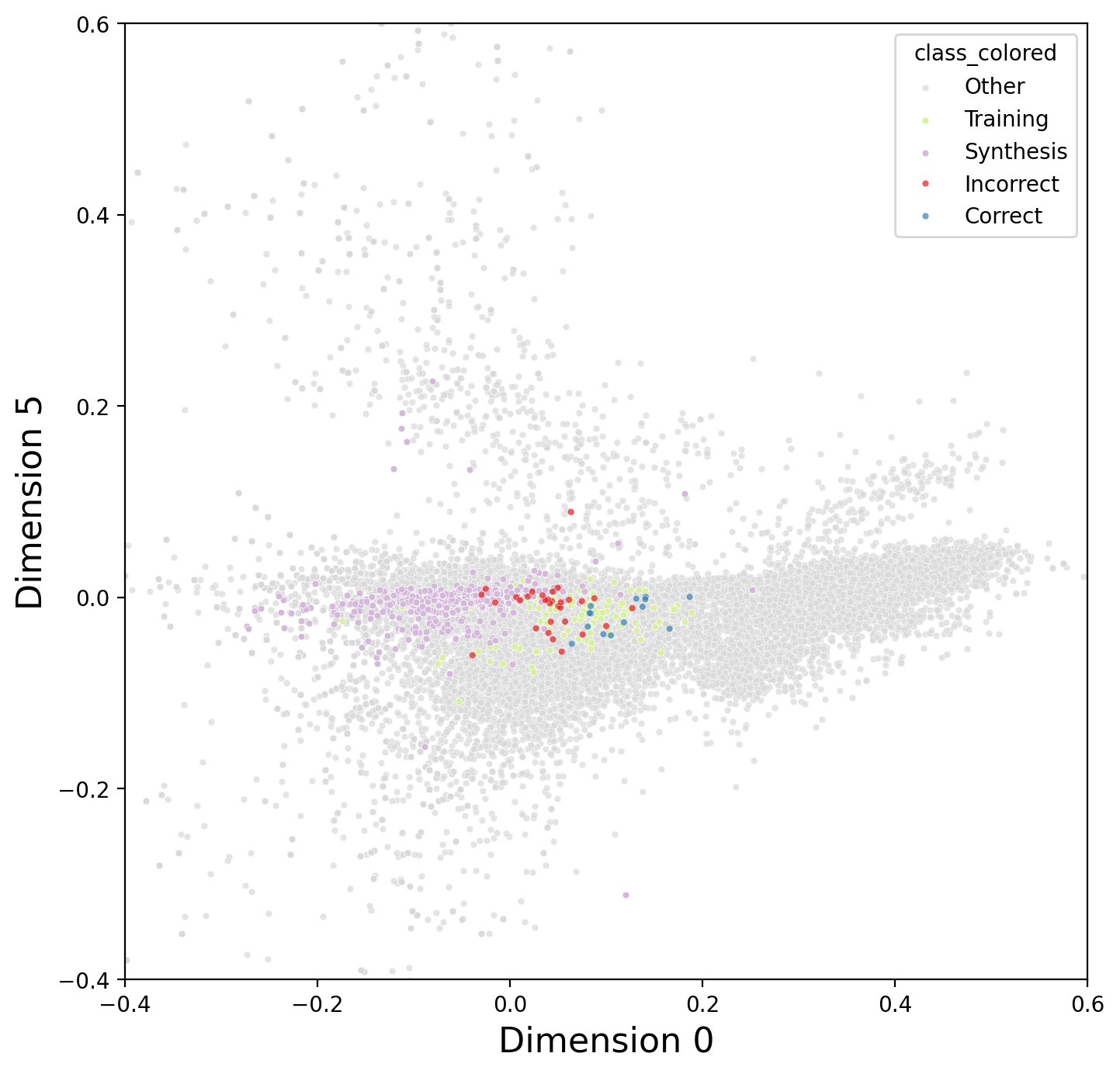} &
        \includegraphics[height=48mm]{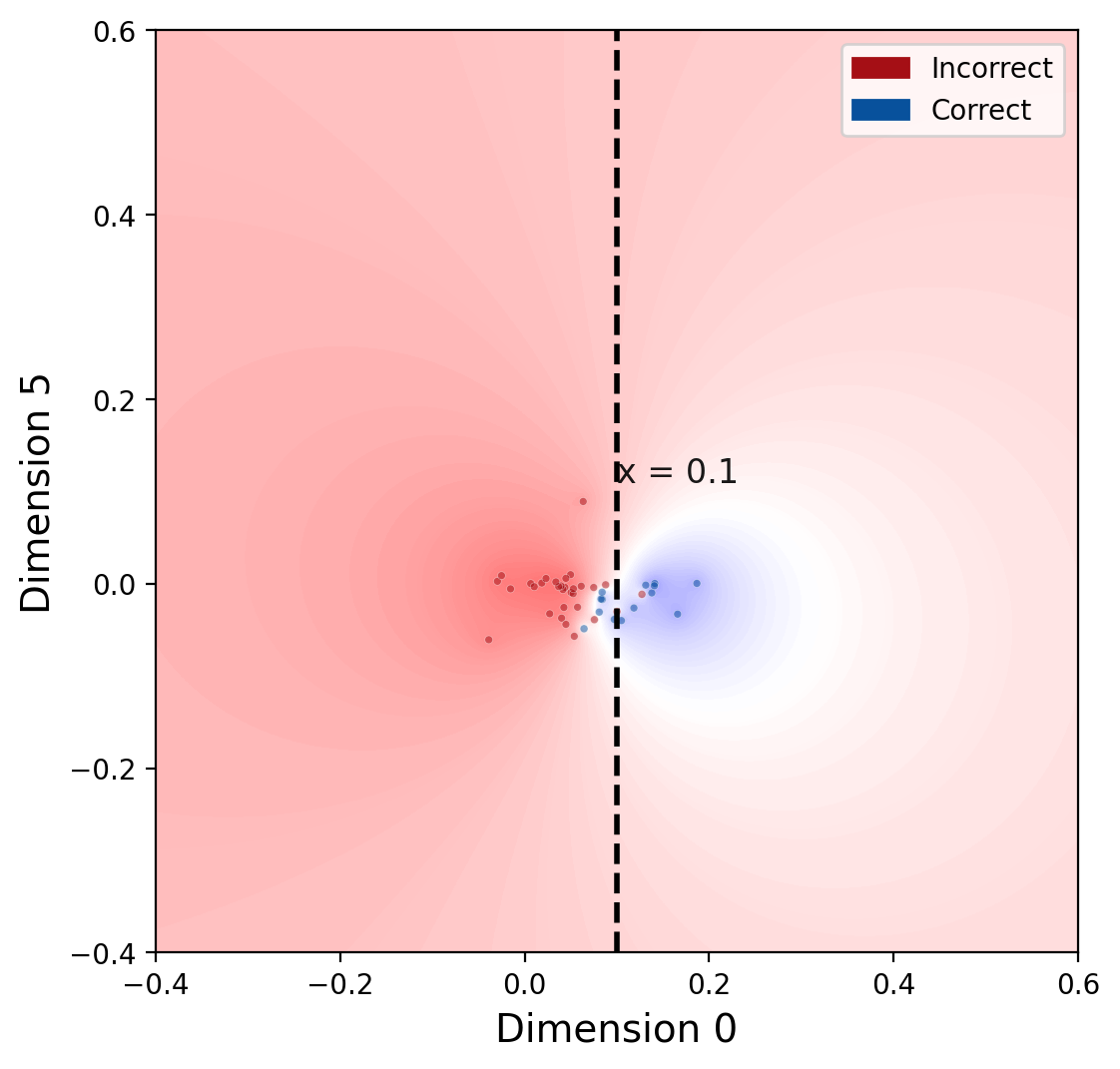} &
        \includegraphics[height=48mm]{figures/RBF_legend.png} &
        \includegraphics[height=48mm]{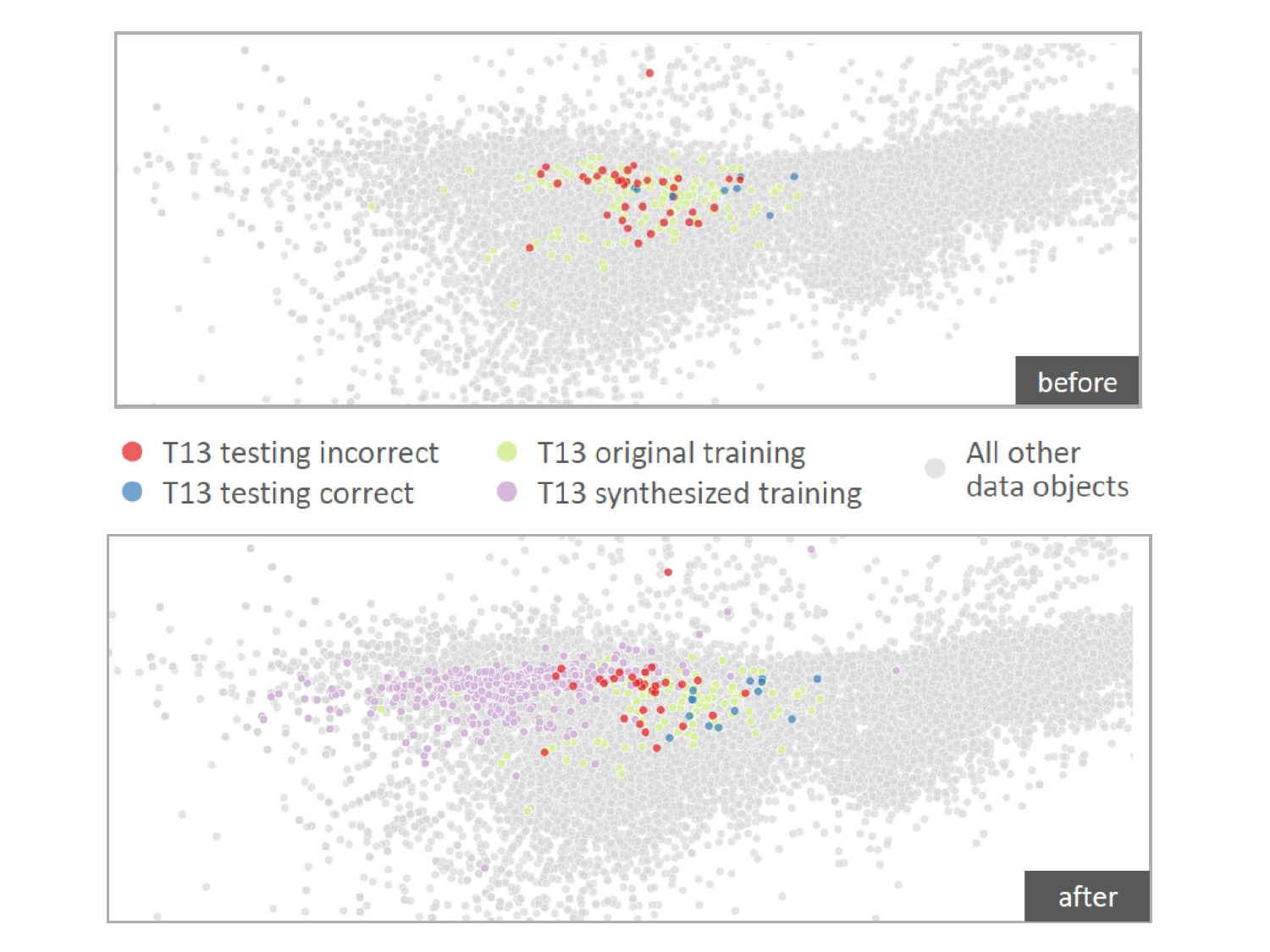} \\
        (d) correctness distribution (after) &
        \multicolumn{2}{c}{(e) noticeable improvement in RBF}  &
        (f) zoomed PCA scatter plots\\
    \end{tabular}
    \caption{The class T13 has the lowest recall among all classes. The scatter plot in (a) indicates more classification errors (red dots) when the data objects are associated with lower values in PCA feature dimension 0. The RBF heatmap in (b) confirms this pattern and enables data synthesis to be targeted at an erroneous cluster on the left as shown in (c). The model retrained with additional LLM-synthesized data is improved in (d). The RBF heatmap for the new testing results in (e) and the zoomed-in scatter plots confirm the improvement.}
    \label{fig:T13}
    \vspace{-4mm}
\end{figure*}

\begin{table*}[t]
\centering
\caption{Example testing results with synthetic data. The \textbf{test} column shows the number of data objects in the testing dataset, which is used to obtain all testing results in the table. The \textbf{train} columns show the numbers of data objects for training. In the five cases of synthetic data, a +num indicates the number of synthetic text messages that were added to the specific class in the training data, while the blank cell indicates that no synthetic data was added to a class. The five $\Delta$-\textbf{recall} columns show the difference between the original recall (column 4) and recall values obtained from testing the five models trained with the corresponding training data (i.e., original training data + synthetic data). More testing results can be found in Appendix \ref{apx:Results}\revise{, including favorable comparison with tests based on randomly selected examples.}}
\label{tab:Synthetic}
\centering

\begin{tabular}{|c|ccc|c@{\hspace{3mm}}c|c@{\hspace{3mm}}c|%
c@{\hspace{3mm}}c|c@{\hspace{3mm}}c|c@{\hspace{3mm}}c|c@{\hspace{3mm}}c}
\textbf{Topic} & \multicolumn{3}{c|}{\textbf{Original Data}}
    & \multicolumn{2}{c|}{\textbf{T11-s1}}
    & \multicolumn{2}{c|}{\textbf{T12-s1}} 
    & \multicolumn{2}{c|}{\textbf{T13-s1}}
    & \multicolumn{2}{c|}{\textbf{T14-s1}}
    & \multicolumn{2}{c|}{\textbf{T15-s1}}
    \\
\textbf{class} & \textbf{test} & \textbf{train} & \textbf{recall}
    & \textbf{train} & $\Delta$-\textbf{recall}
    & \textbf{train} & $\Delta$-\textbf{recall}
    & \textbf{train} & $\Delta$-\textbf{recall} 
    & \textbf{train} & $\Delta$-\textbf{recall} 
    & \textbf{train} & $\Delta$-\textbf{recall} 
    \\ \hline
Overall & 7820 & 31280 & 0.821 &      & \up 0.003 &      & \up 0.009 &      & \up 0.009&      & \up 0.001 &      & \up 0.004 \\ \hline
T1  & 1748 & 6781      & 0.976 &      & \down 0.002 &    & \up 0.003 &     & \down 0.001 &    & \up 0.005 &    & \up 0.005\\
T2  & 2259 & 9091 & 0.943 &     & \up 0.002 &     &\up 0.004&     &\down 0.003&    & \down 0.003 & & \up 0.001\\
T3  &  927 & 3792 & 0.892 &     & \down 0.009 &     &\down 0.005&     &\down 0.004&    & \down 0.019 & & \down 0.012\\
T4  &  281 & 1106 & 0.769 &     & \up 0.039 &     &\up 0.060&     &\up 0.028&    & \up 0.010 & & \up 0.050\\
T5  &  533 & 2222 & 0.732 &     & \up 0.020 &     &\up 0.005&     &\up 0.009&    & \up 0.015 & & \down 0.023\\
T6  &  373 & 1515 & 0.528 &     & \up 0.030 &     &\up 0.059&     &\up 0.038&    & \up 0.035 & & $\enskip$ 0.000\\
T7  &  395 & 1568 & 0.714 &     & \up 0.013 &     &\up 0.040&     &\up 0.025&    & \down 0.020 & & \up 0.018\\
T8  &  236 &  792 & 0.508 &     & \down 0.055 &     &\down 0.046&     &\up 0.077&    & \down 0.046 & & \down 0.067\\
T9  &  278 & 1188 & 0.626 &     & \up 0.011 &     &\up 0.054&     &\down 0.018&    & \down 0.032 & & \up 0.032\\
T10 &  335 & 1364 & 0.427 &     & \up 0.006 &     &\up 0.018&     &\up 0.095&    & \up 0.033 & & \up 0.048\\
T11 &   94 &  377 & 0.926 & +110 & \up 0.021 &     &\up 0.010&     & $\enskip$ 0.000 & & $\enskip$ 0.000 & & \up 0.010\\
T12 &   64 &  294 & 0.375 &     & $\enskip$ 0.000 & +325 &\up 0.031&     &\down 0.016& & $\enskip$ 0.000 & & \down 0.078\\
T13 &   45 &  135 & 0.178 &     & \up 0.155 &     &\down 0.022& +525 &\up 0.133&    & \up 0.066 & & \up 0.066\\
T14 &  135 &  629 & 0.526 &     & \down 0.037 &     &\up 0.007&     &\down 0.022& +1265   & \up 0.148 & & \up 0.015\\
T15 &  117 &  426 & 0.376 &     & \down 0.034 &     &\down 0.060&     &\down 0.034&    & \down 0.034 & +1485 &  \up 0.094\\
\hline
\end{tabular}
\vspace{-4mm}
\end{table*}

\section{LLMs for Synthesizing Training Data}
\label{sec:LLM}
As illustrated in the second workflow in Fig. \ref{fig:Workflows}, the VIS process helped ML developers formulate hypotheses about data-related causes of erroneous results, allowing ML developers to shift their focus from fine-tuning hyper-parameters during the previous months (in the first workflow) to improving training data. As it was not feasible to collect more training data, we decided to experiment with synthetic data generated using large language models (LLMs), i.e., requirement \textbf{R2} (Section \ref{sec:Overview}).

A Generative Pre-trained Transformer (GPT) is a type of LLM. GPT-3, which was developed by OpenAI, includes an ``attention'' mechanism that is able to predict which segments of input text are most relevant, enabling LLMs to focus on these segments \cite{Bahdanau:2015:ICLR}. GPT-3.5 is a sub-class of GPT-3 Models. Based on the GPT-3.5 architecture, Inetum developed a customized LLM, which was optimized for specific tasks and scenarios. The 2023-03-15-preview version of GPT-3.5 API was used in the development. This customized LLM can take some text examples and generate new realistic texts in a range of topics, including the 15 topics in Table \ref{tab:Dataset}.

For example, consider a message in the class T12,
\begin{itemize}
    \item[] ``\emph{Telefono:} <phone number> \emph{Solicita que todas las carpetas de drive de las cuales era propietario mail person (ha causado baja en la empresa) pasen a su usuario mail person}''
\end{itemize}
\noindent which is one of the red dots in the PCA scatter plot in Fig. \ref{fig:PCA-RBF-TTM}(a) was misclassified by an existing ML model. We can use this as a piece of input text to the Inetum LLM. With appropriate parameters such as ``temperature = 0.7'', ``max tokens = 550'', ``top p = 0.5'', ``frequency penalty = 0.3'', ``presence penalty = 0.0'', and so on, the LLM model generates $k$ similar messages as the output:

\begin{enumerate}
    \item \emph{Tel\'{e}fono: 555-5618 Solicita que se actualice la informacion de contacto en el sistema con su nueva direccion de correo electrónico: new email@company.com}
    \vspace{-2mm}
    \item \emph{Tel\'{e}fono: 555-4312 Solicita que se elimine la cuenta de correo electr\'{o}nico asociada al usuario "maria perez" y se transfieran todos los correos a la cuenta de "juan rodriguez}
    \vspace{-2mm}
    \item $\ldots$ Other $k-2$ synthesized messages
\end{enumerate}

Without VIS techniques described in Section \ref{sec:VIS}, we would naturally target all classes with relatively poor recall, e.g., T6, T8-T10, T12-T15. We would have to select example text messages (i.e., data objects) randomly from all data objects in a class. With the VIS techniques, we can target data synthesis to more specific subareas in many ways, e.g.,%
\begin{itemize}
    \item[A.] In a t-SNE scatter plot, there is a relatively isolated class, but its classification results are not satisfactory. Alternatively, in a PCA scatter plot, there is a relatively isolated, small, red clump of data objects. Both scenarios suggest that there are distinct features to enable correct classification, but the ML model may have not learned such features, possibly because there is not enough training data in this subarea of the plot. \textbf{Possible Action:} We can select example text messages from this subarea and synthesize more data for training.
    \vspace{-1mm}
    \item[B.] In a PCA scatter plot, when there is a red clump of data objects mingled with grey data objects, it suggests that the ML model may confuse these text messages with those in other classes, possibly because in this subarea, there are substantially more text messages from other classes (shown in grey) than the class concerned (shown in red or blue). \textbf{Possible Action:} We can select example text messages from this subarea and synthesize more to make the training data more balanced in this subarea. We can anticipate that it is possible that this action may improve the testing results for this class, but may impair those of other classes. We hope that the trade-off is in favor of the improvement.
    \vspace{-1mm}
    \item[C.] In the RBF plot, there are a few scattered red data objects that seem to make a large area red. After a close examination of these text messages directly or using a tag cloud, one notices that these messages are not uncommon in the real world. Likely the data collection was not comprehensive enough, making them appear to be outliers.
    \textbf{Possible Action:} We can select these text messages as examples for synthesizing more training data. Because we are still using collected real-world data for testing, the data objects in this subarea remain to be sparse. Hence the improvement to the testing results may appear to be small. As long as there is improvement, we should consider that the ML model becomes more robust in this subarea.
    \vspace{-1mm}
    \item[D.] In a tag-treemap, when one compares the keywords of two sets of text messages, one may notice that the one with poor testing results may have less expected ordering of keywords. This suggests a possibility that the collected data in this area may be stewed in favor of certain keywords, causing the ML model to have overlooked some other keywords that can better differentiate this class from other classes. \textbf{Possible Action:} We can select text messages with possibly overlooked keywords to synthesize more training data. Similar to scenario B, we need to observe the trade-off between improvement and impairment.     
\end{itemize}

Fig. \ref{fig:PCA-RBF-TTM} suggests that T12 can potentially be divided into two parts based on PCA Dimension 0 or Dimension 13.  

As shown in Table \ref{tab:Dataset}, Class T13 has the lowest recall. One of the PCA scatter plots, as shown in Fig. \ref{fig:T13}(a), reveals that the data objects in the class also exhibit two parts, the left part is less accurate than the right. The RBF heatmap in Fig. \ref{fig:T13}(b) helps us determine a separation line. We then select example messages from the training data on the left part and generate 525 synthetic messages as additional training data. Fig. \ref{fig:T13}(c) shows these synthetic data objects in purple. After retraining the model, the testing results show that the recall of the class improved from 18\% to 31\%. Fig. \ref{fig:T13}(d) indicates noticeable changes in the blue region and the reduction of the shade of red. Fig. \ref{fig:T13}(e) juxtaposes two zoomed-in PCA scatter plots and we can observe the changes of some red dots to blue dots.

As shown in column T13-s1 in Table \ref{tab:Synthetic}, the synthetic data helps improve the recall of T13 by 13.3\%. \revise{Adding synthetic data to T13 training data} has a positive impact on the recall of other six classes, including T4 by 2.8\%, T5 by 0.9 \%, T6 by 3.8\%, T7 by 2.5\%, T8 by 7.7\%, and T10 by 9.5\%. Although the recall of seven topics was reduced, the overall recall improved by nearly 1\%.

In order to maintain the testing consistently, we define a testing dataset with 20\% of data objects from each class. The other 80\% data is used for training. Because we have some very small classes, e.g., T11-T15, such a training-testing division is unavoidable. As described in Section \ref{sec:VIS}, we use tag-treemap to compare the training and testing datasets for each class to ensure that they both represent the class in a similar way.  
When we generate synthetic data using the LLM, we always select example messages from the training data as we are aware that selecting examples from testing data could introduce biases in favor of testing.  

There are many other scenarios, including those involving the use of multiple plots. For instance, an ML developer may first observe two classes overlapping in a t-SNE plot, and then examine closely the PCA plots for both classes in different combinations of dimensions. The ML developer may notice that one of the two classes has poor testing results in the subarea of the scatter plots for two specific dimensions. The ML developer may drill down the details by visualizing the tag-treemap for this subarea, with two tag clouds for the two classes respectively, leading to a scenario of D. There are many different lines of investigations, and hence the above scenarios A$\sim$D represent only a small subset of possible scenarios.

Likely, the more ML developers have 
visualized different patterns and conducted experiments to target data synthesis at different pattern-stimulated hypotheses, the more comprehensive knowledge they will gain about different visual patterns, their association with different hypotheses, and their connection with different parameter settings for data synthesis, and their correlation with the performance of the ML models retrained with additional synthetic data.  

\begin{figure*}[t]
  \centering
  \begin{tabular}{@{}cc@{}}
    \includegraphics[scale=0.27]{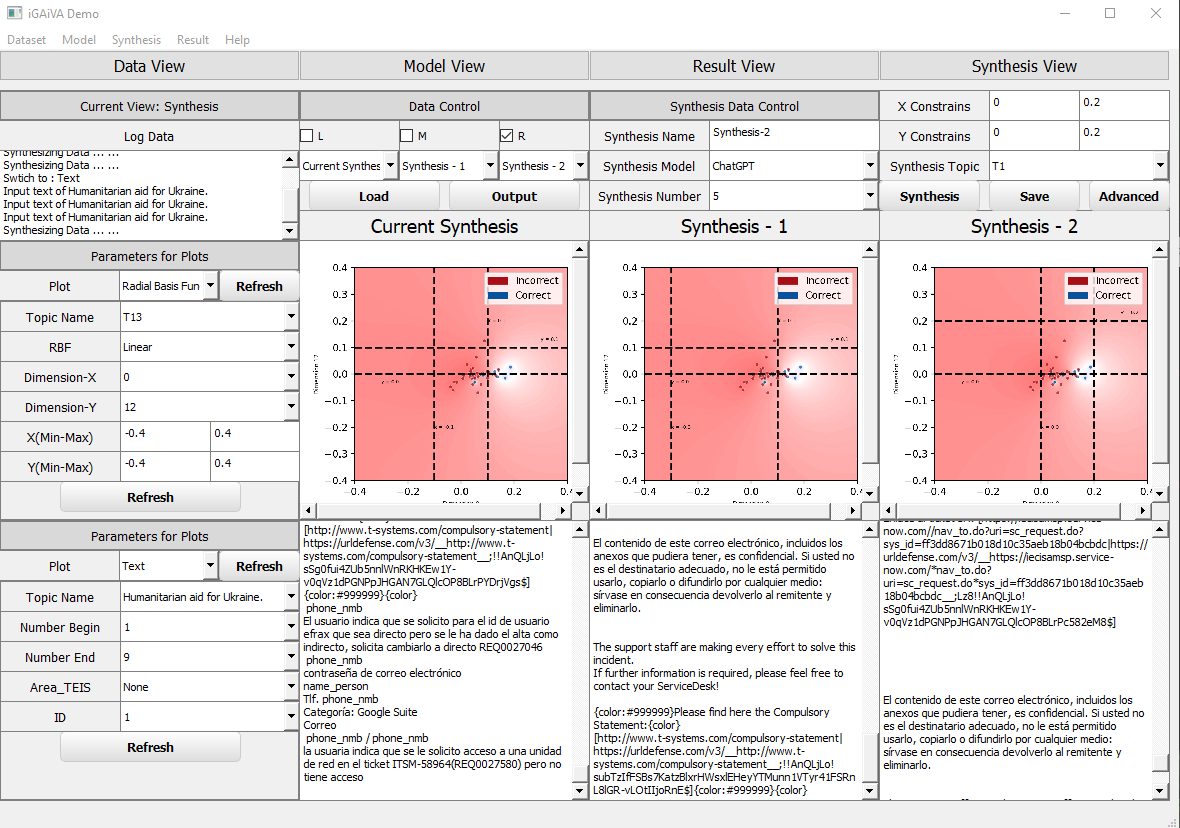} &
    \includegraphics[scale=0.27]{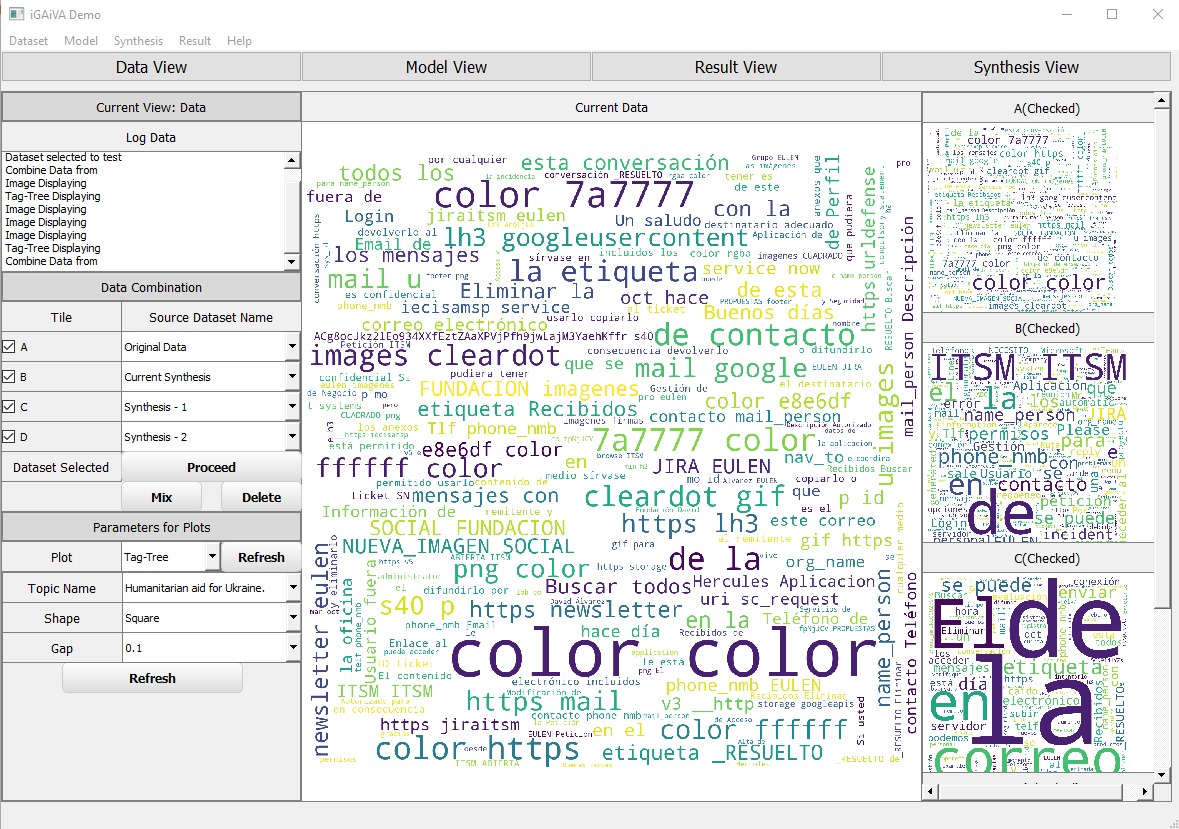} \\
    (a) Synthesis View & (b) Data View\\
    \includegraphics[scale=0.27]{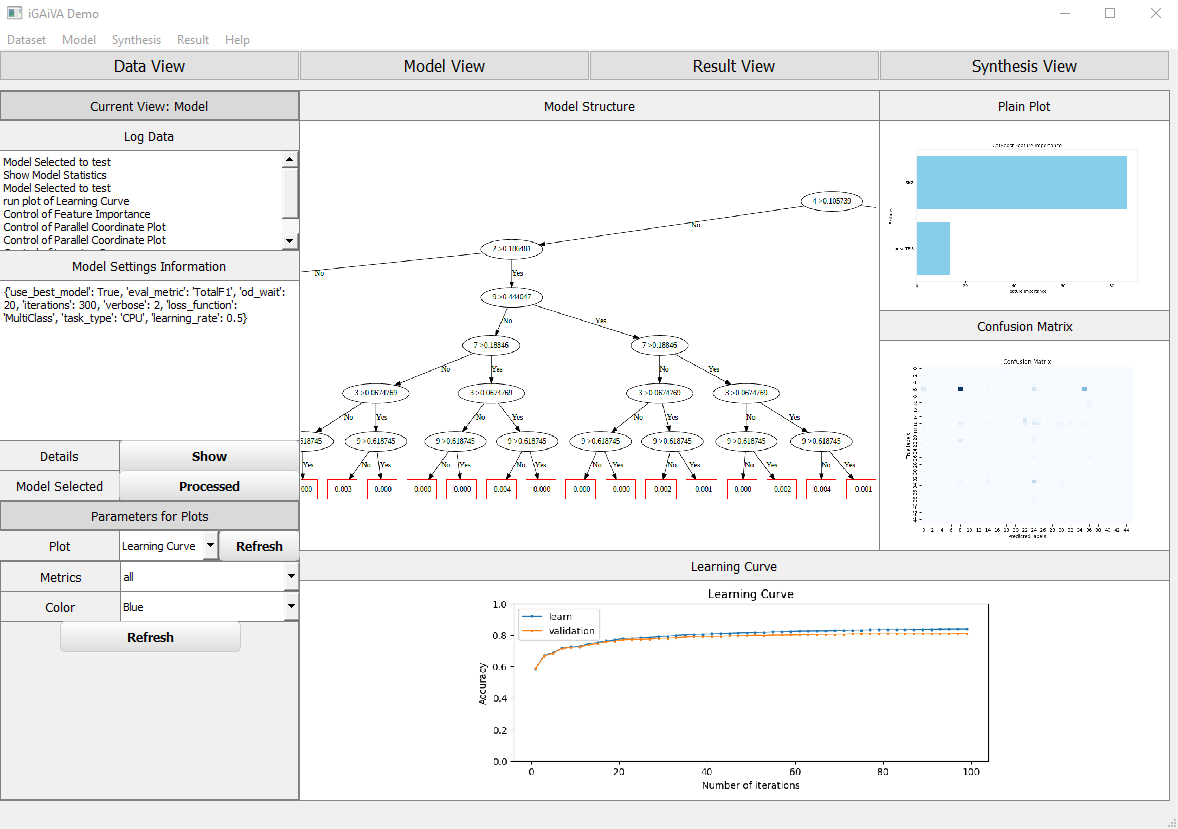} &
    \includegraphics[scale=0.27]{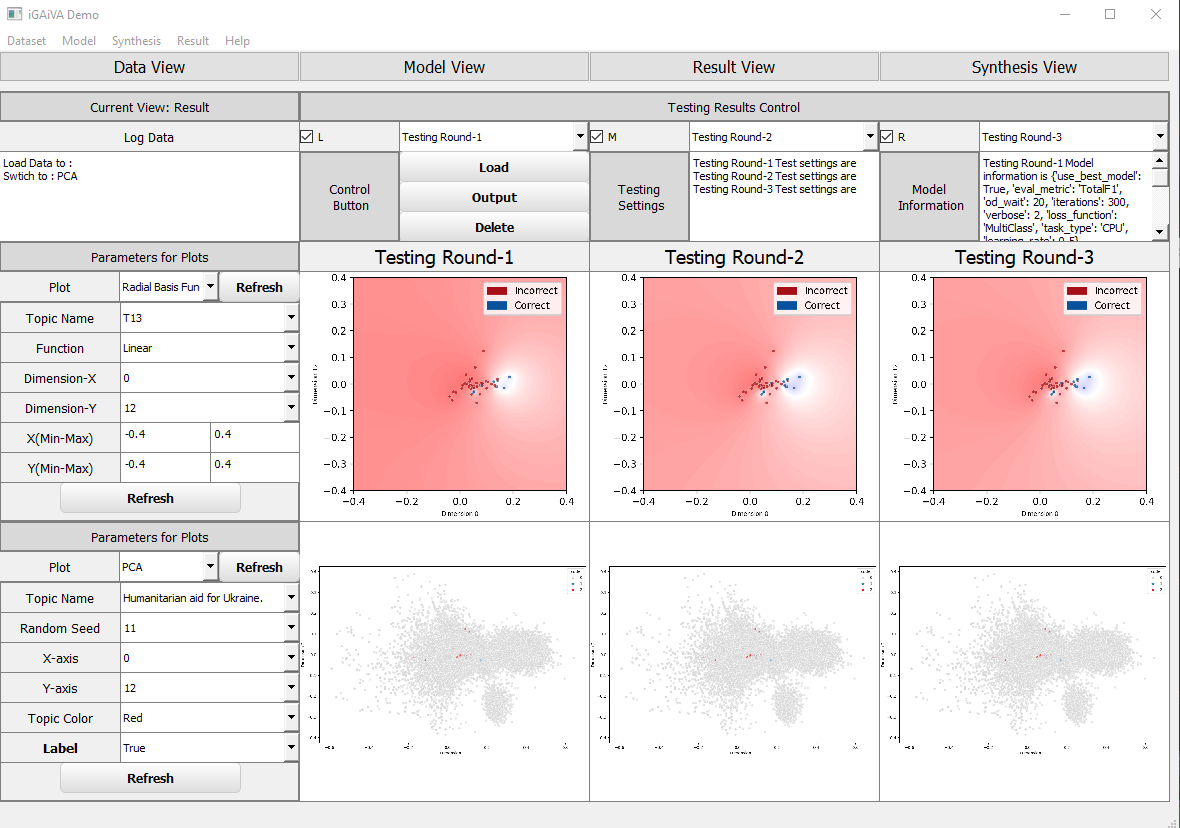} \\
    (c) Model View & (d) Results View
  \end{tabular}
  \caption{Four views of the iGAiVA tool. The user can switch between views using the top menu bar. (a) The Synthesis View is for supporting mainly the tasks for identifying suitable example data objects as inputs to LLMs for generating synthetic data. (b) The Data View is for selecting a subset of synthetic datasets and combining them with the original training data. (c) The Model View is for monitoring the process of retraining a model and running the retrained model against one or more predefined testing datasets. (d) The Results View is for analyzing and evaluating the results of the model's performance.}
  \label{fig:iGAiVA}
  \vspace{-6mm}
\end{figure*}

\section{iGAiVA: A Tool for Data Synthesis in ML}
\label{sec:iGAiVA}
In the previous two sections, we reported the successful use of VIS and LLM techniques for improving the performance of ML models for text classification. As mentioned in Section \ref{sec:Overview}, this successful experience led to the requirement for designing and prototyping a VIS4ML tool where VIS and LLM techniques would be integrated (\textbf{R3}). When we were working on the VIS and LLM techniques in the experimental workflow, we used VIS techniques, including PCA scatter plots, RBF heatmaps, and tag-treemaps frequently. On average, we would view at least 10 plots (e.g., Fig. \ref{fig:PCA-RBF-TTM}) before a run of LLMs to generate some synthetic data. After the synthetic data was generated, one would visualize the synthetic data (e.g., Fig. \ref{fig:T13}(c)) before retraining the model. After the retrained model is tested, we would visualize the testing results (e.g., Fig. \ref{fig:T13}(d,e)). We concluded that the VIS techniques should ideally be available in almost every stage of the workflow.

As illustrated in Fig. \ref{fig:Workflows}, we divided the workflow into four major stages, (A) data synthesis, (B) data selection and integration, (C) model training and testing, and (D) results evaluation. Although an ML developer is most likely to commence at stage (B) for selecting the given training and testing data without any synthetic data, after the first iteration involving (B $\rightarrow$ C $\rightarrow$ D), the ML developer will follow the sequence of
(A $\rightarrow$ B $\rightarrow$ C $\rightarrow$ D)
in most of the subsequent iterations, which typically take more than a few days.

We also notice that the high-level visualization tasks in these four stages are quite different, though some VIS techniques may be required by multiple stages. For example, RBF heatmaps are useful at stage (A) for enabling a user to determine a dividing line as in Fig. \ref{fig:T13}(b), and they are also useful at Stage (D) for comparing the results before and after retraining, e.g., comparing Fig. \ref{fig:T13}(b) and (e). We therefore identify a set of VIS techniques for each major stage and make these VIS techniques available through a user interface that is designed to suit the tasks at each stage. This results in a design of the VIS4ML tool with four ``views'' corresponding to the four major stages of the workflow. We name this VIS4ML tool as \emph{iGAiVA}, which stands for \emph{Integrated Generative AI and Visual Analytics}. Below we describe the functions of each view of iGAiVA.

%
\vspace{1mm}
\noindent\textbf{Synthesis View for Data Synthesis.}
In iGAiVA, this view plays a distinct role in supporting an ML workflow involving data synthesis, while the ML tasks supported by the other three views are comparatively common in conventional ML workflows. 
At this stage, the primary task of an ML developer is to identify a set of example messages that can be used as inputs to LLMs for generating synthetic data. As demonstrated in Sections \ref{sec:VIS} and \ref{sec:LLM}, this task can benefit from VIS techniques extensively, with which the ML developer may use a t-SNE scatter plot to select a class to work on, use PCA scatter plots to observe patterns in different combinations of PCA dimensions; use RBF heatmaps to visualize subareas in a more quantitative and predictive manner; use tag-treemaps to compare the keywords statistics of different subareas; and use PCA scatter plots or RBF heatmaps to determine a subarea for selecting examples (from the training data) as inputs to LLMs.

As there are usually a few dozen of the PCA dimensions to be considered, the ML developer is expected to spend a fair amount of time performing interactive visualization. We, therefore, designed this view as a $2 \times 3$ matrix as shown in Fig. \ref{fig:iGAiVA}(a). The interaction panel above the $2 \times 3$ matrix is used to control what data to be visualized, such as the whole dataset or a class; which class; how a class is subdivided, previous testing results, training data, or synthesized data; and so on. 

The interaction panel on the left of the $2 \times 3$ matrix is used to select what VIS techniques to use and set the corresponding parameters. Two VIS techniques can be used sumptuously and selected from a set of visual representations, including t-SNE, PCA scatter plots, RBF heatmap, tag-treemap, and a number of simple plots as well as viewing text messages directly.
In this way, the ML developer can compare up to three pieces of data using two types of VIS techniques at any moment. There are interaction widgets for setting parameters of different VIS techniques, and commands for refreshing two rows of visualization. 

There are additional commands and pop-up windows for controlling the data synthesis process, e.g., for randomly selecting examples under specific constraints (e.g., regions defined by PCA dimensions), loading data from the file system, selecting data from the iGAiVA cache area, setting parameters for LLMs, saving generated data to the iGAiVA cache area (to be accessed by all views) and to the file system, and deleting synthesized data that is no longer needed.   

%
\vspace{1mm}
\noindent\textbf{Data View for Data Selection and Integration.}
The primary task at this stage is to have the data ready for training an ML model or retraining the model after adding synthesized data. After the first iteration to train an ML model using only the collected real-world data (referred to as the main dataset), there are usually many iterations for retraining. Normally one or more new synthetic datasets are generated for a specific class in each iteration. After a number of iterations, there are many synthetic datasets. iGAiVA maintains an internal cache area for storing about 20 datasets. (The actual number depends on the system parameter for the cache area as well as the sizes of the main dataset and the synthetic datasets.) The Data View provides commands and pop-up windows for managing the cache area, e.g., loading (saving) a dataset from (to) the file system, deleting a dataset from the cache, and merging datasets in the cache.

An ML developer can select one or more synthetic datasets from the cache area, and place them on the small tiles on the right of the screen as shown in Fig.~\ref{fig:iGAiVA}(b). These selected datasets are combined with the main dataset (usually the original training data) and are visualized in the large canvas in the middle.
For example, in Fig.~\ref{fig:iGAiVA}(b), the smaller tag-treemap on the top-right shows the keyword statistics of the original training data in a class (i.e., the main dataset). The two smaller tag-treemaps below show the keyword statistics of two synthetic datasets. The larger tag-treemap in the middle shows the keyword statistics when three datasets are combined. In this way, the ML developer can assess the impact of the synthetic data, e.g., to observe whether the level of skewness increases or decreases.

The VIS techniques commonly used in the Data View are PCA scatter plot, RBF heatmap, tag-treemap, and bar charts for keyword statistics. There are tiles for all selected datasets, including the main dataset. The same VIS technique is applied to the main canvas and all tiles to facilitate consistent comparison. The two interaction panels on the left are for controlling data selection and integration, selecting a VIS technique, and setting its parameters.

\vspace{1mm}
\noindent\textbf{Model View for Model Training and Testing.}
The primary tasks to be performed in the Model View are (i) train or retrain a model and (ii) test the model using a dataset selected or multiple datasets integrated with the Data View. In the Model View, the testing data is normally loaded from the file system, as we normally use the same testing dataset consistently for multiple iterations. There are VIS techniques for monitoring the progress of training and testing.

Usually, training and testing may take some time, and the Model View is designed to focus on unhurried monitoring rather than intensive analysis.  
As shown in Fig. \ref{fig:iGAiVA}(c), the VIS techniques used include a line chart for progress monitoring, tree visualization for decision trees or random forest models, and log data display. The Model View supports the monitoring of intermediate testing results, and there are numerical and visual displays showing performance metrics, e.g., accuracy and confusion matrix.
It was intentional for not providing more complicated VIS techniques to examine the testing results in detail. As the main tasks in Model View are machine-centric, while the tasks of results analysis and evaluation are human-centric, we purposely designed the Results View for the human-centric tasks.      

%
\vspace{1mm}
\noindent\textbf{Results View for Results Analysis and Evaluation.}
The primary task at this stage is to analyze the testing results from the Model View in order to make some high-level decisions about the next iteration, e.g., go to the Synthesis View to generate another synthetic dataset or go to the Data View to configure a different integration. A major part of the analysis is to compare the latest results with those of the previous models. This is one important reason why the Model View cannot support results analysis easily. As shown in Fig. \ref{fig:iGAiVA}(d) it has a similar $2 \times 3$ matrix layout as the Synthesis View. The main difference is that the interaction panel above the $2 \times 3$ matrix is for selecting results data, which is combined with the data objects in the testing data. For example, each data object in the testing data is automatically annotated with correct and incorrect labels, and the testing dataset can be visualized using a PCA scatter plot or an RBF heatmap. There are also commonly used plots and various statistics about the results.

\vspace{2mm}
\noindent\textbf{Using iGAiVA to Conduct Further Experiments.} The four views corresponding to the four major steps in Fig. \ref{fig:Workflows}(c). With iGAiVA, the processes described in Sections \ref{sec:VIS} and \ref{sec:LLM} can be carried out iteratively and systematically. We provide a video in the supplementary material to demonstrate how multiple iterations were used to improve a model, together with an appendix, where we provide further technical details about processes shown in the video.    
\section{User Evaluation}
\label{sec:Evaluation}
\revise{Because of the space constraint, a detailed report on user evaluation can be found in Appendix \ref{apx:Evaluation}. Here we briefly summarize the evaluation process, outcomes, and further actions.}

\vspace{1mm}
\revise{\noindent\textbf{Process.}
In Phase 1 (2 months), the three requirements were identified in multiple steps during the placement of the first author in the FabLab of Inetum, Spain. There were eight in-person meetings (two formal and six informal, 2$\sim$15 people) during this phase, and weekly online meetings involving the third author. The first author wrote two reports and gave two presentations on the VIS and LLM solutions, as part of Phase 1 evaluation.}

\revise{In Phase 2 (4 months), the design and development of iGAiVA was carried out at Oxford, UK. There were weekly in-person meetings and several online meetings involving the second author, who had a 1-month visit to the first and third authors at Oxford. The evaluation was focused on the testing results and the design of iGAiVA.}

\revise{In Phase 3 (2 months), the first author had a further 1-month placement at Inetum, where the iGAiVA tool was evaluated by ML researchers and developers through several meetings (up to 15 people) and individual interviews (5 people). This was followed by efforts for software improvement and technical documentation.}

\vspace{1mm}
\revise{\noindent\textbf{Outcomes.} While the quantitative testing results (Section \ref{sec:LLM} and Appendix \ref{apx:Results}) allowed domain experts to draw a convincing conclusion that the VA+LLM based workflow, i.e., Fig. \ref{fig:Workflows}(b), is effective and efficient, the qualitative evaluation of the iGAiVA enabled domain experts to plan for developing iGAiVA into an industrial software product. The evaluation of three requirements described in Section \ref{sec:Overview} are detailed in Appendix \ref{apx:Evaluation}. }

\vspace{1mm}
\revise{\noindent\textbf{Further Actions.} In addition, the work on iGAiVA inspired further technical development in aspects of machine learning. It is anticipated that the new advancement in machine learning will be integrated into the future development of iGAiVA.}

\section{Conclusions}
\label{sec:Conclusions}
In this paper, we have presented a visual analytics solution for enabling targeted data synthesis in ML workflows. We used VIS techniques that closely coupled analysis and visualization, such as dimensionality reduction and scatter plot (t-SNE), PCA and scatter plot, RBF and heatmap, and keyword statistics coupled with tag cloud and treemap. The code for RBF has been open-sourced in GitHub https://github.com/MattJin19/RBF. The close coupling of VIS and LLMs processes in a workflow represents a high-level integration, demonstrating the successful deployment of VA concepts.

Through this paper, we have demonstrated that with VIS techniques, it has become feasible to target data synthesis to ``areas'', which ML developers can observe visually and can use their knowledge to reason about the possible causes of poor performance and hypothesize potential improvement if synthetic data were introduced in these ``areas''. We have demonstrated that discovering such areas can effectively lead to the improvement of the ML models concerned.

To our best knowledge, this approach has not been reported previously in the literature. We are continuing this work mainly in the form of developing a piece of industrial software based on this approach. Meanwhile, we anticipate that there must be many other visual patterns that could suggest some types of causes of poor performance, and there must be some other iterative pathways from improving the performance of a single class to improving the overall performance. We believe that the more ML developers can use VIS techniques, the more effective pathways will be discovered.

Further discussions can be found in Appendix~\ref{apx:Discussions}.

\newpage
\acknowledgments{
This work has been made possible by the Network of European
Data Scientists. We would like to express our gratitude to the people who
facilitated this project, in particular, Dolores Romero Morales from
Copenhagen Business School.}

\bibliographystyle{abbrv-doi}

\bibliography{ref}

\newpage

\clearpage
\pagenumbering{arabic}
\setcounter{page}{1}
\numberwithin{figure}{section}

\appendix

\begin{center}
\large
APPENDICES OF\\[1mm]
\Large\noindent
\textbf{\textsf{iGAiVA: Integrated Generative AI and Visual Analytics in a Machine Learning Workflow for Text Classification}}\\[2mm]
\normalsize
Yuanzhe Jin$^1$, Adri\'{a}n Carrasco-Revilla$^2$, and Min Chen$^1$\\[1mm]
$^1 $University of Oxford, UK and $^2 $Inetum, Spain
\normalsize
\end{center}

This arXiv report is accompanied by several appendices. The first five appendices provide further technical details for supporting the main body of the paper:
\begin{itemize}
    \item An unnumbered appendix for listing common abbreviations used in the paper.
    \item[A] Further testing results obtained in a multi-iteration workflow supported by iGAiVA.
    \item[B] Further reporting on user evaluation. The second version of the paper was submitted to the conference track of PacificVis 2025, which has a lower page limit than its journal track, where the first version of this paper was submitted to. Because of the lower page limit in the main body, the first version of Section 7 is now included in this appendix, while the new version of Section 7 provides a relatively brief summary.
    \item[C] This appendix provides further discussions on several topics, as requested by the reviewers of the PacificVis 2025 TVCG track.
    \item[D] This appendix provides high-resolution images, as the same images were included in the main body of the paper as smaller images.
\end{itemize}

Version 1 of arXiv:2409.15848 is more or less the manuscript submitted to PacificVis 2025 Journal Track. We were disappointed that the paper was not accepted nor recommended for a fast-track to TVCG, though we were aware that the journal track was highly competitive. The reviewers requested many changes. Some requests, such as formative user evaluation, formal user study, and novel visual design, were commonly-used to reject visualization papers. Recently, IEEE VIS 2024 hosted a discussion panel on ``\emph{(Yet Another) Evaluation Needed? A Panel Discussion on Evaluation Trends in Visualization}''. Senior VIS scientists on the panel critiqued such excessive requests.

As we were aware that the excessive requests for user-centered evaluation had been common, we decided to submit the paper to the conference track of PacificVis 2025, as reviews in the conference track were expected to demand less. Meanwhile, we addressed the reviewers' other requests through revision, including revision to the main body of the paper as well as adding Appendices C and D. The paper in the main body of this version of the arXiv report and five appendices (Common Abbreviations and A$\sim$D) were submitted to the conference track of PacificVis 2025, together with a revision report (Appendix \ref{apx:RevisionReport}). 

Unexpectedly, the same group of reviewers did not remove the original requests for formative user evaluation, formal user study, and novel visual design, and added further requests (e.g., generalizability) while overlooking multiple pieces of text in the paper and mistakenly assuming that the testing data was used to generate synthetic data. In Version 2 of this arXiv report, we also include four additional appendices to help improve the transparency, integrity, professionalism, and fairness in academic review processes.         

\begin{itemize}
    \item[E] Feedback letter sent to the co-chair of PacificVis 2025 Conference track (on January 4, 2025).
    \item[F] Reviews of PacificVis 2025 Conference track (received on December 24, 2024).
    \item[G] Revision Report submitted to PacificVis 2025 Conference Track (on November 20, 2024).
    \item[H] Reviews of PacificVis 2025 Journal Track (received on November 7, 2024). 
\end{itemize}

\section*{\revise{Common Abbreviations}}
\label{apx:Abbreviation}
\revise{In this unnumbered appendix, we provide a list of abbreviations, ordered alphabetically, which are used in this paper. Most of these abbreviations are commonly used in the VIS community and in related subjects such as AI and machine learning.
\begin{itemize}
    \item \textbf{AI} -- Artificial Intelligence.
    \item \textbf{CNN} -- Convolutional Neural Network.
    \item \textbf{GAI} -- Generative AI. 
    \item \textbf{LLM} - Large Language Model. LLMs play a significant part in many text-based GAI systems, such as ChatGPT.
    \item \textbf{ML} -- Machine Learning.
    \item \textbf{PCA} -- Principal Component Analysis.
    \item \textbf{RBF} -- Radial Basis Function.
    \item \textbf{t-SNE} -- t-distributed Stochastic Neighbor Embedding.
    \item \textbf{VA} -- Visual Analytics, which is a branch of VIS. A typical VA approach, method, technique, tool, or workflow involves the combined uses of human-centric processes (e.g., visualization and human-computer interaction) and machine-centric processes (e.g., statistics, algorithms, and ML models)  
    \item \textbf{VIS} -- Visualization and Visual Analytics. The abbreviation was used by the VIS community to encompass all VIS research. Before the three conferences were merged, VIS stands for VA (or VAST), InfoVis, and SciVis. 
    \item \textbf{VIS4ML} -- Visualization and Visual Analytics for Machine Learning. The abbreviation was first introduced by Sacha et al. \cite{Sacha:2018:TVCG}.
\end{itemize}
}


\section{Further Testing Results Obtained in a Multi-Iteration Workflow Supported by iGAiVA}
\label{apx:Results}

In Section \ref{sec:VIS}, we explained the VIS-assisted processes for testing different synthetic data in order to improve an existing text classification model M0. Table~\ref{tab:Synthetic} provides some of the testing results. In this appendix, we provide more testing results, including those produced by further iterations in the workflow. 

\begin{table*}[t]
\centering
\caption{Further results from the first iteration: retraining with different class-based synthetic data. The $\Delta$-recall results were compared with the original model M0 (Major Column 2).}
\label{tab:synthetic-1+}
\centering

\begin{tabular}{|c|ccc|c@{\hspace{3.5mm}}c|c@{\hspace{3.5mm}}c|%
c@{\hspace{3.5mm}}c|c@{\hspace{3.5mm}}c|c@{\hspace{3.5mm}}c|c@{\hspace{3.5mm}}c}

\textbf{Topic} & \multicolumn{3}{c|}{\textbf{Original Data}}     & \multicolumn{2}{c|}{\textbf{T11-s2}}     & \multicolumn{2}{c|}{\textbf{T12-s2}}     & \multicolumn{2}{c|}{\textbf{T13-s2}}     & \multicolumn{2}{c|}{\textbf{T14-s2}}     & \multicolumn{2}{c|}{\textbf{T15-s2}}     \\
\textbf{class}     & \textbf{test} & \textbf{train} & \textbf{recall}     & \textbf{train} & $\Delta$-\textbf{recall}     & \textbf{train} & $\Delta$-\textbf{recall}     & \textbf{train} & $\Delta$-\textbf{recall}     & \textbf{train} & $\Delta$-\textbf{recall}     & \textbf{train} & $\Delta$-\textbf{recall}     \\
\hline
Overall & 7820 & 31280 & 0.821 &  & 0.000 &  & \up 0.008 &  & 0.000  &  & \down 0.002 &  & \up 0.001\\
\hline
T1  & 1748 & 6781 & 0.976 &  & \down 0.001 &  & \up 0.003 &  & \up 0.003  &  & \up 0.003 &  & \up 0.004\\
T2  & 2259 & 9091 & 0.943 &  & \up 0.004 &  & \up 0.002 &  & \down 0.002  &  & \down 0.010 &  & \down 0.008\\
T3  & 927 & 3792 & 0.892 &  & \down 0.037 &  & \down 0.024 &  & \down 0.005 &  & \down 0.014 &  & \down 0.001\\
T4  & 281 & 1106 & 0.769 &  & \up 0.042 &  & \up 0.046 &  & 0.000  &  & \down 0.015&  & \up 0.035\\
T5  & 533 & 2222 & 0.732 &  & \up 0.017 &  & \up 0.028 &  & \down 0.023  &  & \up 0.015 &  & \up 0.002\\
T6  & 373 & 1515 & 0.528 &  & \up 0.022 &  & \up 0.067 &  & \up 0.008  &  & \up 0.040 &  & \up 0.008\\
T7  & 395 & 1568 & 0.714 &  & \up 0.030 &  & \up 0.033 &  & \up 0.013  &  & \down 0.010 &  & \up 0.015\\
T8  & 236 & 792 & 0.508 &  & \down 0.055 &  & \down 0.012 &  & \up 0.034 &  & \down 0.008 &  & \down 0.072\\
T9  & 278 & 1188 & 0.626 &  & \up 0.025 &  & \up 0.050 &  & \down 0.014  &  & \down 0.061 &  & \up 0.039\\
T10 & 335 & 1364 & 0.427 &  & \down 0.030 &  & \down 0.009 &  & \up 0.021  &  & \up 0.027 &  & \up 0.009\\
T11 & 94 & 377 & 0.926 & +110 & \up 0.021 &  & 0.000 &  & 0.000  &  & \up 0.021 &  & \up 0.010\\
T12 & 64 & 294 & 0.375 &  & \down 0.016 & +325 & \up 0.125 &  & \up 0.016 &  & \down 0.016 &  & \down 0.047\\
T13 & 45 & 135 & 0.178 &  & \up 0.066 &  & \up 0.044 & +525 & \up 0.022 &  & \up 0.066 &  & \up 0.022\\
T14 & 135 & 629 & 0.526 &  & \down 0.015 &  & \down 0.022 &  & \down 0.030  & +1265 & \up 0.111 &  & \down 0.022\\
T15 & 117 & 426 & 0.376 &  & \down 0.026 &  & \down 0.026 &  & \down 0.026  &  & \down 0.043 & +1485 & \up 0.086\\
\hline
\end{tabular}
\end{table*}

\begin{table*}
\centering
\caption{Results of the second iteration: retraining with different class-based synthetic data in addition to \textbf{T11-s1} synthetic data. The $\Delta$-recall results were compared with the original model M0 (Major Column 2), while the M1a results are given in Major Column 3.}
\label{tab:synthetic-T11-2a}
\centering

\begin{tabular}{|c|ccc|c@{\hspace{4mm}}c|c@{\hspace{4mm}}c|%
c@{\hspace{4mm}}c|c@{\hspace{4mm}}c|c@{\hspace{4mm}}c|c@{\hspace{4mm}}c}

\textbf{Topic} & \multicolumn{3}{c|}{\textbf{Original Data}} & \multicolumn{2}{c|}{\textbf{T11-s1}} & \multicolumn{2}{c|}{\textbf{T11-s1+T12-s1}} & \multicolumn{2}{c|}{\textbf{T11-s1+T12-s2}} & \multicolumn{2}{c|}{\textbf{T11-s1+T13-s1}} & \multicolumn{2}{c|}{\textbf{T11-s1+T13-s2}} \\
\textbf{class} & \textbf{test} & \textbf{train} & \textbf{recall} & \textbf{train} & $\Delta$-\textbf{recall} & \textbf{train} & $\Delta$-\textbf{recall} & \textbf{train} & $\Delta$-\textbf{recall} & \textbf{train} & $\Delta$-\textbf{recall} & \textbf{train} & $\Delta$-\textbf{recall} \\
\hline
Overall & 7820 & 31280 & 0.821 &  & \up 0.003 &  & \up 0.004 &  & \down 0.003 &  & \down 0.004 &  & \down 0.003 \\
\hline
T1  & 1748 & 6781 & 0.976 &  & \down 0.002 &  & \down 0.006 &  & \down 0.012 &  & \up 0.003 &  & 0.000 \\
T2  & 2259 & 9091 & 0.943 &  & \up 0.002 &  & \up 0.004 &  & \down 0.002 &  & \down 0.009 &  & \down 0.005 \\
T3  & 927  & 3792  & 0.892 &  & \down 0.009 &  & \down 0.006 &  & \down 0.014 &  & \down 0.025 &  & \down 0.012 \\
T4  & 281  & 1106  & 0.769 &  & \up 0.039 &  & \up 0.032 &  & \up 0.014 &  & \up 0.025 &  & \up 0.017 \\
T5  & 533  & 2222  & 0.732 &  & \up 0.020 &  & \up 0.005 &  & \down 0.023 &  & \up 0.013 &  & \down 0.002 \\
T6  & 373  & 1515  & 0.528 &  & \up 0.030 &  & \up 0.024 &  & \up 0.046 &  & \up 0.011 &  & \up 0.022 \\
T7  & 395  & 1568  & 0.714 &  & \up 0.013 &  & \up 0.033 &  & \up 0.025 &  & \down 0.003 &  & \up 0.015 \\
T8  & 236  & 792   & 0.508 &  & \down 0.055 &  & \up 0.056 &  & \up 0.047 &  & \down 0.029 &  & \down 0.059 \\
T9  & 278  & 1188  & 0.626 &  & \up 0.011 &  & \up 0.007 &  & \down 0.047 &  & \up 0.007 &  & \down 0.011 \\
T10 & 335  & 1364  & 0.427 &  & \up 0.006 &  & \down 0.012 &  & \up 0.018 &  & \up 0.036 &  & \up 0.027 \\
T11 & 94   & 377   & 0.926 & +110 & \up 0.021 & +110 & \up 0.031 & +110 & \up 0.021 & +110 & \up 0.010 & +110 & \up 0.010 \\
T12 & 64   & 294   & 0.375 &  & 0.000 & +615 & \up 0.047 & +615 & \down 0.016 &  & \down 0.063 &  & \down 0.078 \\
T13 & 45   & 135   & 0.178 &  & \up 0.155 &  & \up 0.066 &  & \down 0.022 & +585 & \up 0.089 & +585 & \up 0.111 \\
T14 & 135  & 629   & 0.526 &  & \down 0.037 &  & \down 0.104 &  & \down 0.030 &  & \down 0.045 &  & \down 0.007 \\
T15 & 117  & 426   & 0.376 &  & \down 0.034 &  & \down 0.008 &  & \down 0.026 &  & \down 0.077 &  & \down 0.077 \\
\hline
\end{tabular}
\end{table*}

\vspace{2mm}
\textbf{Iteration 1 -- Tables~\ref{tab:Synthetic} (main body) and \ref{tab:synthetic-1+} (this appendix):}
\begin{itemize}
    \item \textbf{T11-s1}: Results in Table~\ref{tab:Synthetic}, Major Column 3.\\
        $\rightarrow$ T11 recall: +0.021, overall recall: +0.003.
    \item \textbf{T12-s1}: Results in Table~\ref{tab:Synthetic}, Major Column 4.\\
        $\rightarrow$ T12 recall: +0.031, overall recall: +0.009.
    \item \textbf{T13-s1}: Results in Table~\ref{tab:Synthetic}, Major Column 5.\\
        $\rightarrow$ T13 recall: +0.133, overall recall: +0.009.
    \item \textbf{T14-s1}: Results in Table~\ref{tab:Synthetic}, Major Column 6.\\
        $\rightarrow$ T14 recall: +0.148, overall recall: +0.001.
    \item \textbf{T15-s1}: Results in Table~\ref{tab:Synthetic}, Major Column 7.\\
        $\rightarrow$ T15 recall: +0.094, overall recall: +0.004.
    \item \textbf{T11-s2}: Results in Table~\ref{tab:synthetic-1+}, Major Column 3.\\
        $\rightarrow$ T11 recall: +0.021, overall recall: $\sim$0.000.
    \item \textbf{T12-s2}: Results in Table~\ref{tab:synthetic-1+}, Major Column 4.\\
        $\rightarrow$ T12 recall: +0.125, overall recall: +0.008.
    \item \textbf{T13-s2}: Results in Table~\ref{tab:synthetic-1+}, Major Column 5.\\
        $\rightarrow$ T13 recall: +0.022, overall recall: $\sim$0.000.
    \item \textbf{T14-s2}: Results in Table~\ref{tab:synthetic-1+}, Major Column 6.\\
        $\rightarrow$ T14 recall: +0.111, overall recall: $-0.002$.
    \item \textbf{T15-s2}: Results in Table~\ref{tab:synthetic-1+}, Major Column 7.\\
        $\rightarrow$ T15 recall: +0.086, overall recall: +0.001.
        
\end{itemize}

\vspace{2mm}
\textbf{Iteration 2A (based on \textbf{T11-s1}) -- Table~\ref{tab:synthetic-T11-2a}:}
\begin{itemize}
    \item \textbf{T11-s1 + T12-s1}: Results in Table~\ref{tab:synthetic-T11-2a}, Major Column 4.\\
    $\rightarrow$ T12 recall: +0.047, overall recall: $+0.004$.\\
    $\rightarrow$ Comparing with M1a (T11-s1) overall recall: $+0.001$.
    \item \textbf{T11-s1 + T12-s2}: Results in Table~\ref{tab:synthetic-T11-2a}, Major Column 5.\\
    $\rightarrow$ T12 recall: $-0.016$, overall recall: $-0.003$.\\
    $\rightarrow$ Comparing with M1a (T11-s1) overall recall: $-0.006$.
    \item \textbf{T11-s1 + T13-s1}: Results in Table~\ref{tab:synthetic-T11-2a}, Major Column 6.\\
    $\rightarrow$ T13 recall: +0.0.089, overall recall: $-0.004$.\\
    $\rightarrow$ Comparing with M1a (T11-s1) overall recall: $-0.007$.
    \item \textbf{T11-s1 + T13-s2}: Results in Table~\ref{tab:synthetic-T11-2a}, Major Column 7.\\
    $\rightarrow$ T13 recall: +0.111, overall recall: $-0.003$.\\
    $\rightarrow$ Comparing with M1a (T11-s1) overall recall: $-0.006$.
\end{itemize}

\vspace{2mm}
\textbf{Iteration 2B (based on \textbf{T13-s1}) -- Table~\ref{tab:synthetic-T11-2b}:}
\begin{itemize}
    \item \textbf{T13-s1 + T11-s1}: Results in Table~\ref{tab:synthetic-T11-2b}, Major Column 4.\\
    $\rightarrow$ T11 recall: +0.021, overall recall: $\sim 0.000$.\\
    $\rightarrow$ Comparing with M1b (T13-s1) overall recall: $-0.009$.
    \item \textbf{T13-s1 + T11-s2}: Results in Table~\ref{tab:synthetic-T11-2b}, Major Column 5.\\
    $\rightarrow$ T11 recall: +0.021, overall recall: $+0.004$.\\
    $\rightarrow$ Comparing with M1b (T13-s1) overall recall: $-0.005$.
    \item \textbf{T13-s1 + T12-s1}: Results in Table~\ref{tab:synthetic-T11-2b}, Major Column 6.\\
    $\rightarrow$ T12 recall: +0.031, overall recall: $+0.004$.\\
    $\rightarrow$ Comparing with M1b (T13-s1) overall recall: $-0.005$.
    \item \textbf{T13-s1 + T12-s2}: Results in Table~\ref{tab:synthetic-T11-2b}, Major Column 7.\\
    $\rightarrow$ T12 recall: $\sim 0.000$, overall recall: $\sim 0.000$.\\
    $\rightarrow$ Comparing with M1b (T13-s1) overall recall: $-0.009$.
\end{itemize}

\begin{table*}[t]
\centering
\caption{Results of the second iteration: retraining with different class-based synthetic data in addition to \textbf{T13-s1} synthetic data. The $\Delta$-recall results were compared with the original model M0 (Major Column 2), while the M1b results are given in Major Column 3.}
\label{tab:synthetic-T11-2b}
\centering

\begin{tabular}{|c|ccc|c@{\hspace{4mm}}c|c@{\hspace{4mm}}c|%
c@{\hspace{4mm}}c|c@{\hspace{4mm}}c|c@{\hspace{4mm}}c|c@{\hspace{4mm}}c}

\textbf{Topic} & \multicolumn{3}{c|}{\textbf{Original Data}} & \multicolumn{2}{c|}{\textbf{T13-s1}} & \multicolumn{2}{c|}{\textbf{T13-s1+T11-s1}} & \multicolumn{2}{c|}{\textbf{T13-s1+T11-s2}} & \multicolumn{2}{c|}{\textbf{T13-s1+T12-s1}} & \multicolumn{2}{c|}{\textbf{T13-s1+T12-s2}} \\
\textbf{class} & \textbf{test} & \textbf{train} & \textbf{recall} & \textbf{train} & $\Delta$-\textbf{recall} & \textbf{train} & $\Delta$-\textbf{recall} & \textbf{train} & $\Delta$-\textbf{recall} & \textbf{train} & $\Delta$-\textbf{recall} & \textbf{train} & $\Delta$-\textbf{recall} \\
\hline
Overall & 7820 & 31280 & 0.821 &  & \up 0.009 &  & 0.000 &  & \up 0.004 &  & \up 0.004 &  & 0.000  \\
\hline
T1  & 1748 & 6781 & 0.976 &  & \down 0.001 &  & \up 0.006 &  & \down 0.007 &  & \up 0.003 &  & \up 0.001 \\
T2  & 2259 & 9091 & 0.943 &  & \down 0.003 &  & \up 0.002 &  & \up 0.005 &  & \down 0.006 &  & \down 0.011  \\
T3  & 927  & 3792  & 0.892 &  & \down 0.004 &  & \down 0.013 &  & \down 0.004&  & \down 0.003 &  & \down 0.011 \\
T4  & 281  & 1106  & 0.769 &  & \up 0.028 &  & \up 0.039 &  & \up 0.053 &  & \up 0.039 &  & \up 0.025 \\
T5  & 533  & 2222  & 0.732 &  & \up 0.009 &  & \down 0.008 &  & \up 0.002 &  & \up 0.032 &  & \up 0.020\\
T6  & 373  & 1515  & 0.528 &  & \up 0.038 &  & \up 0.011 &  & \up 0.003 &  & \up 0.032 &  & \up 0.003\\
T7  & 395  & 1568  & 0.714 &  & \up 0.025 &  & \up 0.033 &  & \up 0.081 &  & \down 0.028 &  & \up 0.008 \\
T8  & 236  & 792   & 0.508 &  & \up 0.077 &  & \down 0.084 &  & \down 0.025 &  & \down 0.004 &  & \down 0.004 \\
T9  & 278  & 1188  & 0.626 &  & \down 0.018 &  & \up 0.018 &  & \down 0.004 &  & \down 0.018 &  & \down 0.014 \\
T10 & 335  & 1364  & 0.427 &  & \up 0.095 &  & \down 0.018 &  & \down 0.006 &  & \up 0.074 &  & \up 0.066 \\
T11 & 94   & 377   & 0.926 &  & 0.000 & +120 & \up 0.021 & +120 & \up 0.021 &  & \up 0.010 &  & \up 0.010 \\
T12 & 64   & 294   & 0.375 &  & \down 0.016 &  & 0.000 &  & 0.000 & +685 & \up 0.031 & +685 & 0.000 \\
T13 & 45   & 135   & 0.178 & +525 & \up 0.133 & +525 & \up 0.044 & +525 & \up 0.089 & +525 & \up 0.133 & +525 & \up 0.178 \\
T14 & 135  & 629   & 0.526 &  & \down 0.022 &  & \down 0.022 &  & \down 0.045 &  & \down 0.022 &  & 0.000 \\
T15 & 117  & 426   & 0.376 &  & \down 0.034 &  & \down 0.034 &  & \down 0.043 &  & \down 0.051 &  & \down 0.094 \\
\hline
\end{tabular}
\end{table*}

\begin{table*}
\caption{Results of the third iteration: retraining with different class-based synthetic data in addition to \textbf{T11-s1} and \textbf{T12-s1} synthetic data. The $\Delta$-recall results were compared with the original model M0 (Major Column 2).}
\label{tab:synthetic-T11-T12-3}
\centering

\begin{tabular}{|c|ccc|cc|cc|cc|cc|}
\textbf{Topic} & \multicolumn{3}{c|}{\textbf{Original Data}} & \multicolumn{2}{c|}{\textbf{T11-T12-s1+T13-s1}} & \multicolumn{2}{c|}{\textbf{T11-T12-s1+T13-s2}} & \multicolumn{2}{c|}{\textbf{T11-T12-s1+T14-s1}} & \multicolumn{2}{c|}{\textbf{T11-T12-s1+T14-s2}} \\
\textbf{class} & \textbf{test} & \textbf{train} & \textbf{recall} & \textbf{train} & $\Delta$-\textbf{recall} & \textbf{train} & $\Delta$-\textbf{recall} & \textbf{train} & $\Delta$-\textbf{recall} & \textbf{train} & $\Delta$-\textbf{recall} \\
\hline
Overall & 7820 & 31280 & 0.821 &  & 0.000 &  & \down 0.007 &  & \down 0.004 &  & \down 0.008 \\
\hline
T1  & 1748 & 6781 & 0.976 &  & \down 0.001 &  & \up 0.005 &  & \up 0.003 &  & \down 0.015 \\
T2  & 2259 & 9091 & 0.943 &  & \down 0.005 &  & \down 0.011 &  & \down 0.008 &  & \down 0.006 \\
T3  & 927  & 3792 & 0.892 &  & \down 0.009 &  & \down 0.023 &  & \down 0.015 &  & \down 0.018 \\
T4  & 281  & 1106 & 0.769 &  & \up 0.039 &  & \up 0.032 &  & \up 0.028 &  & \up 0.039 \\
T5  & 533  & 2222 & 0.732 &  & \up 0.020 &  & \down 0.006 &  & \up 0.030 &  & \up 0.013 \\
T6  & 373  & 1515 & 0.528 &  & 0.000 &  & \down 0.008 &  & \up 0.040 &  & 0.000 \\
T7  & 395  & 1568 & 0.714 &  & \down 0.008 &  & \up 0.010 &  & \down 0.015 &  & \down 0.005 \\
T8  & 236  & 792  & 0.508 &  & \down 0.042 &  & \down 0.105 &  & \down 0.127 &  & \down 0.131 \\
T9  & 278  & 1188 & 0.626 &  & \up 0.011 &  & \up 0.011 &  & \down 0.014 &  & \down 0.047 \\
T10 & 335  & 1364 & 0.427 &  & \up 0.009 &  & \up 0.024 &  & \down 0.003 &  & \up 0.033 \\
T11 & 94   & 377  & 0.926 & +110 & \up 0.010 & +110 & \up 0.010 & +110 & \up 0.021 & +110 & \up 0.010 \\
T12 & 64   & 294  & 0.375 & +615 & \up 0.094 & +615 & 0.000 & +615 & 0.000 & +615 & \down 0.031 \\
T13 & 45   & 135  & 0.178 & +550 & \up 0.200 & +550 & \up 0.155 &  & \down 0.045 &  & \down 0.022 \\
T14 & 135  & 629  & 0.526 &  & \down 0.037 &  & \down 0.074 & +595 & \up 0.030 & +595 & \up 0.089 \\
T15 & 117  & 426  & 0.376 &  & \down 0.034 &  & \down 0.077 &  & \down 0.051 &  & \down 0.008 \\
\hline
\end{tabular}
\end{table*}

\vspace{2mm}
\textbf{Iteration 3 (based on \textbf{T11-s1} + \textbf{T12-s1}) -- Table~\ref{tab:synthetic-T11-T12-3}:}
\begin{itemize}
    \item \textbf{T11-s1 + T12-s1 + T13-s1}: Results in Table~\ref{tab:synthetic-T11-T12-3}, Major Column 3.\\
    $\rightarrow$ T13 recall: $+0.200$, overall recall: $\sim 0.000$\\
    $\rightarrow$ Comparing with M2 (\textbf{T11-s1}+\textbf{T12-s1}) overall recall: $-0.004$.
    \item \textbf{T11-s1 + T12-s1 + T13-s2}: Results in Table~\ref{tab:synthetic-T11-T12-3}, Major Column 4.\\
    $\rightarrow$ T13 recall: $+0.155$, overall recall: $-0.007$\\
    $\rightarrow$ Comparing with M2 (\textbf{T11-s1}+\textbf{T12-s1}) overall recall: $-0.011$.
    \item \textbf{T11-s1 + T12-s1 + T14-s1}: Results in Table~\ref{tab:synthetic-T11-T12-3}, Major Column 5.\\
    $\rightarrow$ T14 recall: $+0.030$, overall recall: $-0.004$\\
    $\rightarrow$ Comparing with M2 (\textbf{T11-s1}+\textbf{T12-s1}) overall recall: $-0.008$.
    \item \textbf{T11-s1 + T12-s1 + T14-s2}: Results in Table~\ref{tab:synthetic-T11-T12-3}, Major Column 6.
    $\rightarrow$ T14 recall: $+0.009$, overall recall: $-0.008$\\
    $\rightarrow$ Comparing with M2 (\textbf{T11-s1}+\textbf{T12-s1}) overall recall: $-0.012$.
\end{itemize}

\vspace{2mm}
After three iterations, we can see that if one wishes to improve only the overall recall, in the first iteration, \textbf{T12-s1} and \textbf{T13-s1} managed to deliver an improvement of the overall recall by 0.9\%. This is followed by \textbf{T12-s2} that delivered 0.8\% improvement of the overall recall.

If one wishes to improve some poor-performing classes, we may consider:
\begin{itemize}
    \item For \textbf{T12}:\\
            \textbf{T11-s1} + \textbf{T12-s1} + \textbf{T13-s1} $\rightarrow$ +9.4\%,\\
            \textbf{T11-s1} + \textbf{T12-s1} $\rightarrow$ +3.1\%,\\
            \textbf{T13-s1} + \textbf{T12-s1} $\rightarrow$ +3.1\%
    \item For \textbf{T13}:\\
            \textbf{T11-s1} + \textbf{T12-s1} + \textbf{T13-s1} $\rightarrow$ +20.0\%,\\
            \textbf{T13-s1} + \textbf{T12-s2} $\rightarrow$ +17.8\%,\\
            \textbf{T11-s1} $\rightarrow$ +15.5\%\\
            \textbf{T11-s1} + \textbf{T12-s1} + \textbf{T13-s2} $\rightarrow$ +15.5\%,\\
            \textbf{T13-s1} $\rightarrow$ +13.3\%,\\
            \textbf{T13-s1} + \textbf{T12-s1} $\rightarrow$ +13.3\%
\end{itemize}


\begin{table*}[t]
\centering
\caption{Synthetic data generated from random samples of T11, T12, T13, T14, and T15 class. The $\Delta$-recall results were compared with the original model M0 (Major Column 2).}
\label{tab:random}
\centering

\begin{tabular}{|c|ccc|cc|cc|cc|cc|cc|}
\textbf{Topic} & \multicolumn{3}{c|}{\textbf{Original Data}} & \multicolumn{2}{c|}{\textbf{T11\_random\_s1}} & \multicolumn{2}{c|}{\textbf{T12\_random\_s1}} & \multicolumn{2}{c|}{\textbf{T13\_random\_s1}} & \multicolumn{2}{c|}{\textbf{T14\_random\_s1}} & \multicolumn{2}{c|}{\textbf{T15\_random\_s1}} \\
\textbf{class} & \textbf{test} & \textbf{train} & \textbf{recall} & \textbf{train} & $\Delta$-\textbf{recall} & \textbf{train} & $\Delta$-\textbf{recall} & \textbf{train} & $\Delta$-\textbf{recall} & \textbf{train} & $\Delta$-\textbf{recall} & \textbf{train} & $\Delta$-\textbf{recall} \\
\hline
Overall & 7820 & 31280 & 0.821 &  & \down 0.003 &  & \up 0.002 &  & \up 0.002 &  & \down 0.001 &  & \down 0.001 \\
\hline
T1  & 1748 & 6781 & 0.976 &  & \down 0.005 &  & 0.000 &  & \up 0.003 &  & \down 0.011 &  & \down 0.010 \\
T2  & 2259 & 9091 & 0.943 &  & \down 0.002 &  & \down 0.001 &  & 0.000 &  & \down 0.002 &  & \up 0.002 \\
T3  & 927 & 3792 & 0.892 &  & \down 0.019 &  & \down 0.010 &  & \up 0.001 &  & \down 0.018 &  & \down 0.023 \\
T4  & 281 & 1106 & 0.769 &  & \up 0.042 &  & \up 0.035 &  & \up 0.039 &  & \up 0.003 &  & \up 0.032 \\
T5  & 533 & 2222 & 0.732 &  & \down 0.030 &  & 0.000 &  & \down 0.030 &  & \up 0.003 &  & \down 0.023 \\
T6  & 373 & 1515 & 0.528 &  & \up 0.003 &  & \up 0.024 &  & \up 0.040 &  & \up 0.040 &  & \up 0.032 \\
T7  & 395 & 1568 & 0.714 &  & \up 0.020 &  & \up 0.020 &  & \down 0.015 &  & \up 0.020 &  & \up 0.058 \\
T8  & 236 & 792 & 0.508 &  & \down 0.025 &  & \down 0.012 &  & \down 0.012 &  & \down 0.008 &  & \down 0.055 \\
T9  & 278 & 1188 & 0.626 &  & 0.000 &  & \down 0.040 &  & \up 0.007 &  & \down 0.025 &  & \down 0.007 \\
T10 & 335 & 1364 & 0.427 &  & \up 0.042 &  & \up 0.066 &  & \up 0.042 &  & \up 0.051 &  & \up 0.036 \\
T11 & 94 & 377 & 0.926 & +250 & \up 0.031 &  & 0.000 &  & 0.000 &  & 0.000 &  & 0.000 \\
T12 & 64 & 294 & 0.375 &  & \down 0.031 & +250 & \down 0.016 &  & 0.000 &  & 0.000 &  & \down 0.031 \\
T13 & 45 & 135 & 0.178 &  & \up 0.044 &  & \up 0.155 & +250 & \up 0.066 &  & \up 0.133 &  & \up 0.044 \\
T14 & 135 & 629 & 0.526 &  & \down 0.037 &  & \down 0.007 &  & \down 0.045 & +250 & \up 0.030 &  & \down 0.015 \\
T15 & 117 & 426 & 0.376 &  & \down 0.026 &  & \down 0.077 &  & \down 0.017 &  & \down 0.094 & +250 & \up 0.034 \\
\hline
\end{tabular}
\end{table*}

\begin{table*}[h!]
\centering
\caption{Synthetic data generated from random samples of T5, T8, T9, T14, and a combination of random samples of T11 and T12 class. The $\Delta$-recall results were compared with the original model M0 (Major Column 2).}
\label{tab:random-2}
\centering

\begin{tabular}{|c|c@{\hspace{3mm}}c@{\hspace{3mm}}c|%
c@{\hspace{3mm}}c|c@{\hspace{3mm}}c|c@{\hspace{3mm}}c|%
c@{\hspace{3mm}}c|c@{\hspace{3mm}}c|c@{\hspace{3mm}}c}

\textbf{Topic} & \multicolumn{3}{c|}{\textbf{Original Data}} & \multicolumn{2}{c|}{\textbf{T5\_random\_s1}} & \multicolumn{2}{c|}{\textbf{T8\_random\_s1}} & \multicolumn{2}{c|}{\textbf{T9\_random\_s1}} & \multicolumn{2}{c|}{\textbf{T14\_random\_s2}} & \multicolumn{2}{c|}{\textbf{T11-T12\_random\_s1}} \\
\textbf{class} & \textbf{test} & \textbf{train} & \textbf{recall} & \textbf{train} & $\Delta$-\textbf{recall} & \textbf{train} & $\Delta$-\textbf{recall} & \textbf{train} & $\Delta$-\textbf{recall} & \textbf{train} & $\Delta$-\textbf{recall} & \textbf{train} & $\Delta$-\textbf{recall} \\
\hline
Overall & 7820 & 31280 & 0.821 &  &  \up 0.001 &  &  0.000 &  & \down 0.001 &  & \up 0.003 &  & \down 0.001 \\
\hline
T1  & 1748 & 6781 & 0.976 &  & \up 0.001 &  & \down 0.006 &  & \down 0.005 &  & \up 0.001 &  & 0.000 \\
T2  & 2259 & 9091 & 0.943 &  & \down 0.005 &  & \down 0.004 &  & \down 0.003 &  & \up 0.003 &  & \down 0.005 \\
T3  & 927 & 3792 & 0.892 &  & \down 0.021  &  & \down 0.006 &  & \down 0.012 &  & \down 0.010 &  & \down 0.004 \\
T4  & 281 & 1106 & 0.769 &  & \up 0.039  &  & \up 0.035 &  & \up 0.025 &  & \up 0.053 &  & \up 0.028 \\
T5  & 533 & 2222 & 0.732 & +150  & \up 0.017 & & \up 0.005 &  & \down 0.030 &  & \down 0.017 &  & \down 0.002 \\
T6  & 373 & 1515 & 0.528 &  & \down 0.005 &  & \down 0.013 &  & \up 0.048 &  & \down 0.037 &  & \up 0.024 \\
T7  & 395 & 1568 & 0.714 &  &  \up 0.030  &  & 0.000&  & \up 0.008 &  & \down 0.003 &  & \down 0.013 \\
T8  & 236 & 792 & 0.508 & & \up 0.026 &+400   & \up 0.051 &  & \down 0.038 &  & \up 0.022 &  & \down 0.016 \\
T9  & 278 & 1188 & 0.626 &  & \up 0.050 &  & \up 0.036 & +175 & \up 0.079 &  & \up 0.021 &  & \up 0.047 \\
T10 & 335 & 1364 & 0.427 &  & \down 0.006 &  & \up 0.015 &  & \down 0.021 &  & \up 0.063 &  & \up 0.027 \\
T11 & 94 & 377 & 0.926 &  & \up 0.010 &  & \up 0.010 &  & 0.000 &  & \up 0.010 & +100 & \up 0.010 \\
T12 & 64 & 294 & 0.375 &  & 0.000 &  & \down 0.047 &  & \up 0.016 &  & \down 0.047 & +100 & \down 0.031 \\
T13 & 45 & 135 & 0.178 &  & 0.000 &  & \up 0.044 &  & \up 0.111 &  & \up 0.111 &  & 0.000 \\
T14 & 135 & 629 & 0.526 &  & \down 0.045 &  & \down 0.030 &  & \down 0.007 & +125 & \down 0.015 &  & \down 0.015 \\
T15 & 117 & 426 & 0.376 &  & \down 0.068 &  & \down 0.043 &  & \down 0.068 &  & \down 0.017 &  & \down 0.043 \\
\hline
\end{tabular}
\end{table*}


\noindent\textbf{Randomly-Selected Examples.} We also performed tests to see how the VA-assisted process for selecting examples for LLMs to generate synthetic data has any advantage over a random process for selecting examples. Tables \ref{tab:random} and \ref{tab:random-2} show 10 of such examples.

The ten experiments are:
\begin{itemize}
    \item \textbf{T11-random-s1}: Results in Table~\ref{tab:random}, Major Column 3.\\
        $\rightarrow$ T11 recall: $+0.031$, overall recall: $-0.003$.
    \item \textbf{T12-random-s1}: Results in Table~\ref{tab:random}, Major Column 4.\\
        $\rightarrow$ T12 recall: $-0.016$, overall recall: $+0.002$.
    \item \textbf{T13-random-s1}: Results in Table~\ref{tab:random}, Major Column 5.\\
        $\rightarrow$ T13 recall: $+0.066$, overall recall: $+0.002$.
    \item \textbf{T14-random-s1}: Results in Table~\ref{tab:random}, Major Column 6.\\
        $\rightarrow$ T14 recall: $+0.030$, overall recall: $-0.001$.
    \item \textbf{T15-random-s1}: Results in Table~\ref{tab:random}, Major Column 7.\\
        $\rightarrow$ T15 recall: $+0.034$, overall recall: $-0.001$.
    \item \textbf{T5-random-s1}: Results in Table~\ref{tab:random-2}, Major Column 3.\\
        $\rightarrow$ T5 recall: $+0.017$, overall recall: $+0.001$.
    \item \textbf{T8-random-s1}: Results in Table~\ref{tab:random-2}, Major Column 4.\\
        $\rightarrow$ T8 recall: $+0.051$, overall recall: $\sim0.000$.
    \item \textbf{T9-random-s1}: Results in Table~\ref{tab:random-2}, Major Column 5.\\
        $\rightarrow$ T9 recall: $+0.079$, overall recall: $-0.001$.
    \item \textbf{T14-random-s2}: Results in Table~\ref{tab:random-2}, Major Column 6.\\
        $\rightarrow$ T14 recall: $-0.015$, overall recall: $+0.001$.
    \item \textbf{T11+T12-random-s1}: Results in Table~\ref{tab:random-2}, Major Column 7.\\
        $\rightarrow$ T11 recall: $+0.010$, T12 recall: $-0.031$, overall recall: $-0.001$.
\end{itemize}

For the five experiments in Table \ref{tab:random}, five sets of synthetic data were generated using LLMs for classes \textbf{T11}, \textbf{T12}, \textbf{T13}, \textbf{T14}. and \textbf{T15}.
Each set was created by randomly selecting 50 unique messages as examples from a class and then generating 5 synthetic samples per example. This process resulted in 250 synthetic messages per class. These messages were added to the training data, and the model was retrained. The performance of the retrained model is given in the corresponding column. 

For the first four experiments in Table \ref{tab:random-2}, examples were randomly selected from classes \textbf{T5}, \textbf{T8}, \textbf{T9}, and \textbf{T14} respectively. In these cases, we used different numbers of examples, i.e., 30 for \textbf{T5}, 80 for \textbf{T8}, 35 for \textbf{T9}, and 25 for \textbf{T14}. Again LLMs were used to generate 5 synthetic messages per example.

Lhe last column in Table \ref{tab:random-2} is an experiment where 100 examples were randomly selected from \textbf{T11} and another 100 examples were randomly selected from \textbf{T12}. This can be considered as a two-iteration experiment, with \textbf{T11} first followed by \textbf{T12} or \textbf{T12} first followed by \textbf{T11}.

The experiments reported in this appendix are a subset of all experiments conducted in this work. In general, randomly-selected examples can lead to performance improvement or declination in the range $[-0.3\%, +0.3\%]$ in terms of overall recall. On the other hand, VA-assisted approach for selecting examples has led to performance improvement or declination in the range $[-0.7\%, +0.9\%]$ in terms of overall recall. As shown in Table \ref{tab:Synthetic} in the main body of the paper and tables in this appendix, +0.9\% improvement was achieved multiple times. Meanwhile, $\leq-0.4\%$ declination all occurred in the third iteration (Table \ref{tab:synthetic-T11-T12-3}), when model improvement becomes difficult.

As shown in Tables \ref{tab:random} and \ref{tab:random-2}, for individually-targeted classes, randomly-selected examples can lead to performance improvement or declination in the range $[-3.1\%, +7.9\%]$ in terms of recall. VA-assisted approach for selecting examples has led to performance improvement or declination in the range $[-3.1\%, +17.8\%]$ in terms of recall. Our experience showed that the users can usually make a reasonably good prediction about a positive outcome for the targeted class when selecting examples through the visualization plots as shown in Section \ref{sec:VIS}. The overall performance is less predictable, especially when the targeted area in a PCA-based scatter plot overlaps extensively with other classes (i.e., the gray dots behind).

\newpage

\section{\revise{Further Reporting on User Evaluation}}
\label{apx:Evaluation}
\revise{\emph{Because of the page limit in the main body of the paper, the original version of Section \ref{sec:Evaluation} is included in this appendix, while the new version of Section \ref{sec:Evaluation} provides a relatively brief summary.}}

As described in Section \ref{sec:Overview}, the three requirements were identified in multiple steps during the 2-month placement of the first author in the FabLab of Inetum, Spain. The experimental workflow in Fig. \ref{fig:Workflows}, including the VIS and LLM solutions described in Sections \ref{sec:VIS} and \ref{sec:LLM}, was developed and evaluated during the placement as there were daily contacts between the first author (who is specialized in both VIS and ML) and the ML researchers and developers in the FabLab. There were two formal meetings and six informal meetings during the 2-month period. \revise{There were up to 15 people in the formal meetings.} The first author wrote two reports and gave two presentations on the VIS and LLM solutions in the experimental workflow. In addition, there were weekly online technical meetings, involving the third author, who visited the Inetum, Spain before the placement.

During the development of iGAiVA, the first and third authors had weekly design meetings to analyze the requirements, and evaluate interim designs as well as the prototype under development. There were several online meetings, involving the second author, who paid a 1-month visit to the first and third authors at Oxford.
This was followed by a 1-month placement by the first author at Inetum, Spain, where the iGAiVA tool was evaluated by ML researchers and developers through several meetings \revise{(up to 15 people)}, five interviews, and many informal discussions.

Through these frequent engagements and the development of VIS and LLM solutions, the three requirements emerged gradually. These requirements were evaluated continually when iGAiVA was developed. Below we summarized the major feedback on the technical work presented in Sections \ref{sec:VIS}, \ref{sec:LLM}, and \ref{sec:iGAiVA}, which correspond to the three requirements. All texts in \emph{italic} are from the minutes of meetings and transcripts of conversations. 

\vspace{2mm}
\noindent\textbf{R1: Using more VIS techniques to help identify possible causes of errors.} The initial evaluation was welcoming but uncertain about whether VIS is really useful. The uncertainty was largely removed after VIS was used to help the LLM-based data synthesis process.

$\blacktriangleright$ When t-SNE scatter plots and PCA scatter plots were first used to analyze the training data and testing results, the ML developers were interested but unsure about what information they could gain. They commented: \emph{They} [(i.e., these plots)] \emph{are nice. We saw similar pictures in some research papers and presentations. We have not used these in our workflows} [for developing ML models.] \emph{We like these pictures. Hopefully when you} [(i.e., the first author)] \emph{are here, we can find out how useful they are.}

$\blacktriangleright$ On RBF and tag-treemap, \emph{We have not seen these before in ML. Very interesting. It may take some time for us to learn to use them.}

$\blacktriangleright$ After some experimental results for the requirement \textbf{R2} became available, ML developers are convinced that VIS techniques can be used to solve problems. They commented: \emph{We can now see these plots can show where to chose examples for data generation. Seeing these pictures is a bit like seeing targets in sports.}

$\blacktriangleright$ \emph{Being able to see different parts of a class have different behaviors, we can use a stacked model to help some classification tasks.}

$\blacktriangleright$ \emph{Visualization can be used during the interaction between ML developers and the clients.}

$\blacktriangleright$ \emph{We can use visualization to monitor a model's behavior in a dynamic environment as the whole dataset ``refreshes'' periodically.}

$\blacktriangleright$ \emph{I can see two benefits for the company to receive from visualization methods: cost saving and understanding the data.}

\vspace{2mm}
\noindent\textbf{R2: Using large language models (LLMs) to generate synthetic data for training and using VIS techniques to guide the data synthesis process.} The combined use of VIS and LLM helped improve the ML model for the specific ticketing system in a relatively shorter period and more controllable manner (in comparison with hyper-parameter tuning.) The focus of the evaluation was quickly shifted to the long-term impact of the approach on the business processes.

$\blacktriangleright$ \emph{The current results} [for the dataset used in this work] \emph{are very promising. We can see} [the ML models for] \emph{other ticketing systems can also be improved in this way.}

$\blacktriangleright$ \emph{Clients are sometimes only interested in certain categories} [(i.e., classes),] \emph{it is acceptable when these certain categories improve} [their accuracy,] \emph{while other categories lose some accuracy.}

$\blacktriangleright$ \emph{From a business point of view, targeted data generation will be very useful. I can remember one case,} [where] \emph{a client wanted to have a class with only a few messages} [(i.e., training and testing data)]. \emph{After training, the model only achieved about 10\% accuracy. The client explained the reason: they knew that the messages for this class would increase quickly.}

$\blacktriangleright$ \emph{Other business workflows can also potentially be changed. At the moment, after a ticketing system has been released, the client's feedback} [about the classification model(s)] \emph{is difficult to deal with, because retraining models cannot target individual issues. Now we can. This is wonderful. It will be possible that we can receive feedback regularly from clients and update ML models, possibly monthly.}
\emph{Our software for these ticketing systems is very sophisticated. ML models can be updated easily.}

\vspace{2mm}
\noindent\textbf{R3: Designing and developing a VIS4ML tool for supporting ML workflows involving data synthesis.} The iGAiVA prototype is useful for informing future design and development in the industry. The company will invest more resources to transform it into a web-based prototype suitable as a ``demo'' in the company. The discussion on industrial software engineering is informative to academic researchers. A research prototype is not the same as an industrial prototype.

$\blacktriangleright$ \emph{It is a good idea to build a software tool for supporting ML developers. The current design} [of iGAiVA] \emph{is suitable for a piece of academic work. It can help shape the design of an industrial software tool, which will take many months and will require the company to allocate a fair amount of resources} [for the development].

$\blacktriangleright$ \emph{The design idea of four views is sensible at the moment. However, the software engineering process in a large industrial company is very different from} [an] \emph{academic} [development]. \emph{The next step will be for us} [the company] \emph{to allocate a small amount of resources to transform this} [i.e., iGAiVA] \emph{to a web-based tool. It is still a prototype. We call this ``demo''} [in the company]. \emph{So the ML people in the company across different regions can get their hands on it. In this way, we can use this demo} [i.e., web-based prototype] \emph{in different contexts and receive comprehensive feedback. I am sure that this will happen.}

$\blacktriangleright$ \emph{From the} [comprehensive] \emph{feedback about the web-based prototype, the company will make a long-term decision. It is about how to integrate the new VIS and LLM functions into the ML} [development] \emph{workflows in the company. This could be a standalone tool or we can add these functions into existing tools. It is too early to predict such a decision. But, the combined VIS and LLM solution and the iGAiVA prototype are setting the ball rolling.}

\section{\revise{Further Discussions}}
\label{apx:Discussions}
\revise{\emph{One reviewer of this paper recommended for us to discuss four topics. Due to the 9-page space constraint for the main body of the paper, we include further discussions in this appendix.}}

\vspace{1mm}
\revise{\noindent\textbf{Scalability.}
The term ``scalability'' is defined in relation to a scaling variable $x$. 
Given $n$ messages in a training dataset, there are numerous ways to select $k$ messages. The total number of possible combinations for a fixed $k$ is:
\[
    C\dbinom{n}{k} = \frac{n!}{k! (n-k)!}
\]
In our example, $n = 39,100 \times 80\% = 31,280$ (Table \ref{tab:Dataset}) and $20 < k < 300$ (Table \ref{tab:Synthetic}, 5 synthetic messages per example message). When $k$ is varying from 1 to $n$, the total possible combinations is:
\[
    \sum_{k=1}^n C\dbinom{n}{k}
\]
Among these combinations, there is one or more combinations that can be used to generate the most effective set of synthetic data using LLMs for improving a model. In addition, finding the optimal set of example messages also depends on many other factors, e.g., the parameter settings for LLMs, choices of ML frameworks, design options of the architectures of the text classification models, hyperparameters for training the models, and so on. Hence the search space for an optimal set of messages in the training data is gigantic, and it is not feasible to search the whole space using a brute-force strategy, either manually or using the computer. In other words, when $x = n$, neither humans or computers are scalable to the whole space.}

\revise{In our work, we found that (i) targeting one class was usually better than targeting multiple classes simultaneously; and (ii) the success rate for VA-assisted selection was higher than random selection (see Appendix \ref{apx:Results}). With this finding, iGAiVA was designed for ML developers to observe PCA scatter plots and RBF heatmaps and to select examples in a class-by-class manner, though the iGAiVA users can merge synthetic data generated using examples chosen from multiple classes.}

\revise{As iGAiVA represents a new approach. According to our observation, when a user works with a specific class, the user pays relatively little attention to other classes (i.e., the grey dots in PCA scatter plots). Hence, the effort for processing $m$ classes is largely linearly related to $m$.
Let us consider $x$ is the number of classes of a text classification problem, e.g., $x = 15$ in our example. Since most text classification problems have a relatively small number of classes, it is not difficult to conclude that the VA approach is scalable in this context.}

\revise{In many informal conversations, the term ``scalability'' is often used to convey a doubt, e.g., \emph{can humans handle more data than machines or can humans process data faster than machines?} One often hears the question of whether a human-centric process is scalable from colleagues who are interested in automatic techniques. Mathematically, the term ``scalability'' is not correctly used here. For a given problem with a well-defined algorithmic solution, of course, humans usually cannot handle more data than machines and humans cannot process data faster than machines. However, one should not draw a general conclusion that VA is not good enough, or use the incorrect term, ``VA is not scalable''. Meanwhile, a well-defined algorithm that is faster than humans can still be not scalable. Most timetabling algorithms are of NP complexity, and they are not scalable. However, they are faster than humans and that is why they are commonly deployed in universities.}

\revise{Currently, the visual patterns discussed in Section \ref{sec:VIS} are not well-defined and there is no adequate algorithm to identify such patterns. Hence VA allows humans to perform tasks that machines cannot perform. In addition, from identified patterns, an experienced ML developer can formulate hypotheses as discussed in Section \ref{sec:VIS}. There is no known algorithm that can transform such visual patterns into hypotheses. Hence in this context, the algorithmic ability of machines does not exist, and we cannot take the machines' advantages in terms of processing more data and processing faster.}

\vspace{1mm}
\revise{\textbf{Potential Challenges with LLMs.}
There must be many potential challenges with LLMs. It will not be appropriate for this group of authors to cover a broad spectrum of such potential challenges. From the perspective of this work, the current LLM-based GAI technology can help solve a major industrial problem, i.e., generating synthetic data for improving text classification models. During this work, we were involved in several discussions about the shortcomings of LLMs. For example, some questioned that LLMs might not generate realistic or correct text messages, assuming that the ML models must be trained with realistic and correct text messages. From an industrial perspective, the answer was rather simple. Many collected messages contain spelling and grammatical errors. The styles of some messages could easily lead to a conclusion as unrealistic if one did not know that it was collected. There are also mislabeled messages. Training ML models using imperfect data is a norm in the industry. As long as any less-ideal synthetic data can help improve a model, using such data is the correct approach.}

\revise{Some suggested that the cost of using LLMs in an industrial setting could be an issue. However, this does not affect the application of this research. Because LLMs are not deployed in ticketing systems, the hosts of ticketing systems do not need to bear the cost of using LLMs. The text classification models are developed, improved, and maintained in the VA+LLM workflows residing at the industrial partner. When the developed models are deployed in different ticketing systems, LLMs are not used. }

\vspace{1mm}
\revise{\textbf{Lessons learned,}
Both academic and industrial partners gained a lot of experience in this collaboration. We had the opportunity to read a collection of research papers. Academic researchers had the opportunity to work on real-world problems. Industrial partners had the opportunity to receive ideas about VA approaches that were unfamiliar to them before. Of course, there are many things that we could do differently, such as in some cases, meeting domain experts individually was more effective than a large meeting. In general, the project was conducted smoothly according to the plan.      }

\vspace{1mm}
\revise{\textbf{Limitations.}
The main contributions of this work are
(i) proposing a novel VIS4ML approach in which VIS techniques are used to guide the processes of data synthesis using LLMs.
(ii) Developing a software tool, iGAiVA, which integrates generative AI and visual analytics into a unified ML workflow.
(iii) Demonstrating the effectiveness of targeted data synthesis in
improving ML model accuracy through visual analytics.
In comparison with the existing workflow as shown in Fig. \ref{fig:Workflows}(a), the new workflow has many advantages as discussed throughout this paper and Appendix \ref{apx:Evaluation}. Perhaps the only perceived ``problem'' is that Inetum has to invest additional resources to transform iGAiVA to an industrial software product. This is a good problem to have than not so. 
}

\revise{A limitation that cannot be improved is not really a limitation. In comparison with a piece of future technology that improves upon some aspects of iGAiVA, those aspects that will be improved are limitations from future perspectives. For example, if there would be a better visual representation than the RBF heatmap, the current RBF heatmap has limitations. We can easily anticipate that there will be future technologies that will improve some aspects of the iGAiVA approach, such as improved LLMs, new and better visual designs, better user interface designs, and so on. Hence, from a future perspective, it is possible that every aspect of this work can be a limitation. That is why many published research works are now obsolete.}

\newpage

\section{High-resolution Images}
\label{apx:HighResImages}
In this appendix, we provide high-resolution versions of the images in Figs. \ref{fig:PCA-RBF-TTM}, \ref{fig:T13}, and \ref{fig:iGAiVA}.

\begin{figure*}
    \centering
    \textbf{Original Fig. \ref{fig:PCA-RBF-TTM} Caption}
    \parbox[c]{160mm}{Two examples of detailed visual analysis for investigating class T12. The two PCA scatter plots on the left show that Dimension 0 in (a) and Dimension 13 in (b) can separate the data objects into two regions, and data objects in one region have higher recall, while the overall class recall is only 37.5\%. Each RBF plot in the second column makes the boundary between the high-recall and low-recall regions clearer, enabling the selection of a division line to study the summary statistics of the messages in the two regions using a tag-treemap on the right.}\\[10mm]
    \includegraphics[height=125mm]{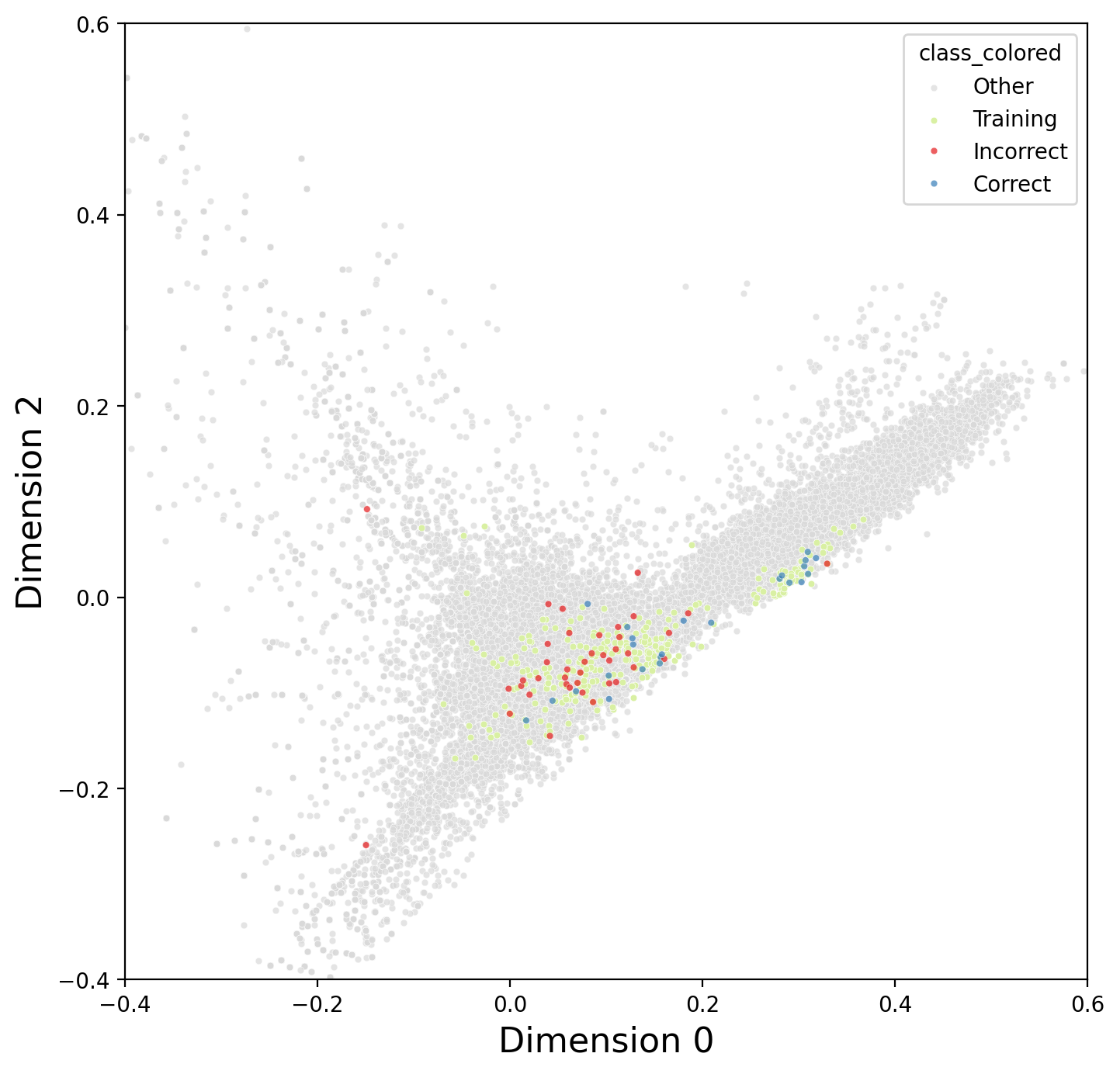}
    \caption{High resolution Fig. \ref{fig:PCA-RBF-TTM}(a): column 1.}
\end{figure*}

\begin{figure*}
    \centering
    \includegraphics[height=125mm]{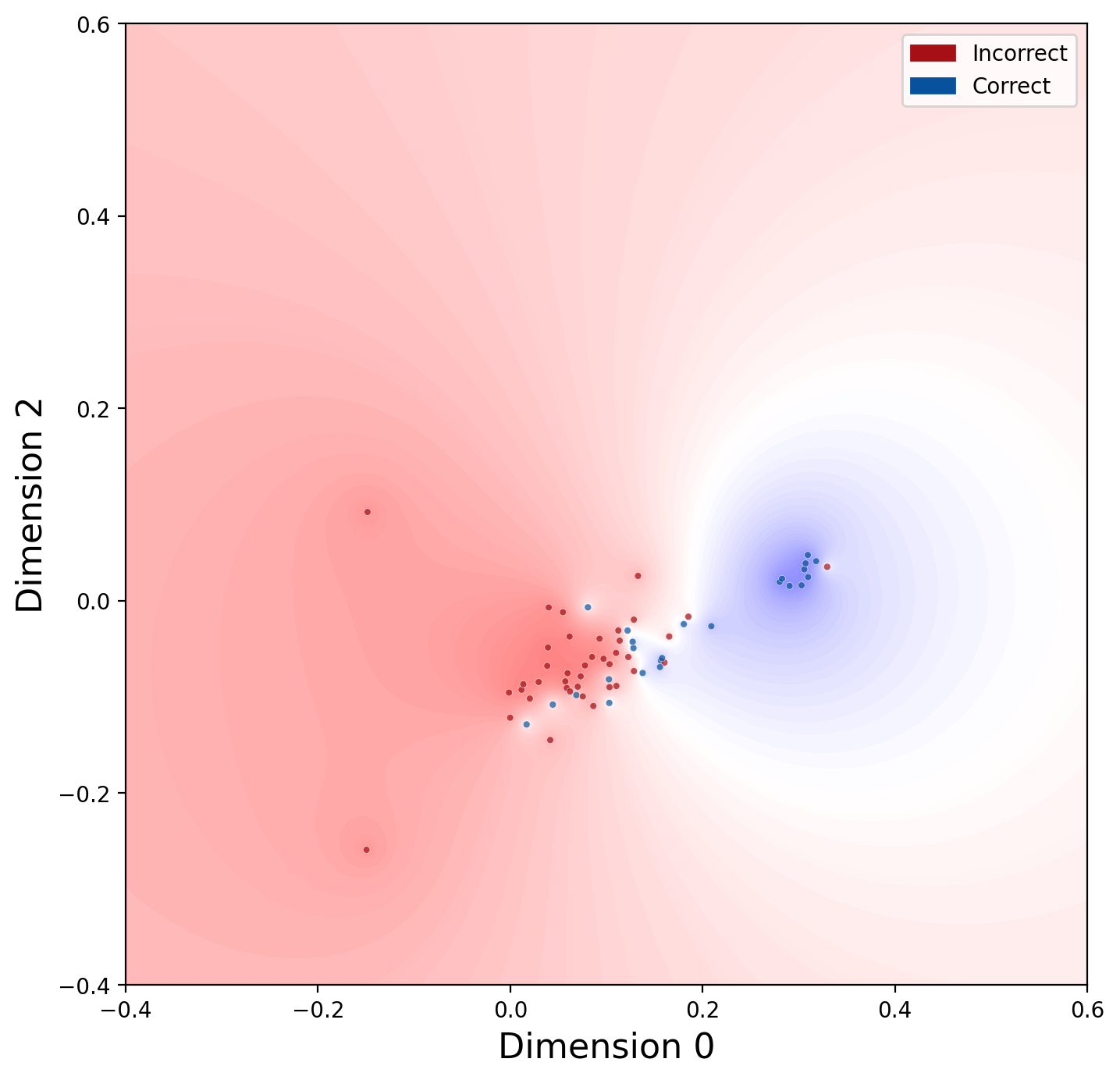}
    \includegraphics[height=125mm]{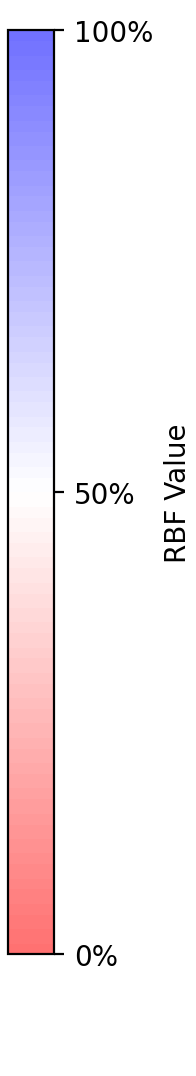}
    \caption{High resolution Fig. \ref{fig:PCA-RBF-TTM}(a): column 2.}
\end{figure*}

\begin{figure*}
    \centering
    \includegraphics[height=125mm]{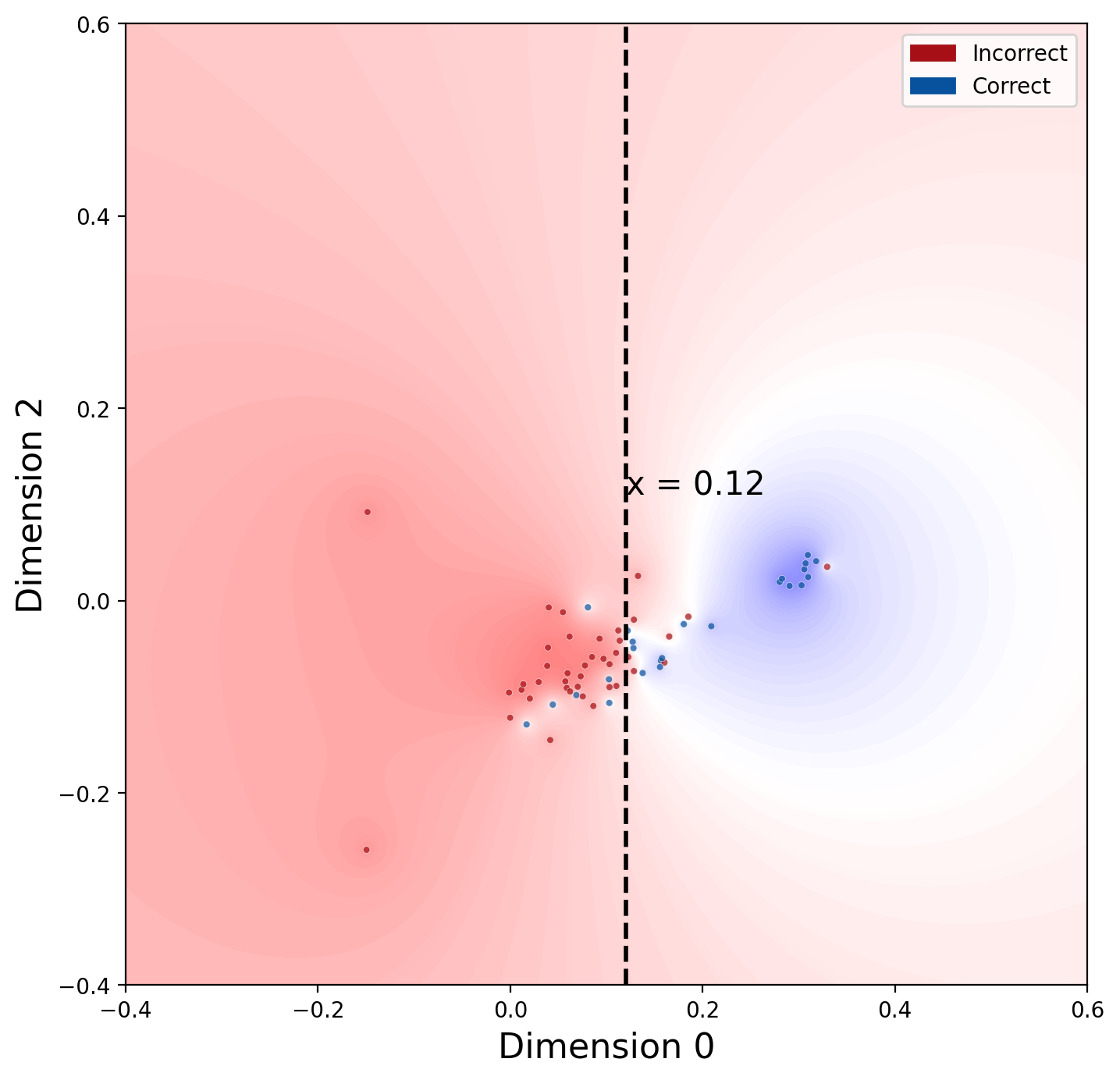}
    \includegraphics[height=125mm]{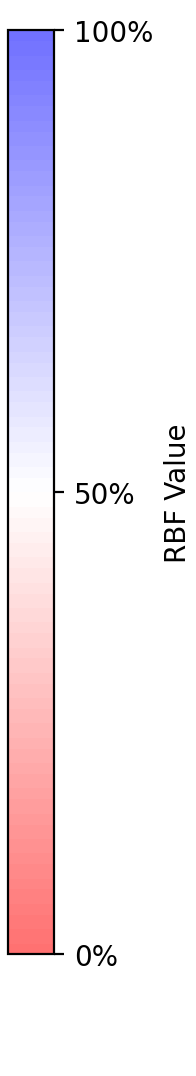}
    \caption{High resolution Fig. \ref{fig:PCA-RBF-TTM}(a): column 3.}
\end{figure*}

\begin{figure*}
    \centering
    \includegraphics[height=160mm]{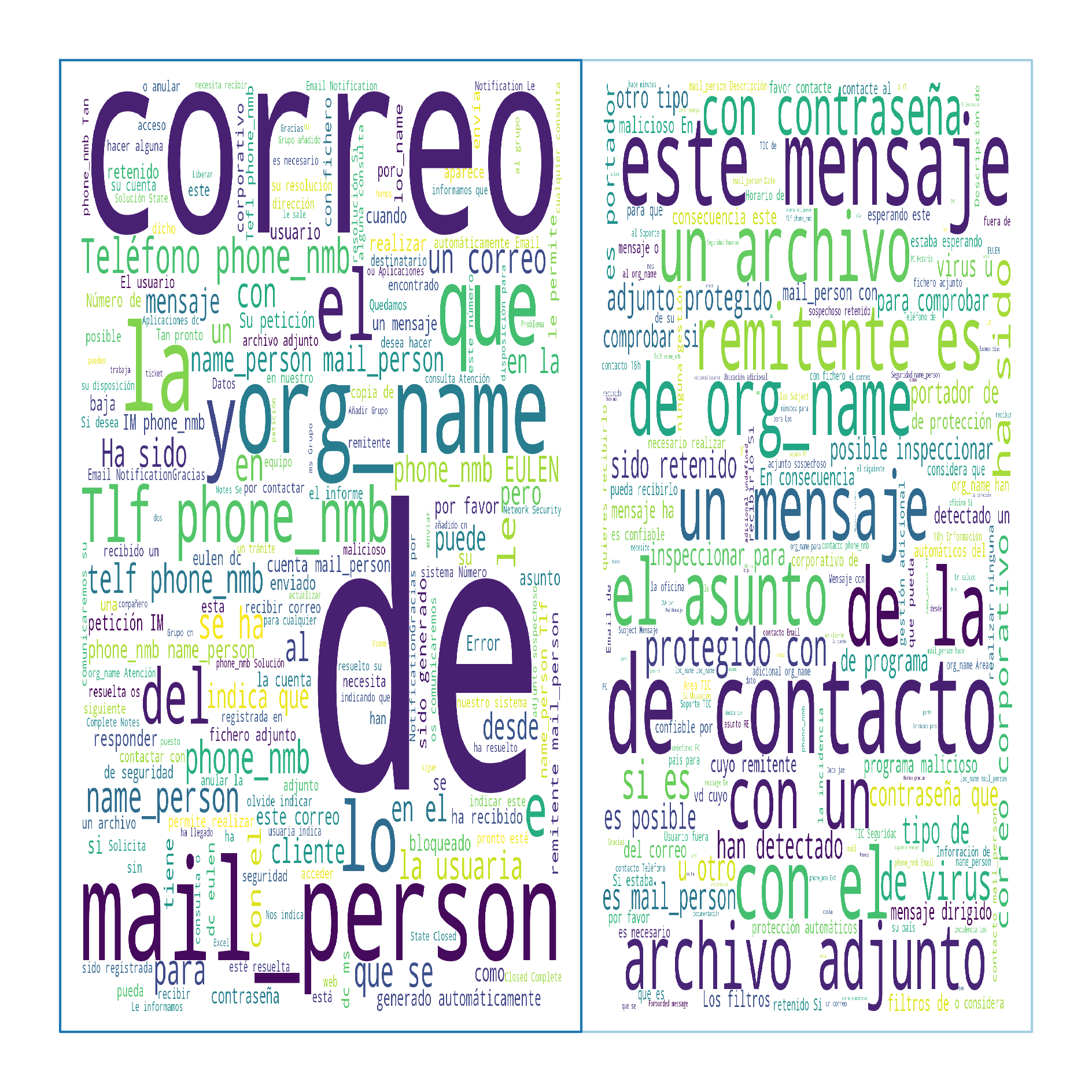}
    \caption{High resolution Fig. \ref{fig:PCA-RBF-TTM}(a): column 4.}
\end{figure*}

\begin{figure*}
    \centering
    \includegraphics[height=125mm]{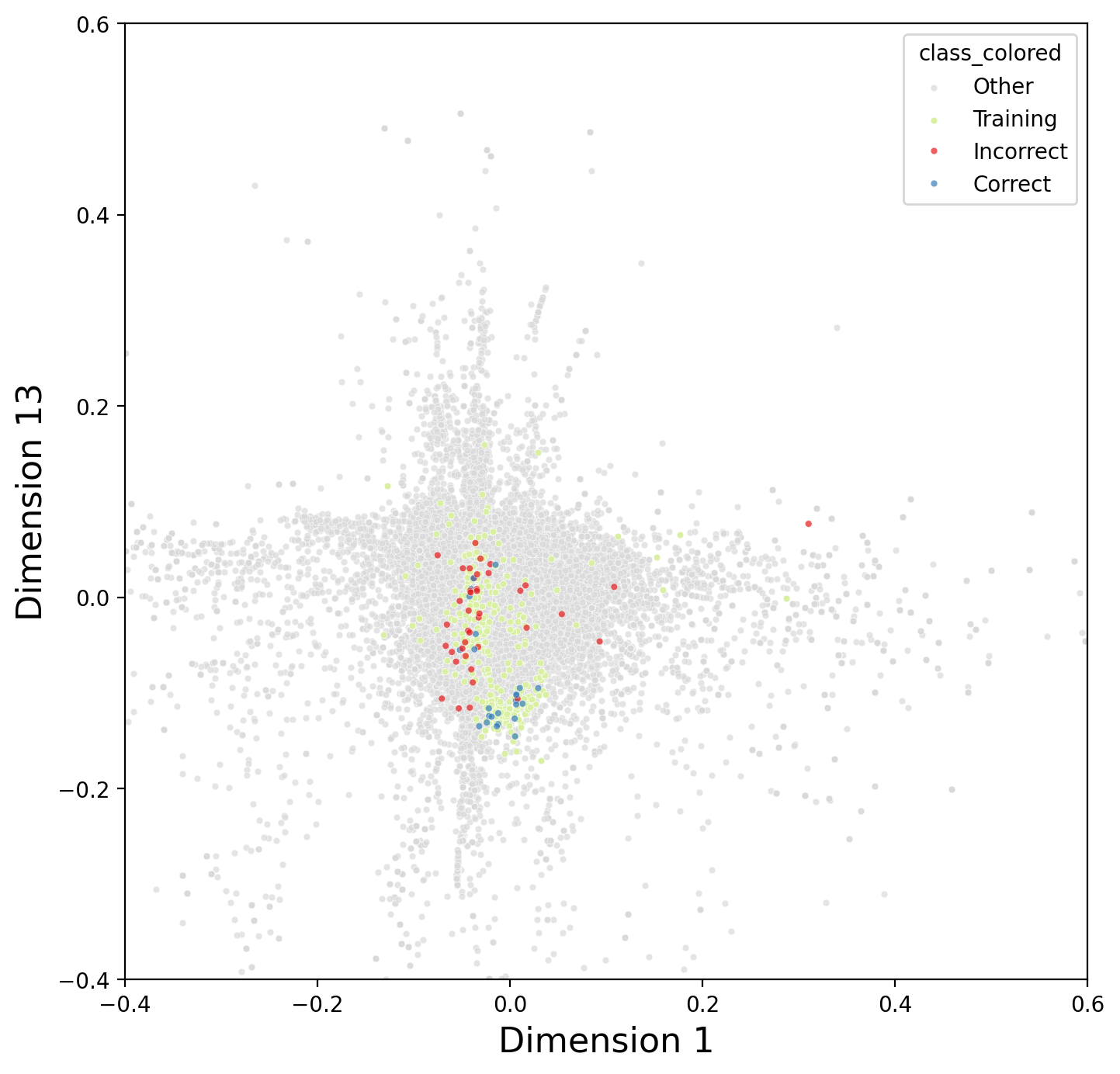}
    \caption{High resolution Fig. \ref{fig:PCA-RBF-TTM}(b): column 1.}
\end{figure*}

\begin{figure*}
    \centering
    \includegraphics[height=125mm]{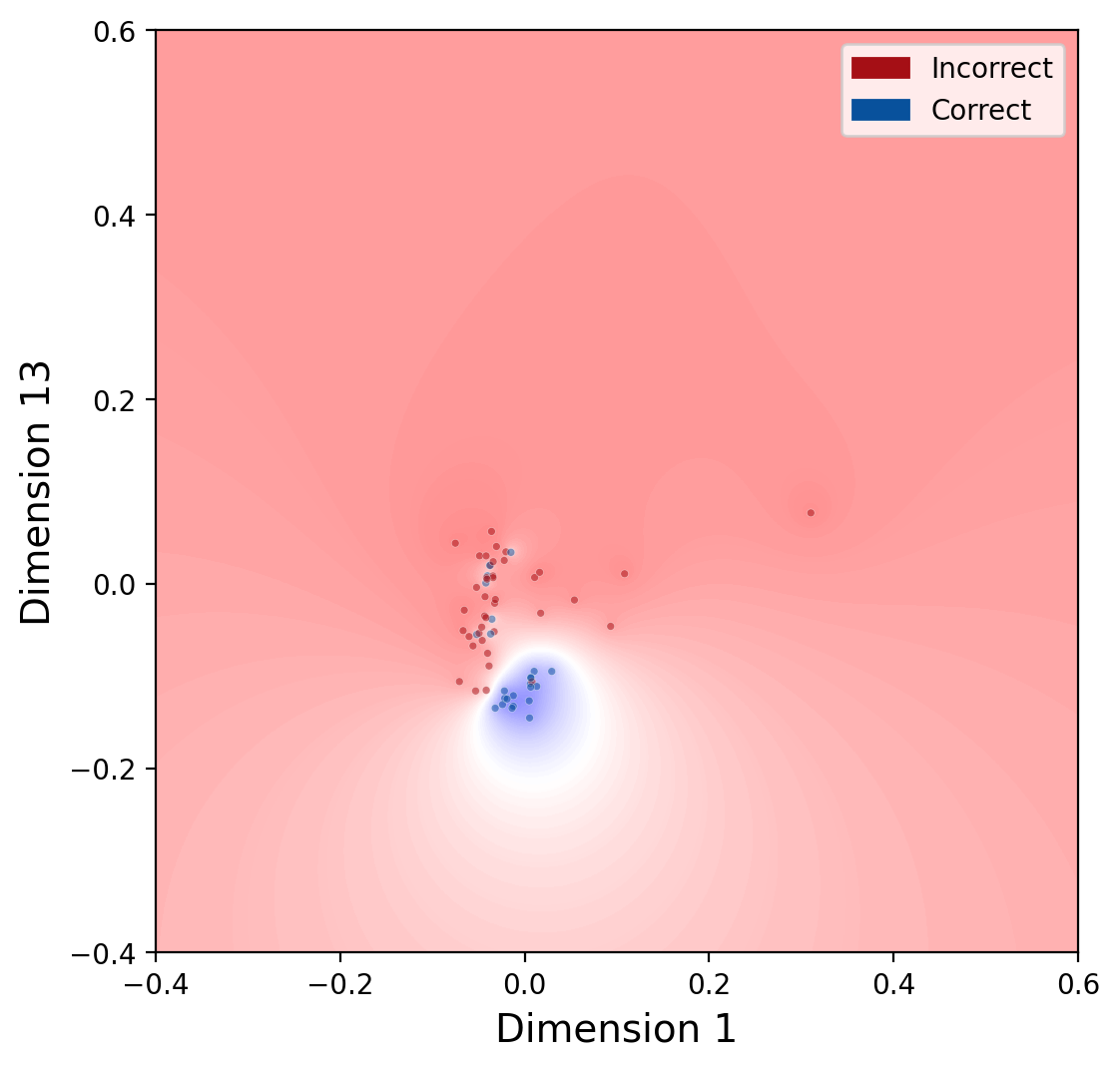}
    \includegraphics[height=125mm]{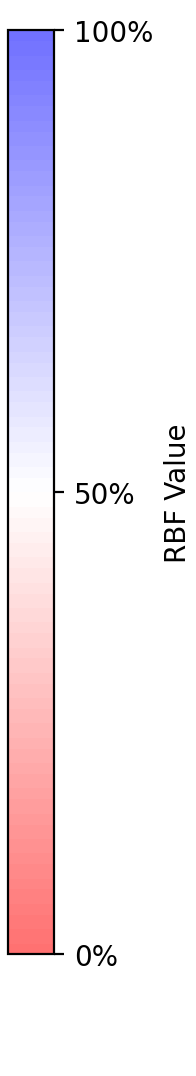}
    \caption{High resolution Fig. \ref{fig:PCA-RBF-TTM}(b): column 2.}
\end{figure*}

\begin{figure*}
    \centering
    \includegraphics[height=125mm]{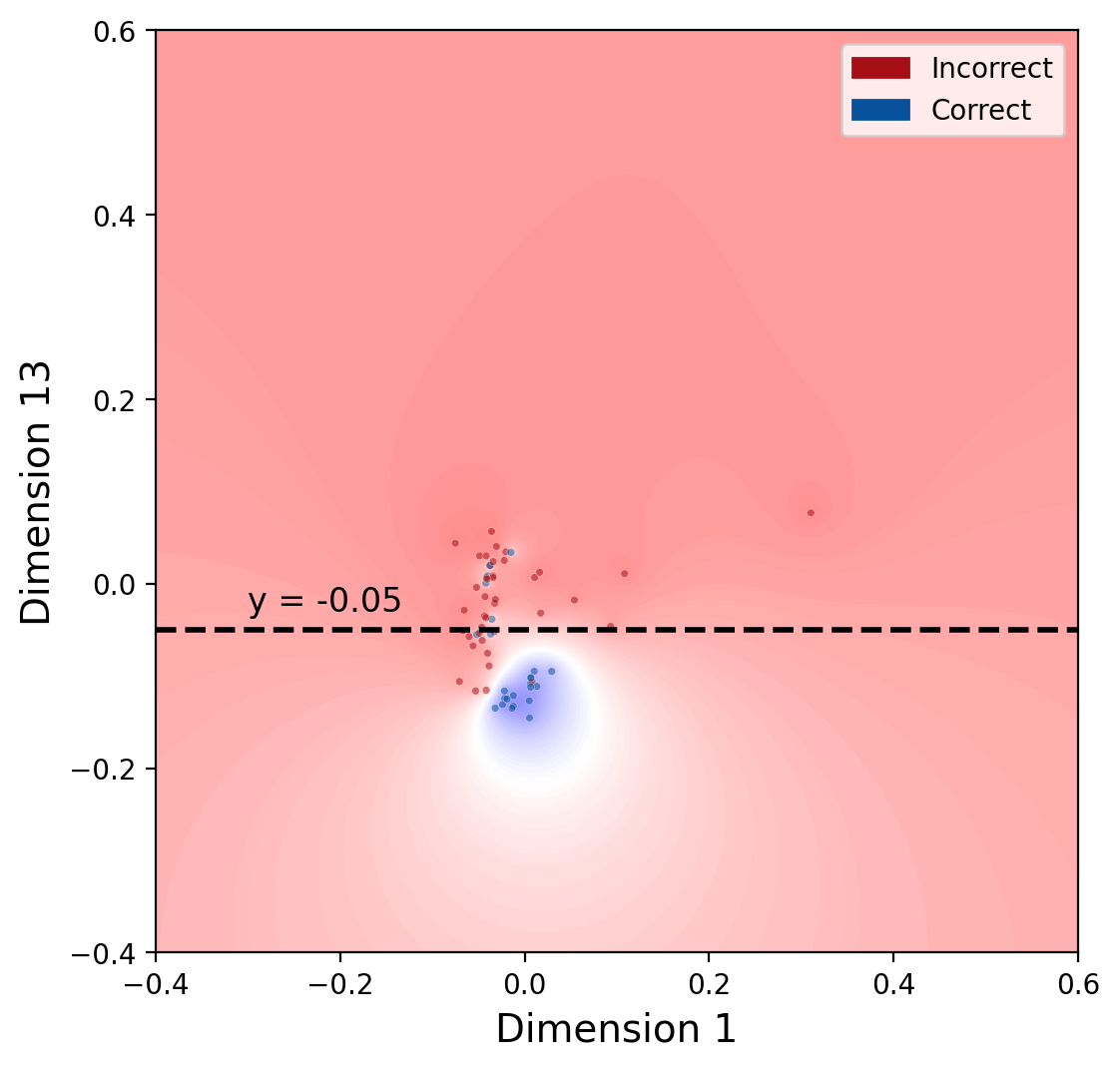}
    \includegraphics[height=125mm]{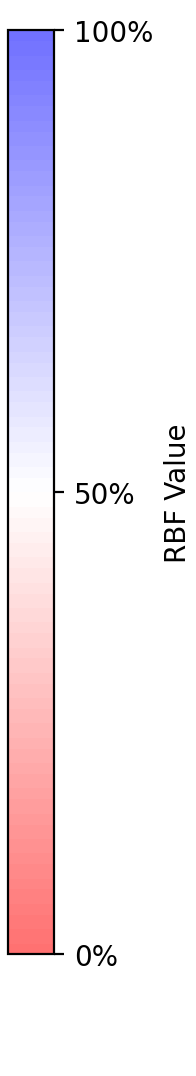}
    \caption{High resolution Fig. \ref{fig:PCA-RBF-TTM}(b): column 3.}
\end{figure*}

\begin{figure*}
    \centering
    \includegraphics[height=160mm]{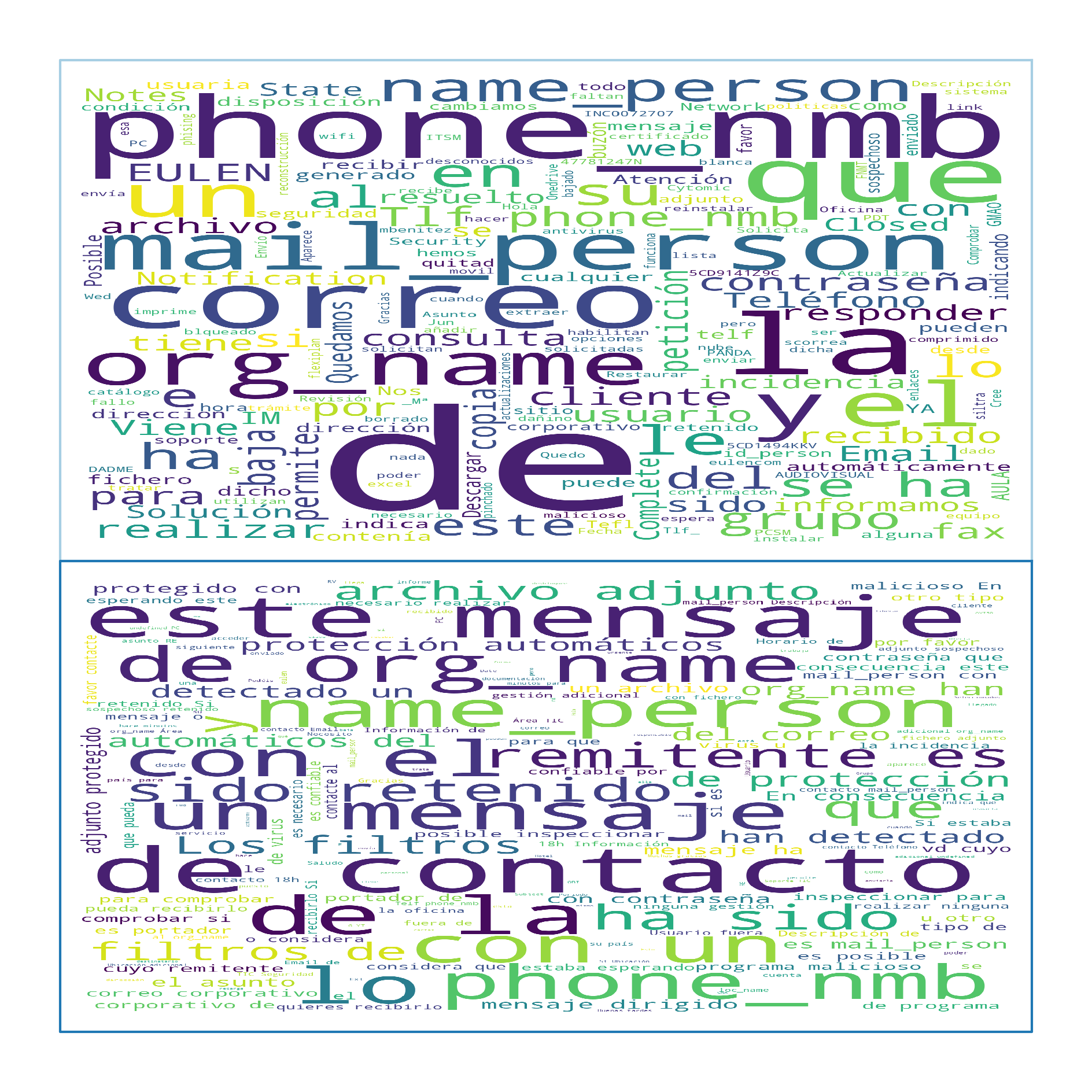}
    \caption{High resolution Fig. \ref{fig:PCA-RBF-TTM}(b): column 4.}
\end{figure*}


\begin{figure*}
    \centering
    \textbf{Original Fig. \ref{fig:T13} Caption}\\
    \parbox[c]{160mm}{The class T13 has the lowest recall among all classes. The scatter plot in (a) indicates more classification errors (red dots) when the data objects are associated with lower values in PCA feature dimension 0. The RBF heatmap in (b) confirms this pattern and enables data synthesis to be targeted at an erroneous cluster on the left as shown in (c). The model retrained with additional LLM-synthesized data is improved in (d). The RBF heatmap for the new testing results in (e) and the zoomed-in scatter plots confirm the improvement.}\\[10mm]
    \includegraphics[height=125mm]{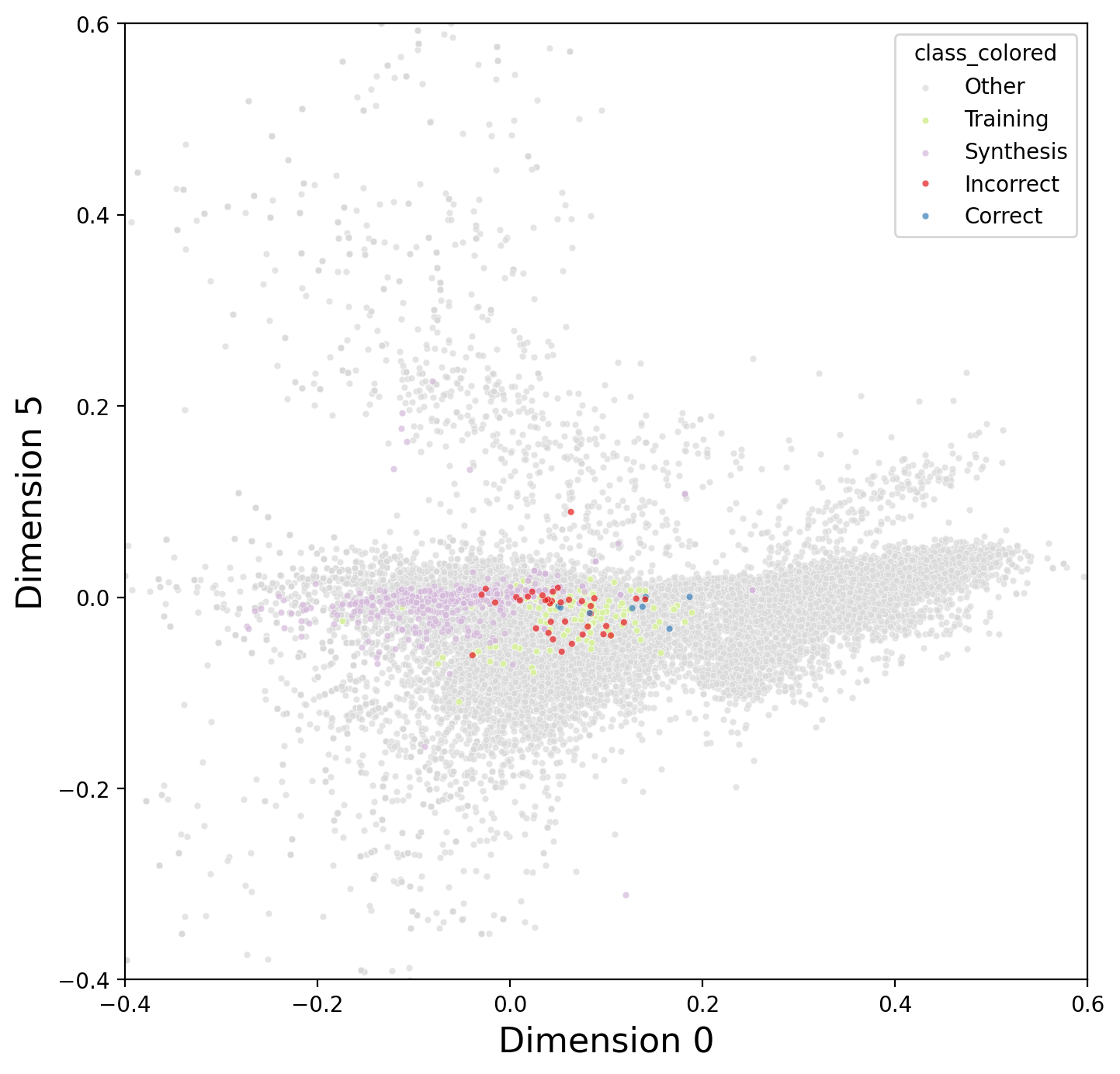}
    \caption{High resolution Fig. \ref{fig:T13}(a): correctness distribution (before).}
\end{figure*}

\begin{figure*}
    \centering
    \includegraphics[height=125mm]{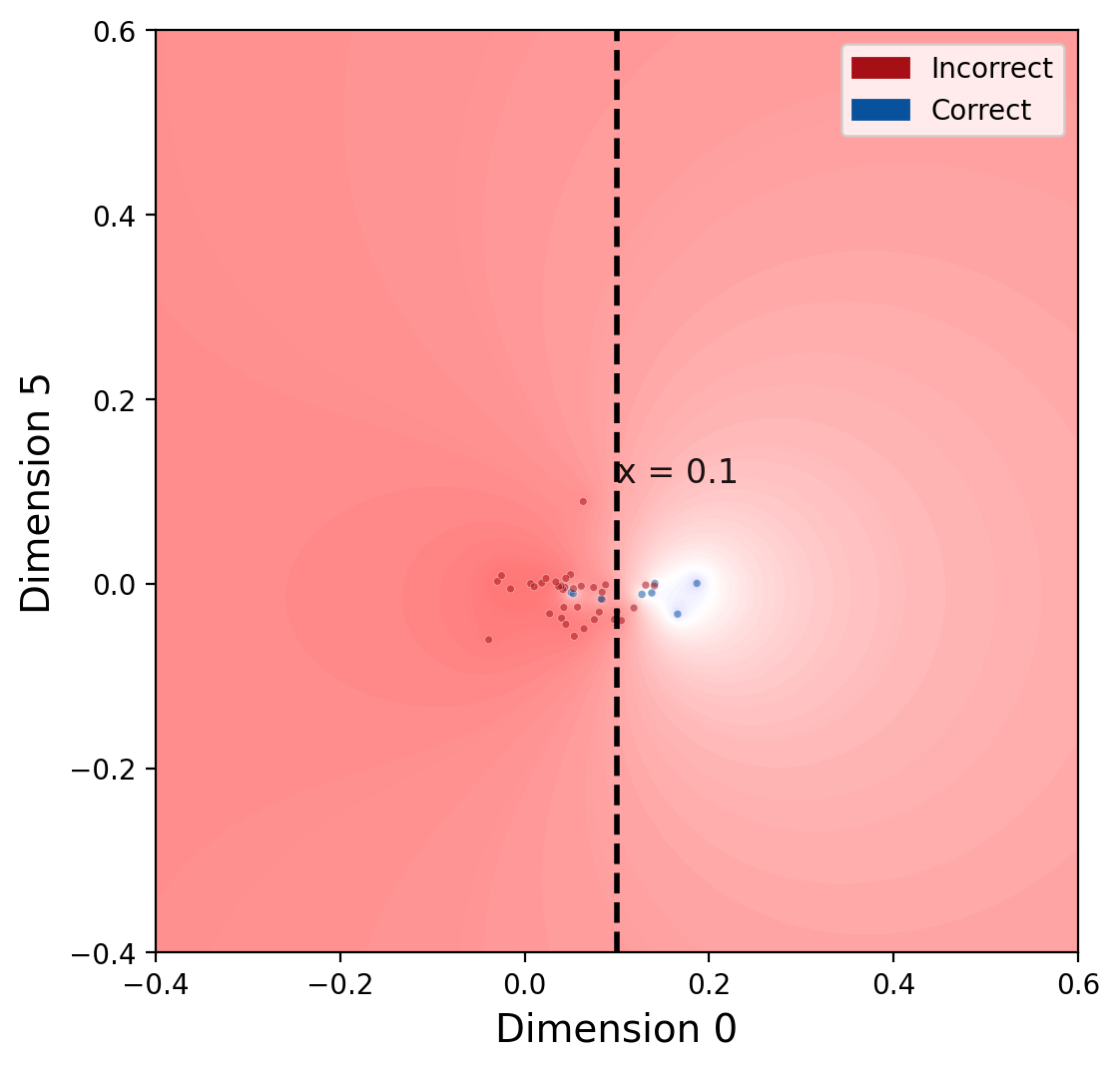}
    \includegraphics[height=125mm]{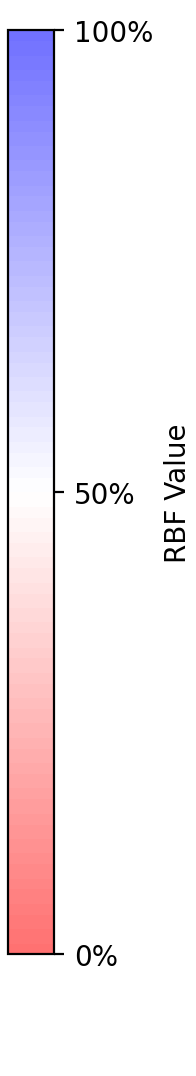}
    \caption{High resolution Fig. \ref{fig:T13}(b): two area divided by dimension 0.}
\end{figure*}

\begin{figure*}
    \centering
    \includegraphics[height=125mm]{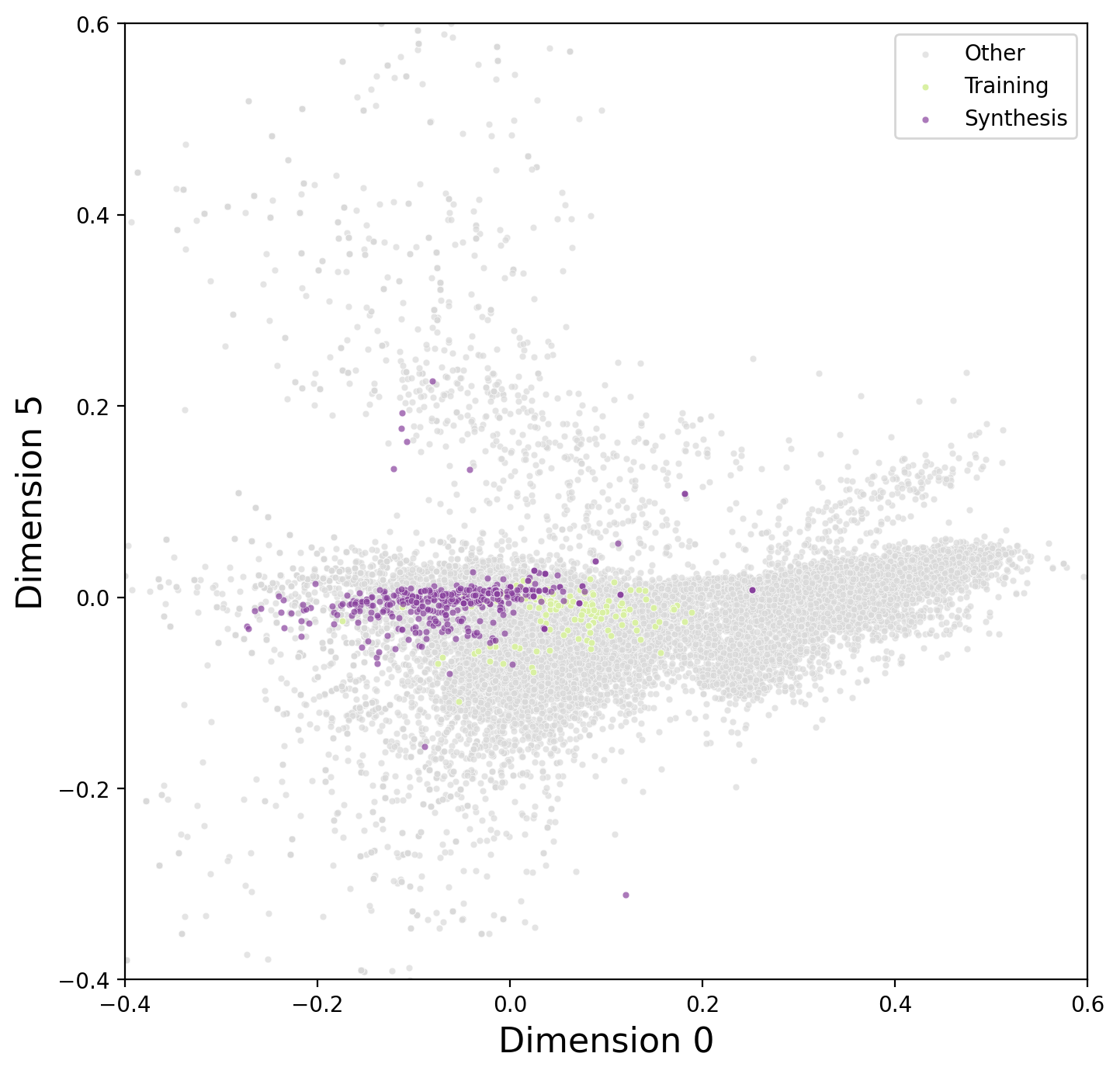}
    \caption{High resolution Fig. \ref{fig:T13}(c): distribution of synthesized data.}
\end{figure*}

\begin{figure*}
    \centering
    \includegraphics[height=125mm]{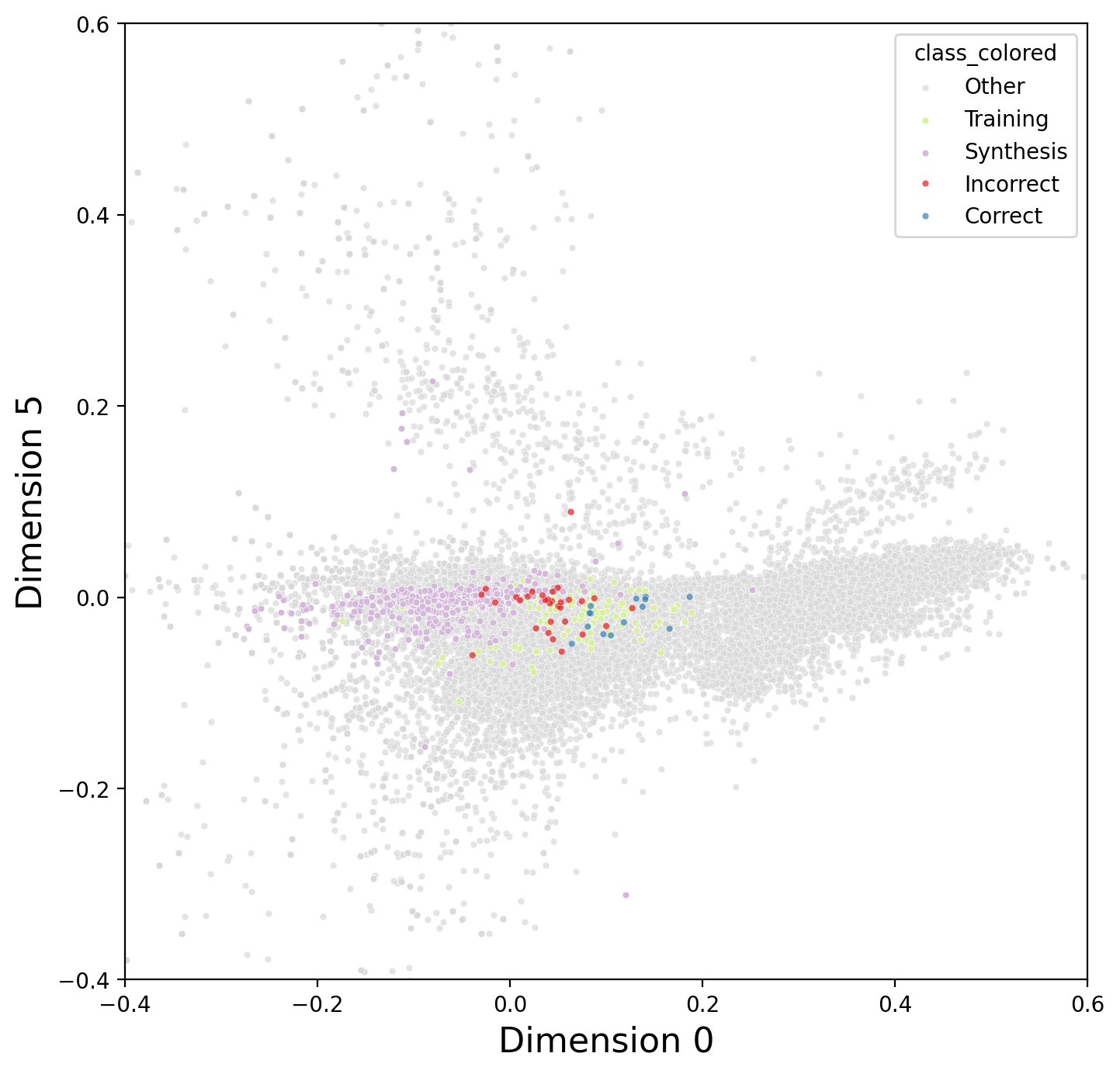}
    \caption{High resolution Fig. \ref{fig:T13}(d): correctness distribution (after).}
\end{figure*}

\begin{figure*}
    \centering
    \includegraphics[height=125mm]{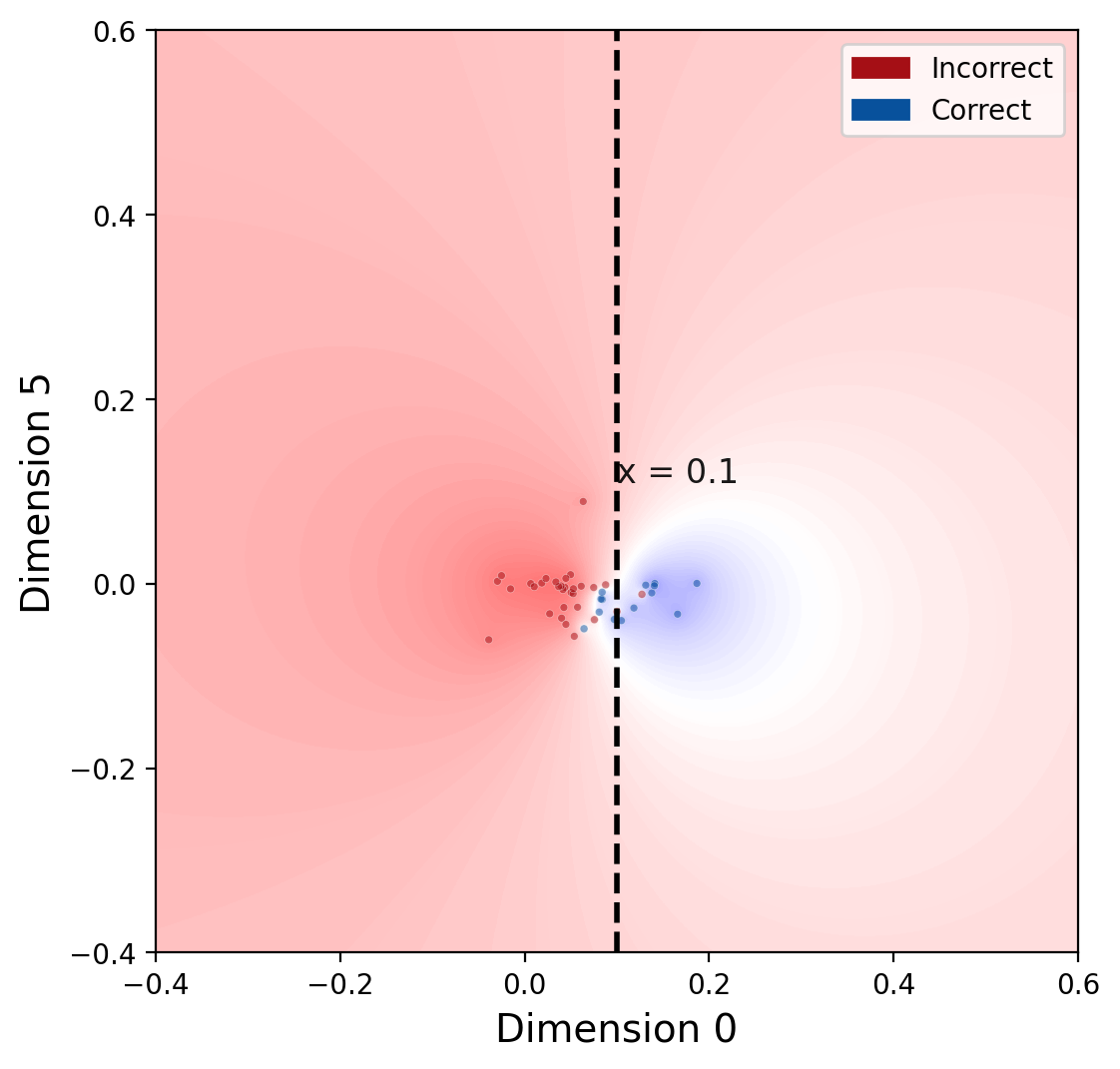}
    \includegraphics[height=125mm]{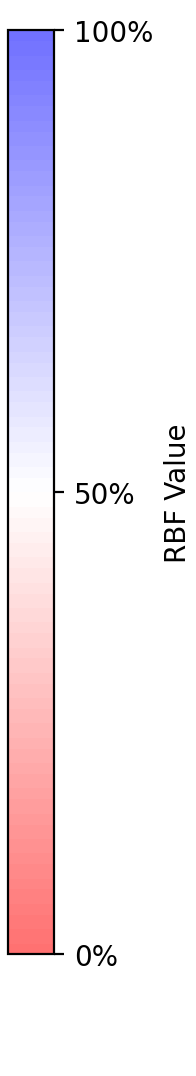}
    \caption{High resolution Fig. \ref{fig:T13}(e): noticeable improvement in RBF.}
\end{figure*}

\begin{figure*}
    \centering
    \includegraphics[height=125mm]{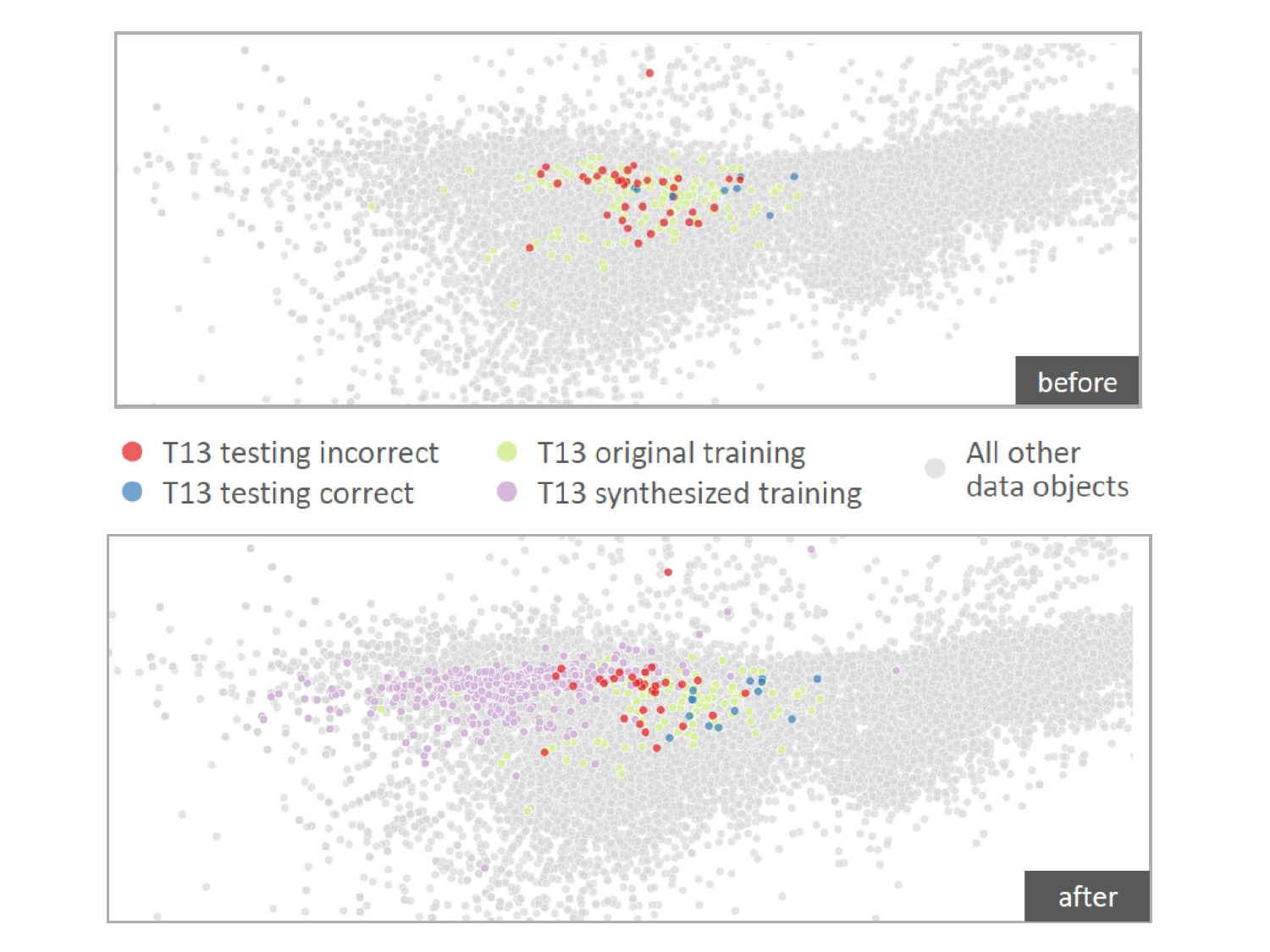}
    \caption{High resolution Fig. \ref{fig:T13}(f): zoomed PCA scatter plots.}
\end{figure*}
\begin{figure*}
    \centering
    \textbf{Original Fig. \ref{fig:iGAiVA} Caption}\\
    \parbox[c]{160mm}{Four views of the iGAiVA tool. The user can switch between views using the top menu bar. (a) The Synthesis View is for supporting mainly the tasks for identifying suitable example data objects as inputs to LLMs for generating synthetic data. (b) The Data View is for selecting a subset of synthetic datasets and combining them with the original training data. (c) The Model View is for monitoring the process of retraining a model and running the retrained model against one or more predefined testing datasets.}\\[10mm]
    \includegraphics[width=160mm]{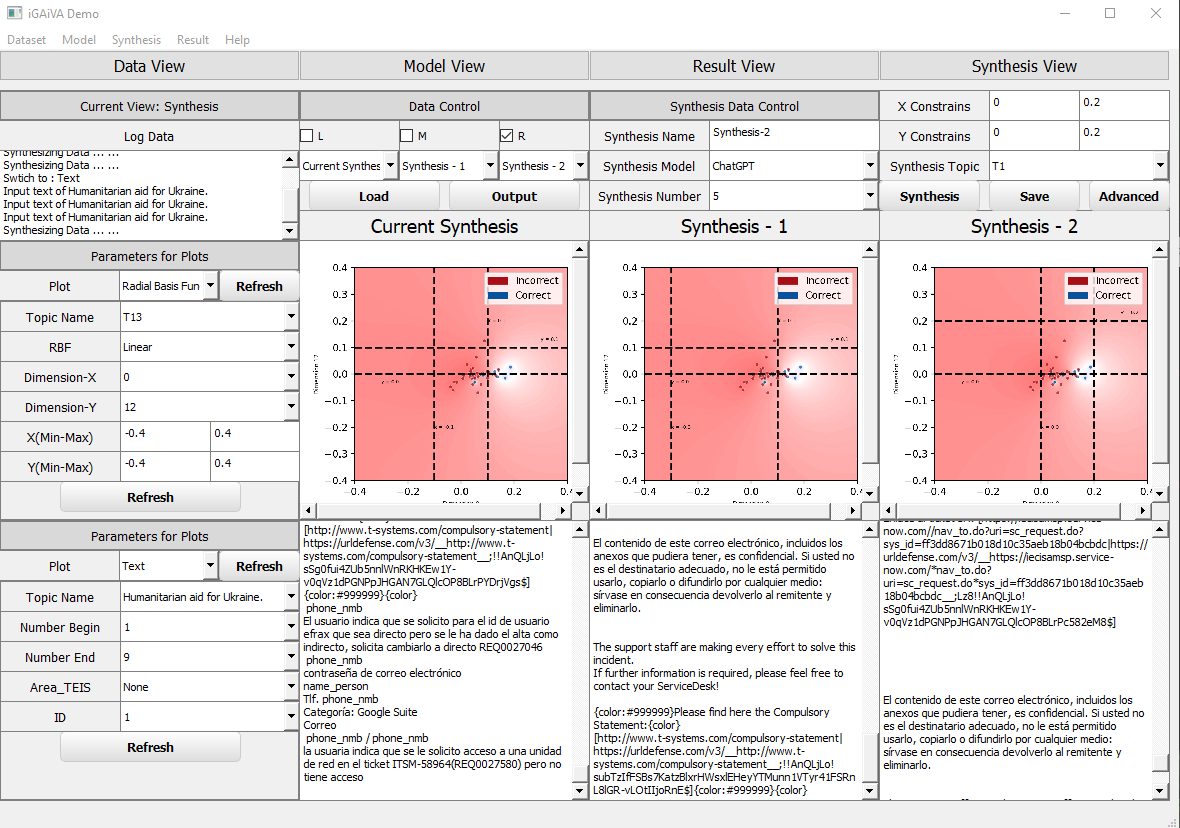}
    \caption{High resolution Fig. \ref{fig:iGAiVA}(a): Synthesis View.}
\end{figure*}

\begin{figure*}
    \centering
    \includegraphics[width=160mm]{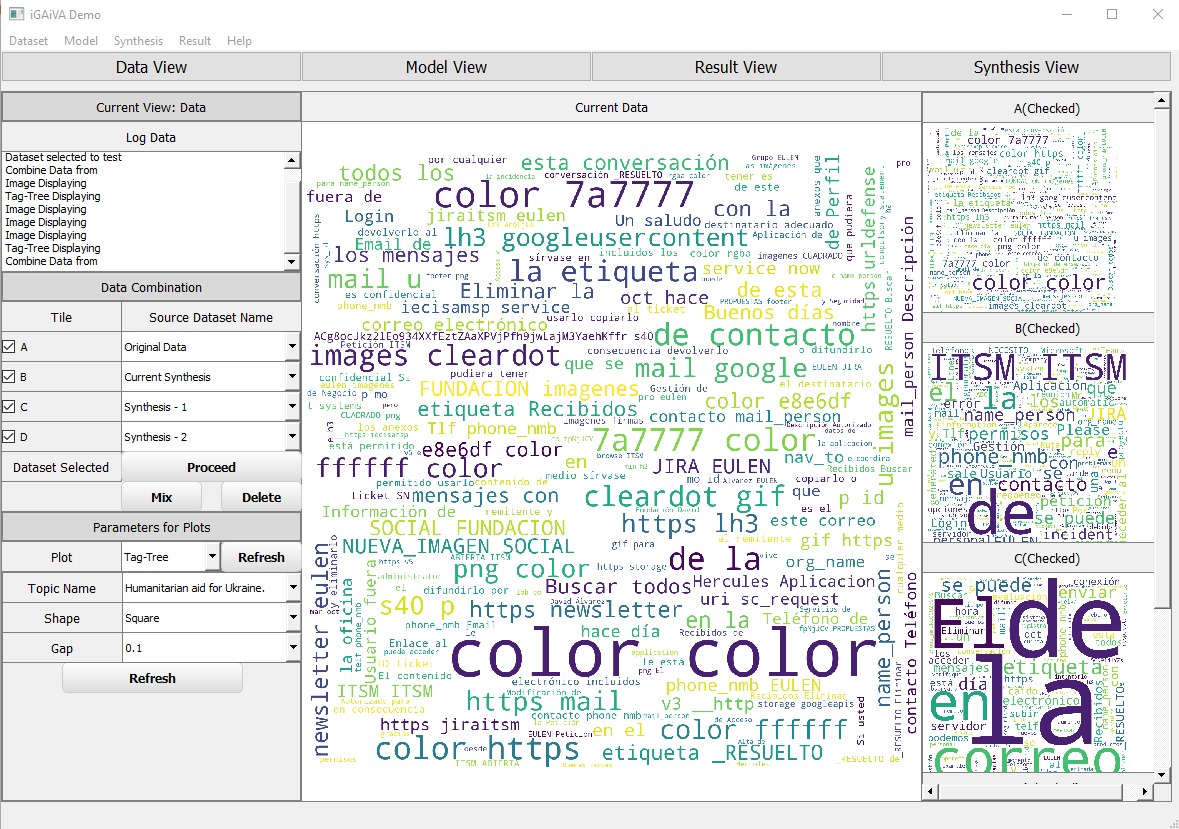}
    \caption{High resolution Fig. \ref{fig:iGAiVA}(b): Data View.}
\end{figure*}

\begin{figure*}
    \centering
    \includegraphics[width=160mm]{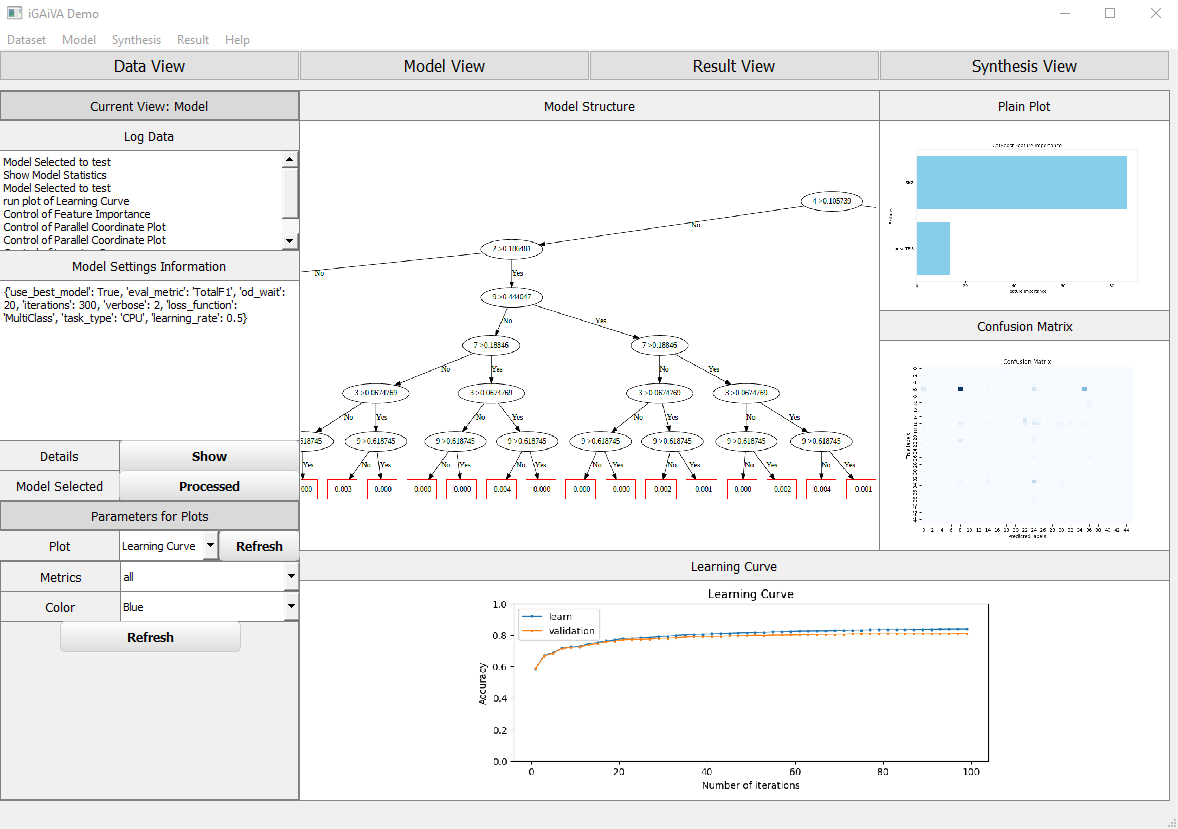}
    \caption{High resolution Fig. \ref{fig:iGAiVA}(c): Model View.}
\end{figure*}

\begin{figure*}
    \centering
    \includegraphics[width=160mm]{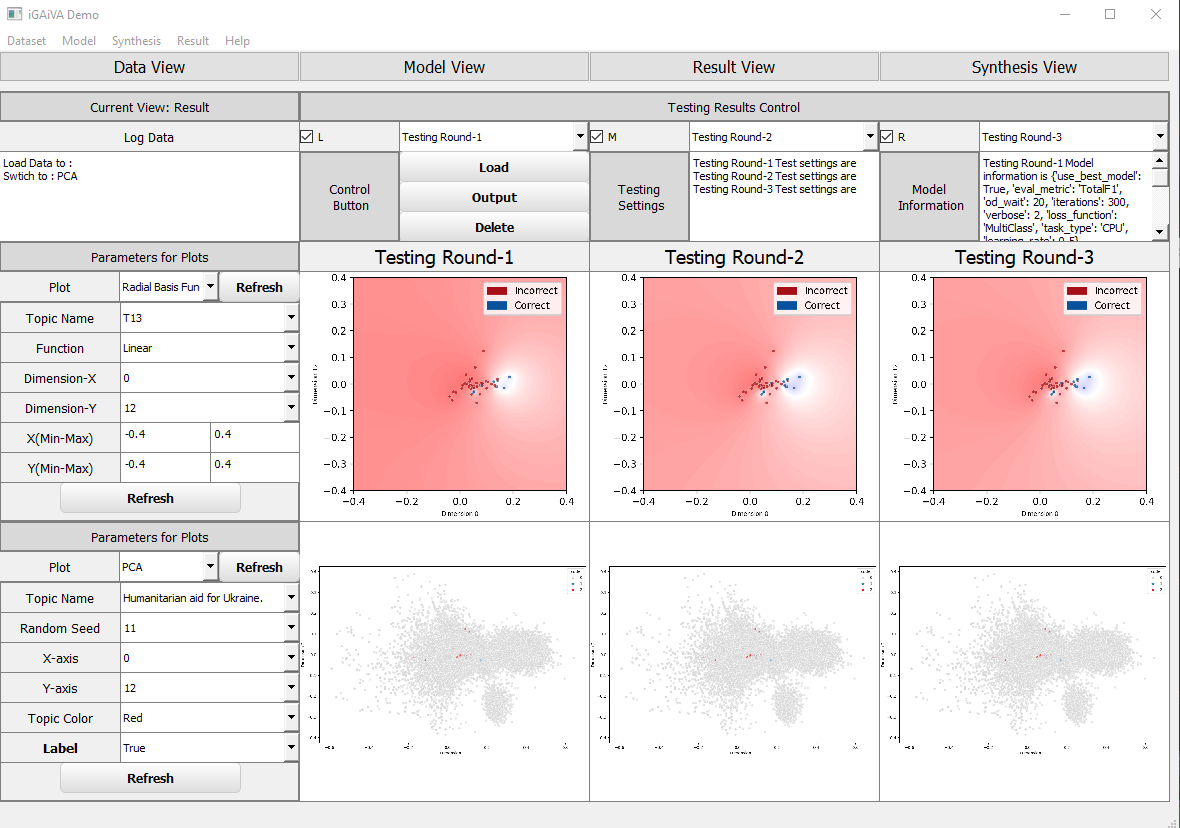}
    \caption{High resolution Fig. \ref{fig:iGAiVA}(d): Results View.}
\end{figure*}

\newpage

\section{Feedback Letter to the Co-Chairs and Reviewers of PacificVis 2025 Conference track}
\label{apx:FeedbackLetter}
\emph{The following feedback letter was sent to the co-chair of PacificVis 2025 Conference track on January 4, 2025.}

\begin{helveticasection}
\begin{small}
\vspace{2mm}\noindent
Dear PacificVis 2025 Conference Track Co-Chairs,\\
Dear \# 5992 and \#9555 Reviewers

\vspace{2mm}\noindent
\textbf{RE:	\#9555 iGAiVA: Integrated Generative AI and Visual Analytics in a Machine Learning Workflow for Text Classification}

\vspace{2mm}\noindent
Thank you for your time and effort in considering this work. The authors are fully aware that the review process has finished, and the final decision has been made. While this feedback letter has no material implication, the authors would like to draw the reviewers’ attention to the misjudgement about the leakage in the experimental setup. If such a judgment were correct, it would be fair to reject this paper. However, there were explicit statements in the paper about the fact that only training data was used to generate synthetic data. This is rather unexpected as the reviewers’ oversight about existing texts happened in the second review cycle rather than the first one. For details, please see the point A2 following this letter. We believe that the reviewers’ new comment on this leakage was the critical element of the rejection decision, as all other issues raised were either addressed or discussed in the revision report following the TVCG track submission.

\vspace{2mm}\noindent
We believe that letting reviewers know such a misjudgement is helpful for maintaining the scientific integrity and professionalism of the review process. The author plan to make this conference track version available at arXiv, together with the TVCG track reviews, our revision report, Conference track reviews, and this feedback letter. The previous submission to the TVCG track was placed on the arXiv on 24/09/2024 (https://arxiv.org/abs/2409.15848). Hopefully, this may provide future reviewers of this work a little help to avoid making the same misjudgement.

\vspace{2mm}\noindent
\emph{This letter and the previous revision report were both written in the name of the last author (Min Chen), because some feedback texts in both documents could be read as critical to some of the review comments, and the last author would like to take the full responsibility for such texts. Hope that any negative feeling due to such texts will not be directed towards other authors, especially the student first author.}

\vspace{2mm}\noindent
Happy New Year,\\
Yours sincerely,\\
$\langle$the third author's name$\rangle$, on behalf of $\langle$two other authors' names$\rangle$

\vspace{6mm}\noindent
Summary Review:

\vspace{2mm}\noindent
Q1.	\texttt{The feedback from the previous round of reviews has not been adequately accounted for.}

\vspace{2mm}\noindent
A1.	The authors argued in the revision report that some of the requests in the previous reviews (e.g., conducting formative evaluation and featuring novel visual designs) are not justified for the TVCG track submission. Such requests appear even less justifiable for a conference track submission. We also noted that the reviewers did not provide a counter-argument to justify such request. (See also authors’ previous revision report.)

\vspace{2mm}\noindent
Q2.	\texttt{In particular, all reviewers expressed concerns about the issue of leakage in the experimental setup, which makes the improvement that is reported as having been achieved with the system unreliable.}

\vspace{2mm}\noindent
A2. ``\texttt{overfit to the test data}'', ``\texttt{… red points … then essentially you are using the label information of the test set to get more training data.}'' [R1, R2] -- This is a new review comment. Reviewers guessed this incorrectly. In both versions of the paper, there were explicit statements stating that only the training data was used to generate synthetic data. They are:

\vspace{2mm}\noindent
TVCG track version: Section 5, Page 6, Left Column, Paragraph 2:\\
Conf. track version: Section 5, Page 7, Left Column, Paragraph 1:\\
``\textrm{We then select example messages from the training data on the left part and generate 525 synthetic messages as additional training data.}''

\vspace{2mm}\noindent
TVCG track version: Section 5, Page 6, Left Column, Last three lines:\\
Conf. track version: Section 5, Page 7, Left Column, Paragraph 3:\\
``\textrm{When we generate synthetic data using the LLM, we always select example messages from the training data as we are aware that selecting examples from testing data could introduce biases in favor of testing.}''

\vspace{2mm}\noindent
TVCG track version: Section 6, Page 7, Left Column, Paragraph 3:\\
Conf. track version: Section 6, Page 8, Left Column, Paragraph 1:\\
``\textrm{... and use PCA scatter plots or RBF heatmaps to determine a subarea for selecting examples (from the training data) as inputs to LLMs.}''

\vspace{2mm}\noindent
It is possible that some reviewers made a guess when viewing the plots in Fig. 5 (page 6 in both versions) and guess that the red dots were selected as examples. In the TVCG track version, there were two passages on the same page talking about using training data only. It is possible that other reviewers just agreed with these reviewers without checking.

\vspace{2mm}\noindent
We knew the potential leakage from the very beginning of the project, and that is why the user interface of iGAiVA provides the facility for selecting an area, rather than selecting red dots. There is no text in the paper that mentions about using testing data as examples for generating synthetic data.

\vspace{2mm}\noindent
If reviewers’ this comment had come up in the review cycle for the TVCG track submission, the authors would be able to provide feedback, pointing the reviewers to the relevant texts in the paper. If the reviewers had been aware of the uncertainty and the risk of misjudgement, the reviewers could have asked the co-chairs to contact authors to seek clarification.

\vspace{2mm}\noindent
\textbf{The meaning of overfitting.} The words overfitting and underfitting are negative. However, the word better-fitting is NOT negative. Sometimes, some students and colleagues confused such terms by considering all ``fitting'' as ``overfitting''. Using the plots in Figs. 4 and 5 as examples, the testing results (red and blue dots) indicate the model performance about a specific class X. The yellow dots show where the training data of Class X is in a feature space, while the grey dots show the data of other classes. Because the data in different classes are unbalanced, one relatively common issue is that the errors in class X (red dots) were caused by a model that may ``better-fit'' the training data of some other classes but fit class X relatively poorly. So from class X's perspective, the model overfits the data of one or more other classes. Another common issue is that the model may fit well for some class X testing data, but not so well for other class X testing data (could be due to either overfit or underfit the training data). In a ticketing system, the class definition is often related to the organizational structure of a customer company. Multiple topics could easily be merged into a class, and the customer company may not have other formal mechanisms to define such multi-topic classes.

\vspace{2mm}\noindent
If such erroneous results are clustered, iGAiVA allows the ML developers to find a localised solution to train a better model that provides better fit in a local area. Synthetic data can be used to deal with both overfitting and underfitting problems. An experienced ML developer can usually work out after one or a few trials. VA can help conduct such trials more effectively and efficiently.

\vspace{2mm}\noindent
We also wonder whether R1 might mistake the distance in 2D plots, which are orthogonal projections of 20D to 2D, as the actual distance in the 20D space. So a cluster of red dots suggests these test messages may be similar in terms of the two features used for the 2D plots, but likely be distributed widely in the other 18D space (not mentioning in the actual textual space). The figure below (\emph{on the next page}) illustrates the variations in just one of the 18 dimensions. The ML developers using iGAiVA have a good knowledge and adequate mental image about the 20D feature space, and they never raised the overfitting or information leakage question as they knew that the testing data was not used for generating synthetic data.

\begin{figure*}[t]
    \centering
    \includegraphics[width=0.9\linewidth]{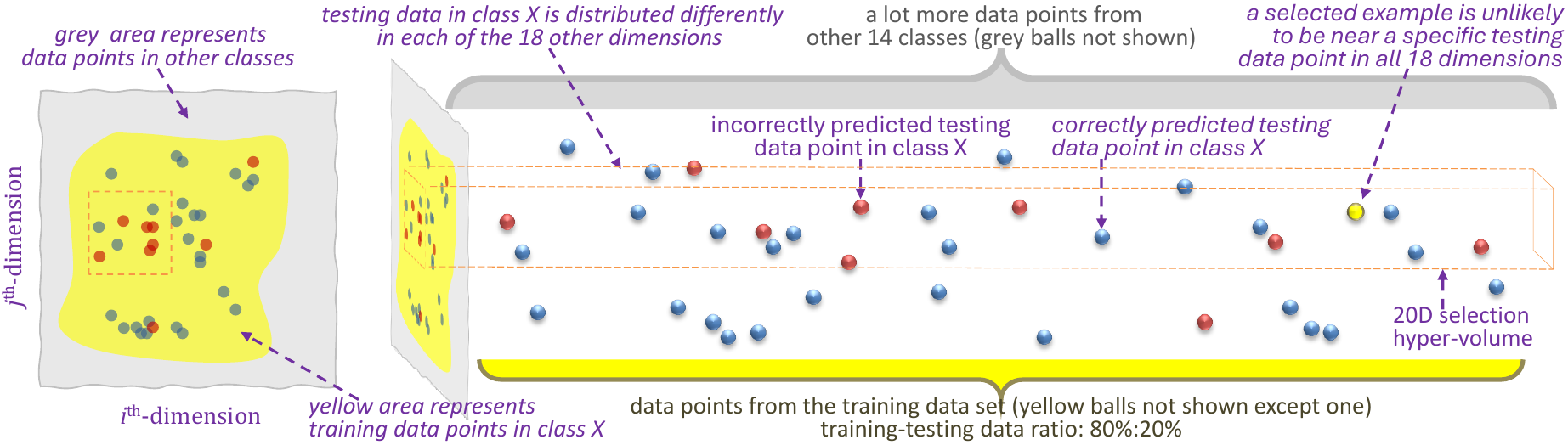}
    \label{fig:20D-features}
\end{figure*}

\vspace{2mm}\noindent
We encourage the reviewers to consider such an approach is similar to scenarios where school-teachers identify the weakness of an individual pupil through tests, and then set more assignments targeted at such weaknesses. Note the teachers do not give the test questions to the pupil as the extra assignments. Those extra assignments can differ from the test questions in at least 18 different aspects. We would not label these teachers overfitting.

\vspace{2mm}\noindent
\textbf{A bigger philosophical question about VIS4ML.} Note that R2's previous comment (in the TVCG Track review cycle) on experiment setup was different from this comment. The previous comment argues that ML developers ``\texttt{must never see the test data}''. Otherwise, the experiment set up would be an issue. Almost all experienced ML researchers and developers inspect testing data and results. Many previous VIS4ML papers were about helping ML developers inspect testing data and results. Observing such data allows ML developers to transfer information about the testing data and results to hypotheses about how to improve the model through changes to the training data, algorithms, and parameters as well as model architectures, The hypotheses are then transferred to actions for new experiments and iterations.

\vspace{2mm}\noindent
If such information leakage were considered to be inappropriate, the majority research in the ML field would be inappropriate. In active learning, selecting training data after observing interim testing results would be inappropriate. In reinforcement learning or genetic algorithms/programming, changing fitness functions or mutation methods, after observing some testing data, would be inappropriate. In one recently Oxford paper (KDD2024, 10.1145/3637528.3672004), their model ensemble algorithm was based on testing results. This would be inappropriate.

\vspace{2mm}\noindent
In the 2017 VAST best paper (TVCG, 10.1109/TVCG.2016.2598829), Tam et al. showed that VA enables human to inject knowledge into an ML workflow, and this is a good thing rather than a bad thing. The whole ethos of VIS4ML is based on this thinking.

\vspace{2mm}\noindent
When ML colleagues talking about avoiding information leakage, it is usually in the context of comparing two similar automated methods, e.g., decision tree vs random forest. In such a case, adding a new feature to ONLY one method after observing testing data would result in an unfair comparison. However, when comparing two significantly different methods, e.g., decision tree vs CNN, such information leakage is not considered as unfair. At the end, one just wants to compare an optimized decision tree against an optimized CNN. Both optimization processes inevitably based on observing testing results in the development stage and early iterations of comparison.

\vspace{2mm}\noindent
When considering VIS4ML methods (and to a large extent, active learning), the main idea is to enable humans to transfer their knowledge and the information they observed to actions that may lead to a better model in a more effective way than with such knowledge and information.

\vspace{2mm}\noindent
In industrial applications, the focus is rarely about which ML method is better, but about the question ``can a model be further improved?'' If human knowledge or observed information can improve a model, why not. This work was carried exactly in such an environment. The evaluation section/appendix in the paper shows that the ML developers in the industry can appreciate the benefits of using VA to help improve their ML models. The argument ``\texttt{must never see the test data}'' would mean that the ML developers in Inetum (Spain) should not see plots such as in Fig. 5 in order to improve their models.

\vspace{2mm}\noindent
R1 and R2 might have realised that this argument about ``\texttt{must never see the test data}'' is fundamentally questionable, since it is a common practice in almost all ML workflows. In this review cycle, they used an alternative description based on their guess about the uses of ``red dots''. As mentioned earlier, this guess was not correct.

\vspace{2mm}\noindent
Q3.	\texttt{Moreover, multiple reviewers thought that the benefit of involving a human in the loop is not sufficiently clear, and that automated data augmentation might achieve similar results.}

\vspace{2mm}\noindent
A3.	As mentioned in the previous revision report, this VA work stimulated a separate project (through another student's placement at Inetum (Spain) working on machine learning solutions). A paper on this new development is currently under review. From perspective of this iGAiVA project,

\begin{itemize}
    \item[(i)] The VA work stimulated the second project on machine learning solutions. In this way, VA contributed to the scientific development. It is important for the VIS community to receive credits for such technical advancement. Hopefully, VIS reviewers will also appreciate such importance in the interest of VIS as a scientific subject and community.
    \item[(ii)] When there is no automated solution to be used (i.e., before ML solutions become deployable), VA allows humans to address the practical needs in an efficient, effective, and understandable way. So the benchmark should not be a speculation that an automated solution is feasible but a situation where this is no automated solution. Fig. 1 in the paper illustrates the comparison with such a benchmark situation.
    
    In general, many visualization and visual analytics papers presented technical solutions for tasks that potentially can be automated. It will not be appropriate to undervalue such work by speculating an automated solution. For similar reasons, the UK funding body advises their reviewers not to make critical comments based on speculation. The VIS community is already facing a huge pressure about the values of human-in-the-loop. The last thing we want is for VIS researchers themselves to undermine the values of VIS research through speculation.
    \item[(iii)] Because the authors have done both VA and ML for this industrial application, the authors have some initial findings about the relative merits. The authors are indeed conduct new projects to compare and evidence such merits. Nevertheless, discussing such relative merits is beyond the scope of this paper, especially when we do not have any space to talk about our automated solutions.
\end{itemize}

\vspace{2mm}\noindent
Q4.	\texttt{Finally, there were some concerns about missing technical details, gaps in the explanations, and several other issues that relate to the positioning with respect to prior work, and the general quality of presentation.}

\vspace{2mm}\noindent
A4.	The authors welcome and appreciate the suggestions for further clarifications. We have already revised the paper after receiving similar comments in the previous review cycle (e.g., adding texts and references). In the same way, we will certainly study each of such comment and make further improvements.

\vspace{2mm}\noindent
Q5.	\texttt{Some questions in individual reviews}

\vspace{2mm}\noindent
A5.	Some reviewers asked technical questions, which can be seen as discussion points between authors and reviewers. The authors welcome such question-based discussions, reflecting that these reviewers have the professionalism and confidence to indicate that they are not so sure when making a point. Such discussion points may sometimes result from a misunderstanding, and inadequate writing, and sometimes both. In some ML conference review processes, there is a particular phase for reviewers to ask questions anonymously. Here we give a brief answer to each question in the review.

\begin{itemize}
    \item[a)] \texttt{R1: I fully understand that it's the visualization that reveals the existence of red clusters, but once human understanding is achieved, who not automate this process?}
    
    Yes. ``Once, … and then …''! However, we found that it is not as simple as the question suggested. It is not always the case that after finding a red cluster, selecting some collected training data (i.e., some yellow dots) in the area as examples to generate synthetic training data will always lead a positive outcome. Some further results in Appendix A shows such cases. The success or failure depends on many factors, including the distribution patterns of the red, blue, yellow, and grey dots in the area where the red cluster appears as well as in the other 18 dimensions (see the figure for Q1), the semantic characteristics of the two features used to plot these dots, the language aspects of the messages, and the ability of LLM to generate a “suitable” set of synthetic messages. Currently, we found that targeting a 20D area of training data hinted by a cluster of red dots (testing data) has much higher probability to success than random trials. We also found that an experienced ML developer has good intuition in selecting the area and have better success rate than an inexperienced ML developer, who just target red clusters. In Section 5, we discuss such human pattern recognition in some details. Note that this approach is new, and the ML developers are still trying to formalize their experience and intuition into transferable knowledge. This may take many years and may need a community effort. Meanwhile, we are also using ML to learn successful patterns in different projects. Hopefully, when both are combined, there will be explainable understanding.
    \item[b)] \texttt{R1: it would be better to provide both precision and recall.}
    
    We tried this. It does not support human or machine analysis very well. Given N classes, for class [i], recall calculation is based only on the data objects in the class [i]. while precision calculation involves data objects from other N-1 classes. When class sizes vary significantly, the big classes add a lot of false positive errors into the precision of class [i]. So ML developers find that the indication of a low precision is not quite actionable. It is relatively easy to interpret recall of class [i], and decide actions on this. In a different project inspired by iGAiVA, we also tested precision as a metric for automated optimization. The result was not as good as some other measures, suggesting that ML did not find precision as useful as some other measures.
    \item[c)] \texttt{R1: The team used two ML methods, namely gradient-boosted decision trees (using CatBoost) and convolutional neural network (CNN, using TensorFlow)” Do you mean character CNN? Or using CNN to classify based on images?}
    
    The CatBoost model and CNN model mentioned in this context are traditional classification models (widely studied in the literature) for predicting text labels. These models are designed to process text features rather than images. In general, the iGAiVA approach and software can be used in conjunction with any text classification model that is a function:
    \[
        F(\text{text feature vector}) \rightarrow \text{label}
        \quad \text{or} \quad
        G(\text{text message}) \rightarrow \text{label}
    \]
    and the model is trained with pre-labelled data.
    \item[d)] \texttt{R1: Section 4.1: how did you provide the t-SNE embeddings? Do you first embed the messages in high dimensions? If so, what algorithm? (word2vec? Sbert? Tf-idf? char-CNN embedding?)}
    
    Yes, TF-IDF method, 28,927-dimensions.
    \item[e)] \texttt{R1: Section 4.2: what are these "good number of features" (related to the previous question) that PCA is applied to?}
    
    Similar to t-SNE, the 28,927-dimensional feature vectors were used as the input to PCA (sklearn). We studied the first 20 principal components using various plots similar to those in Figs. 4 and 5.
    \item[f)] \texttt{Section 5: "D. … one may notice that the one with poor testing results may have less expected ordering of keywords." Do you mean, by "ordering", the counts of words? Tag cloud is a word cloud if I understand correctly, and does not show ordering of words.}
    
    Thank you for asking this question. We actually meant to describe scenarios where one keyword was common in one set of data objects but less common in another set, or vice versa. We use the word ``order'' [of the frequency] as a short way to say this. Tag cloud depicts mainly the statistical frequency, and allows the observation of the ordering of different words in terms of usage frequency. They are not about the ordering of words appearing in messages. We will try to find a better way to explain this.
    \item[g)] \texttt{R1: Why not automate by identifying the regions of high “redness” and generate synthetic data based on points there?}
    
    In short, we are working on automating this. It is not straightforward (please see our answer (a)). In some way, VA solution is still useful and VA is also helping the development of the automated solution. It is necessary to note that VA inspired our projects to develop automated approaches. We would like the field of VIS to receive some credits before the automated work is published. Unfortunately, it is not always easy for VIS to get the deserved credits as a good number of colleagues in ML, and occasionally some colleagues in VIS as well, may not see the benefit of using human knowledge in optimising ML models. As most model optimisation is fundamentally an NP problem, mathematics tells us VIS and human knowledge can help reduce the search space (Chen, 2020, 10.1093/oso/9780190636685.003.0016).
\end{itemize}

\end{small}
\end{helveticasection}
\section{Reviews of PacificVIS 2025 Conference Track}
\label{apx:PVIS-Conf-reviews}

\emph{The following email was received on December 24, 2024.}

\begin{tiny}
\begin{verbatim}
Dear <name of the first author>

We regret to inform you that we are unable to accept your IEEE PacificVis 2025
conference paper track submission:

 9555 - iGAiVA: Integrated Generative AI and Visual Analytics in a Machine
 Learning Workflow for Text Classification

The reviews are included below.

There were many factors considered in evaluating the reviews. Although numerical
scores are important, we also read reviewers’ comments and discussions closely to
understand their reasons for the ratings. We arrived at the final decision by
balancing these points of view.

Many of the submissions that were not accepted present exciting work and ideas.
We hope you will find the reviewers' comments informative, especially if you
revise your paper for submission to the next PacificVis conference, IEEE TVCG,
or elsewhere.

We particularly encourage such revision where submissions were positively received
by reviewers, but the revisions required were deemed to be beyond the scope of the
conference review cycle. If you address the issues raised and subsequently submit
to IEEE TVCG, please make reference to the IEEE PacificVis 2025 submission and
include a description of how you have addressed the reviewers' comments. The journal
often reinvites some of the same reviewers to be part of the peer-review process.

Please also consider submitting work that was received positively to the IEEE
PacificVis 2025 Visualization Notes program or Posters program. This will enable
you to still present your work to the PacificVis audience and get further feedback
on your ideas. For more information, please see
https://pacificvis2025.github.io/pages/index.html. If your work is related to
machine learning/artificial intelligence, please also consider submitting to the
VisMeetsAI workshop (https://vismeetsai.github.io/), co-located with IEEE PacificVis.
Please contact the workshop chairs for related questions. In addition, PacificVis
2025 has a new workshop on Visualization for the Digital and Public Humanities
(DPH-Vis) (https://soumyadutta-cse.github.io/dphvis/). Please consider the submission
if your work is related to digital and public humanities. 

We thank you for submitting your paper to IEEE PacificVis 2025 conference paper track.
We wish you the best in your endeavors and hope to see you at the conference. Please
refer to the conference website for updates on the program.

Sincerely,
<names of co-chairs>
IEEE PacificVis 2025 Papers Co-Chairs

----------------------------------------------------------------

primary review (reviewer 2)
score 2/5

  Second Round Recommendation

    (blank)

  Second Round Review Text

    (blank)

  Summary of all reviews from the Primary Reviewer

    The reviewers of this resubmission thought that the feedback from the previous
                round of reviews has not been adequately accounted for. They still
                have crucial concerns about this work that cannot be addressed within
                the conference cycle.

    In particular, all reviewers expressed concerns about the issue of leakage in the
                experimental setup, which makes the improvement that is reported as
                having been achieved with the system unreliable. Moreover, multiple
                reviewers thought that the benefit of involving a human in the loop is
                not sufficiently clear, and that automated data augmentation might
                achieve similar results. Finally, there were some concerns about
                missing technical details, gaps in the explanations, and several other
                issues that relate to the positioning with respect to prior work, and
                the general quality of presentation.

  Recommendation from the Primary Reviewer

    Probably reject

  The Review

    This manuscript is a resubmission of a work that was previously submitted to the
                journal track of PacficVis. Changes to the manuscript are relatively
                minor. Instead, the senior author decided to address the majority of
                the reviewer comments by claiming that they should not have been
                reasons for rejection. I strongly disagree.

    My own primary concern in the previous round was to point out that the
                experimental setup is invalid, since it allows use of the proposed
                system to overfit to the test data, and does not evaluate whether this
                leads to any benefits on independent data. My suggestion to overcome
                this was to conduct experiments with a standard three-way data split
                into training, validation, and test data, where the user would be
                allowed to generate synthetic training data based on observing errors
                that arise on validation data that was not included in the training.
                However, the final test data cannot be made available to the user of
                the VA system, and can only be used once, to finally evaluate the
                effectiveness of the VA system.

    The effort of conducting experiments with such a valid setup is not much larger
                than performing them with the invalid setup that has instead been
                chosen. Therefore, I strongly disagree with the senior author's
                suggestion that this request "is an ambitious long-term endeavor" and
                therefore unreasonable.

    Also, such a three-way data split DOES NOT violate "the guidelines and conventions
                for evaluation" (as claimed by the senior author), but is rather a
                widely understood necessity if one wishes to draw conclusions such as
                "(i) using VA to select example data for LLMs, it can deliver a higher
                amount of improvement than random selection, (ii) using VA and LLMs,
                the model developers can achieve a noticeable amount of improvement
                that cannot easily be achieved in the existing workflow that focuses
                on parameter tuning and model architectures, (iii) the improvements
                can be made to most of the classes using synthetic data". Without such
                a split, we simply do not obtain any reliable estimate of the
                "improvement" that has been achieved with the system, which would be
                required to support such claims. This fact is explained in any
                introductory textbook to machine learning, e.g., the first few pages
                of the chapter on "Model Assessment and Selection" in the popular
                textbook by Hastie et al., "The Elements of Statistical Learning",
                from which I took for following quote:

    "Ideally, the test set should be kept in a “vault,” and be brought out only at the
                end of the data analysis. Suppose instead that we use the test-set
                repeatedly, choosing the model with smallest test-set error. Then the
                test set error of the final chosen model will underestimate the true
                test error, sometimes substantially."

    The fact alone that the evaluation presented in this manuscript violates such
                basic principles makes it unfit for publication. In addition, I still
                stand by my original assessment that for a VA paper, the specific
                benefit of involving a human in the loop via the newly proposed
                workflow, and the new prototype, should be evaluated in comparison to
                the obvious alternative of performing LLM-based data augmentation
                fully automatically.

  Overall Rating

    2 - Reject.<br>The paper is not ready for publication in PacificVis. The work may
                have some value but the paper requires major revisions or additional
                work that are beyond the scope of the conference review cycle to meet
                the quality standard. Without this I am not going to be able to return
                a score of '4 - Accept'.

  Expertise

    Knowledgeable

----------------------------------------------------------------

secondary review (reviewer 1)
score 2.5/5

  The Review

    The paper described in detail a project conducted while one of the authors was in
                a placement with an industrial partner (the details in how long the
                project is etc seems irrelevant and I would suggest removing these in
                the paper). The main thrust of the paper is to use t-SNE/PCA embedding
                of features (I did not find what these features are, see detailed
                comments later) to identify regions that have high concentrations of
                incorrectly classified points, then manually takes message in these
                region, and use ChatGPT to generate more synthetic data.

    On the positive side, the paper is very detailed and an interesting read in the
                thought process of this project, and contains good details on the
                challenges an ML oriented company faces in the real world. I also like
                the idea of trying to understand where the model failed via
                visualization. The system this project developed (Fig 6) also seems to
                be comprehensive and non-trvial.

    On the negative side, one big concern I have is that, if I understand correctly,
                the synthetic data are generated based on the Vis4Ml system, and in
                that system, the red points are incorrectly classifier points in the
                **test set**. If this understanding is correct, then essentially you
                are using the label information of the test set to get more training
                data, which constitutes a label leak. So any improvement is not
                reliable.

    The second issue I can see is that the process seems to be quite manual, where a
                human operator inspects the t-SNE/PCA plots, identifies clusters of
                red points, and uses ChatGPT to generate more synthetic data. I fully
                understand that it’s the visualization that reveals the existence of
                red clusters, but once human understanding is achieved, who not
                automate this process? (see comments later)

    Detailed comments:

    Table 1. “according to the organizational structure, incoming messages are to be
                classified into
    15 task categories” If this is a multiclass classification, it would be better to
                provide both precision and recall.

    “ The team used two ML methods, namely gradient-boosted decision trees (using
                CatBoost)
    and convolutional neural network (CNN, using TensorFlow)” Do you mean character
                CNN? Or using CNN to classify based on images?

    Section 4.1: how did you provide the t-SNE embeddings? Do you first embed the
                messages in high dimensions? If so, what algorithm? (word2vec? Sbert?
                Tf-idf? char-CNN embedding?)

    Section 4.2: what are these “good number of features” (related to the previous
                question) that PCA is applied to?

    Section 5: would be helpful to highlight the figure to specify which area the
                reader is supposed to look at. E.g.,”A. In a t-SNE scatter plot, there
                is a relatively isolated class, but its classification results are not
                satisfactory.” Furthermore, “Both scenarios suggest that there are
                distinct features to enable correct classification”, I am not
                understanding this statement. If there is a cluster of red, it means
                that samples with similar features are NOT being classified correctly.

    Section 5: “D. …  one may notice that the one with poor testing results may have
                less expected ordering of keywords.” Do you mean, by “ordering”, the
                counts of words? Tag cloud is a word cloud if I understand correctly,
                and does not show ordering of words. “can select text messages with
                possibly overlooked keywords to synthesize more training data” – who
                is to decide what’s “overlooked”, do you need a human with domain
                knowledge to decide, which seems to be not scalable?

    Section 5, under table 2 “The RBF heatmap in Fig. 5(b) helps us determine a
                separation
    line. We then select example messages from the training data on the left part and
                generate 525 synthetic messages as additional training data.” – This
                confirmed my understanding that this system is quite manual and not
                scalable. Why not automate by identifying the regions of high
                “redness” and generate synthetic data based on points there?

  Overall Rating

    2.5 - Between Reject and Possible Accept.

  Expertise

    Knowledgeable

----------------------------------------------------------------

reviewer review (reviewer 3)
score 3/5

  The Review

    The issues mentioned in summary review is partly solved but not completely. In
                particular there are two major problems.

    1. The authors make a clarification of their consideration on abbreviations. But
                the introduction part is still unsatisfying. The structure of the
                introduction can be refined. Immediately after the first paragraph,
                there should be a brief overview of the practice of generative AI-
                based data augmentation. Next, the shortcomings of this approach can
                be highlighted, followed by an introduction to the proposed VIS
                solution and its benefits, as mentioned towards the end of your second
                paragraph. Finally, it is important to clearly define the specific
                context and task, namely the computerized ticketing system, and at
                least provide a concise explanation of your system design. The
                introduction should be made more reader-friendly, ensuring that
                readers are engaged and easily able to grasp the core aspects of your
                work.

    2. A random selection for augmentation was added to compare and verify the
                effectiveness of the VA-assisted approach. However, R1’s first
                question seems to have been either misunderstood or overlooked. R1 was
                likely referring to the potential issue of data leakage, which arises
                from using test set information to help select the text messages. In
                this case, the validity of the quantitative evaluation itself is
                questionable. Also, the comparison between your proposed approach with
                random selection is unfair. A conventional interpretation would be
                that the test set is actually serving as the validation set. A
                possible correction would be to demonstrate the effectiveness of your
                work in later usage in reality, when the model encounters unseen data,
                or to just follow the three-way data split for evaluation.

    I suggest the paper be accepted after the two problems are fixed.

  Overall Rating

    3 - Possible Accept.<br>The paper is not acceptable in its current state, but
                might be made acceptable with revisions within the conference review
                cycle. If the specified revisions are addressed fully and effectively
                I may be able to return a score of '4 - Accept'.

  Expertise

    Expert

----------------------------------------------------------------

reviewer review (reviewer 4)
score 2/5

  The Review

    First and foremost, I was extremely shocked to receive such a response letter. I
                am equally disappointed in the authors' disrespectful approach to the
                peer review process. The comments provided during the previous review
                round were intended to help improve the submission. If there are areas
                of disagreement, it is sufficient to state the reasons objectively.
                Most reviewers are highly skilled and approach their reviews with a
                friendly and objective mindset. This response letter, however,
                reflects an inappropriate attitude that does not respect the
                reviewers' efforts.

    Setting aside this condescending tone, I find this round of revisions deeply
                disappointing for the following reasons:

    1. The paper disregards many review comments and presents itself in an overly
                forceful manner. PVIS papers typically adhere to a consistent
                structure, including a well-developed Discussion section, which this
                paper lacks. Placing such content in the appendix is equally
                unacceptable, as the appendix is not part of the main text.

    2. Consistency in terminology is a fundamental requirement for academic papers.
                The reviewers are certainly aware of the meanings of terms like ML,
                generative AI, and LLMs. The issue is not about understanding the
                terms but about their inconsistent usage. For instance, in the prior
                review, R2 noted that terms like ML, generative AI, and LLMs were used
                confusingly, and the reviewer simply suggested using them consistently
                and contextually rather than mixing them arbitrarily.

    3. In the previous round, a reviewer highlighted the unclear logic in the
                introduction. For example, the first paragraph discusses the link
                between visualization and ML, but the second paragraph abruptly shifts
                to generative AI. Similarly, the authors mention data collection
                limitations affecting ML model performance and list three reasons but
                fail to address how previous research has tackled these issues. It is
                unacceptable and disrespectful that the authors completely ignored
                these comments without offering any response.

    4. Section 2.1 reviews various visualization techniques for text analysis, but it
                does not clarify their connection to the main problem addressed in
                this paper. The gap between this paper’s approach and prior
                techniques, as well as the improvements it offers, remains unclear.
                Section 2.2 briefly mentions applying ML to text analysis but fails to
                focus on how ML or LLMs are specifically used for the task. Instead,
                it highlights the use of LLMs for generating synthetic training data,
                which misaligns with the prior work on text analysis. If the focus is
                on addressing missing labels in text data, the related work should
                include methods like interactive text labeling or machine learning-
                based labeling techniques. For example B. Grimmeisen et al., "Visgil:
                Machine learning-based visual guidance for interactive labeling," The
                Visual Computer, 2023. B. Miller et al., "Active learning approaches
                for labeling text," Political Analysis, 2020. J. Bernard et al.,
                "VIAL: A unified process for visual interactive labeling," The Visual
                Computer, 2018. C. Felix et al., "The exploratory labeling assistant,"
                UIST Proceedings, 2018. Section 2.3 should be central to the paper,
                emphasizing the contribution of the proposed ML workflow. It should
                connect to prior VIS work on interpretability and highlight the
                research gap. This section currently lacks that critical focus.

    5. The authors’ response to R3’s comments on requirement analysis is
                disappointing. Reviewers emphasized that mining pain points from real-
                world application domains require identifying existing practices and
                their limitations. The first requirement of using visualization
                techniques to spot errors lacks a discussion of the current practices'
                shortcomings or their frequency of use. Similarly, the second
                requirement, which involves using LLMs to create synthetic data, lacks
                justification. Other methods, such as federated learning, could
                address data shortages. These justifications should ideally be based
                on expert interviews rather than personal experiences. While some
                details are included in Appendix B, they belong in the Background,
                Motivation, and Process Overview sections, not in User Evaluation.

    6. Although the authors complain about space constraints, some content could be
                streamlined. For example, Section 4.1’s introduction of t-SNE
                techniques could be omitted. Additionally, the explanation of
                visualization techniques and system design conflates the process of
                identifying patterns in the case study with the introduction of the
                visualization techniques themselves. Section 4 still lacks logical
                coherence, an issue highlighted in the first round but ignored in this
                revision.

    7. The evaluation focuses on the benefits of GPT-based data augmentation rather
                than assessing the advantages of the VA approach as a whole. To
                evaluate whether involving a human justifies the time and effort
                invested, a more meaningful comparison would be against a fully
                automated data augmentation approach. Meanwhile, I agreed with the
                primary reviewer about the setting of evaluation, which allows the
                proposed VA system to overfit the test data without evaluating its
                benefits on independent data.

  Overall Rating

    2 - Reject.<br>The paper is not ready for publication in PacificVis. The work may
                have some value but the paper requires major revisions or additional
                work that are beyond the scope of the conference review cycle to meet
                the quality standard. Without this I am not going to be able to return
                a score of '4 - Accept'.

  Expertise

    Knowledgeable
    
\end{verbatim}
\end{tiny}
\section{Revision Report Submitted to PacificVis 2025 Conference Track}
\label{apx:RevisionReport}

\emph{Following a careful study of the reviews (Appendix \ref{apx:PVIS-TVCG-reviews}) of our submission to the PacificVis 2025 Journal Track (arXiv:2409.15848, version 1), we decided to submit a shortened version to the PacificVis 2025 Conference Track, as this was encouraged by the journal track co-chairs (Appendix \ref{apx:PVIS-TVCG-reviews}). The main body of this version and the first five appendices are the submission, while the revision report that accompanied the resubmission is included in this appendix.}

\begin{helveticasection}
\begin{small}
\vspace{2mm}\noindent
Dear co-chairs, PC members, and reviewers for \#5992,

\vspace{2mm}\noindent
Re: \#5992 - iGAiVA: Integrated Generative AI and Visual Analytics in\\
\indent a Machine Learning Workflow for Text Classification\\
\indent \emph{Yuanzhe Jin, Adrián Carrasco, and Min Chen}

\vspace{2mm}\noindent
We thank the reviewers for the submission \#5992 in the previous cycle for the TVCG track. While we appreciated their effort, comments, and suggestions, we noticed that most comments (see Q1, Q2, Q4 following the letter) are of a clarification nature, which should not be reasons for a rejection decision. Meanwhile, the technical criticism for not using data imputation (see Q5) is not correct, as using LLMs to generate synthetic data is a form of data imputation. The requirements for additional evaluation do not follow the guidelines and conventions for evaluating such a piece of VA work (see Q3).

\vspace{2mm}\noindent
We are extremely disappointed by the low scores received and the rejection decision, which we believe do not reflect the contributions of this work correctly. Because this work was conducted in an EU project that finished recently and there is a need for the work to be published timely. As required by the EU, the previous version of the paper is already available at arXiv (2409.15848). We therefore, reluctantly, resubmit this work to be considered for the conference track of PacificVis 2025. We thank the reviewers in advance for their further reviewing effort.

\vspace{2mm}\noindent
Attached below is our feedback based on the rejection reasons outlined in summary reviews. For each given reason, we have located the related comments in individual reviews before we provide our clarifications and actions.

\vspace{2mm}\noindent
Yours sincerely,\\
$\langle$one author name$\rangle$,\\
on behalf of the first and second authors $\langle$other two author names$\rangle$

\vspace{6mm}\noindent
\textbf{Summary Review:}

\vspace{2mm}\noindent
\textbf{Q1.} \texttt{The introduction is too brief and there are Conceptual confusion, inconsistencies and lacking motivations. (R1, R3, R4).}

\vspace{2mm}\noindent
\textbf{A1.} R1 and R2 queried about the terms VIS and VA. R3 and R4 pointed out the need
to introduce VIS4ML and LLM4Data. Firstly we provide our rationale about using the
terms:

\begin{itemize}
    \item \textbf{VIS} and \textbf{VA}. The IEEE VIS community decided many years ago to use VIS as an abbreviation for Visual Analytics, InfoVis, and SciVis. The merge of the three conferences a few years ago resulted in a single conference name for Visualization and Visual Analytics (abbreviated as VIS). As a collective community decision, VIS stands for both visualization (or lowercase vis) and VA. We are aware of the possible confusion, but the authors merely tried to follow the community decision. Perhaps the VIS management can communicate such decisions to the VIS community better as an individual paper has very limited mandate and capacity to do so.
    \item \textbf{VIS4ML}. This term was the term used by Sacha et al. 2018 ([53, check]). They considered the option of VA4ML, but decided to use VIS instead of VA to cover the broader scope of VIS.
    \item Our work falls into the broad scope VIS4ML as well as the narrow scope of VA4ML. We choose to avoid the term VA4ML as it was less commonly used and might cause another debate as VIS vs VA.
    \item Our work consists of the three main components of VA as defined by the VA sub-community in the 2000s and 2010, i.e., analysis (feature space analysis, the LLM-based data synthesis), visualization (various plots used), and interaction (throughout the iGAiVA system). We therefore described iGAiVA as a VA system for supporting a VA workflow.
    \item \textbf{LLMs} and \textbf{Generative AI (GAI)}. We did not anticipate that readers would need more introduction on the links between the two terms as it is widely known that GAI systems such as ChatGPT were built primarily on LLMs. Nevertheless, we appreciate that readers may not be familiar with how to use LLMs to generate synthetic data. In our paper, we have provided Section 5 specifically for informing readers of this process with examples.
\end{itemize}

\noindent
\textbf{ACTIONS:} \textrm{All VIS submissions have page limits. The previous version of this paper just fits into 9 pages of the TVCG format, and we have to remove about 0.6 pages of the text in order to fit the conference format. When asking for adding a fair amount of texts, some reviewers would suggest what texts or figures may be removed. Without such suggestions, it is often difficult for authors to act on suggestions for addition as in our case. We therefore decide to add one small abbreviation list (before Appendix A) for the terms that we used in the paper. In the cases VA, VIS, and VIS4ML, we provide further clarifications. We also added additional clarification text to link LLMs with generative AI.}

\vspace{2mm}\noindent
\textbf{Q2.} \texttt{Insufficient Comparison with Related Work and overlook how previous research has dealt with these issues, especially regarding data imputation. (R3, R4)}

\vspace{2mm}\noindent
\textbf{A2.} R3 commented that ``\texttt{the authors ... overlooked how previous research has dealt with these issues, especially regarding data imputation.}'' R4 suggested papers to be included in the literature studied. Thank you for the suggestions.

\vspace{2mm}\noindent
\textbf{ACTION:} \textrm{Following R3's comment, we have added a paragraph in Section 2 on data augmentation and data imputation. We added further references on these two highly-related topics, including three papers published in VIS venues on using VIS to support data imputation processes. The two papers suggested by R4 fall into the VIS4ML area, and we added these two references in Section 2.}

\vspace{2mm}\noindent
\textbf{Q3.} \texttt{There are problems in evaluation setup, result analysis (R1, R2, R3)}

\vspace{2mm}\noindent
\textbf{A3.} R1 requested for more tests on generalizability. R2 requested for a broader evaluation beyond the company involved. R3 commented on the lack of direct evaluation of visual representations. We consider that R1, R2, and R3's requests are beyond the standard requirements for evaluating a VA system. The design study paper by Sedlmair et al. (2012) suggested involving 4+ domain experts in the evaluation process. We fulfilled this requirement. We believe that our thoroughness in the evaluation process is well above the average standard of VA papers published in VIS/TVCG. In addition, the merits of using the VA-LLM-based workflow were also evaluated quantitatively through a large amount of testing. Some of the results were presented in the main body of the paper, e.g., Table 2, and more results were provided in Appendix A. The quantitative testing results provided very strong evidence in the evaluation process, which was enough to convince the industrial partner that the VA+LLM approach was effective and efficient.

The request for a broader evaluation beyond the company is unreasonable. Inetum Spain is not a small company, and the company has already been actively involved in the evaluation process. A good number of Inetum colleagues were in several meetings (up to 15 people) and participated in the interviews (5 people in addition to the second co-author). Only one Inetum colleague is a co-author of the paper, all other colleagues can be considered as independent. The evaluation focused on three requirements. We did not choose a survey style evaluation as it would be very unnatural, and unlikely we would have obtained a rich list of comments as in the paper (previously Section 7, now Appendix B). These were genuine efforts for the company to understand the research carried out and discussed strategies for further industrial development, while providing feedback to the academic researchers. As an industrial research activity, it is particularly difficult to take ongoing research work outside the company before its publication. We believe most academic researchers working with industry can appreciate that. The main contributions of this work is the new VA workflow and new prototype iGAiVA.

The request for evaluating visual representations is unreasonable. The evaluation in the paper focused on whether such a VA workflow could help model developers to improve their models. The baseline is the current workflow in Fig. 1(a). In the recent VIS2024 panel on evaluation, all panelists, including a VIS2024 co-OPC, and Michael Sedlmair himself, agreed that the demands for evaluation should not go beyond the main contributions of the paper. Although this paper used less-commonly
seen visualization plots, i.e., RBF-based heatmap and hierarchical treemaps, we did not state them as main contributions. Their relative merits and demerits in comparison with scatter plots and single treemaps are expected to be subtle, and are more suitable to be studied in separate research activities, such as an empirical study. As the VIS2024 panel pointed out, it would be unreasonable to demand such a study in a VA or application paper focusing on VA workflows.

R1’s request is an ambitious long-term endeavor for confirming the general
applicability of a VA solution. What this research can confirm is that (i) using VA to select example data for LLMs, it can deliver a higher amount of improvement than random selection, (ii) using VA and LLMs, the model developers can achieve a noticeable amount of improvement that cannot easily be achieved in the existing workflow that focuses on parameter tuning and model architectures, (iii) the improvements can be made to most of the classes using synthetic data, and (iv) with iGAiVA, the time required to achieve improvement is less than without such a tool. On the one hand, (iii) indicates the generalizability of the proposed workflow to different classes in a dataset. Our more recent study for a different paper (inspired by this paper, but not a VIS paper, under review) also confirmed the general applicability to other datasets. Meanwhile, logically, it is also easy to conclude that the generalizability is limited by how many times the iGAiVA workflow is used to improve a model, as most improvement curves will reach a plateau over a period. Like all ML techniques, the improvement cannot be generally assumed if a technique X is
deployed.

\vspace{2mm}\noindent
\textbf{ACTION.} \textrm{We will improve our conclusions to include the discussions related to R1's comments.}

\vspace{2mm}\noindent
\textbf{Q4.} \texttt{The paper lacks a discussion section, which should address aspects like scalability, potential challenges with LLMs, lessons learned, and limitations.}

\vspace{2mm}\noindent
\textbf{A4.} The VIS submission has page limits. It is not always easy to add a discussion section as the reviewer suggested. If each topic takes a paragraph, it would need about a page for the topics to be addressed.

\vspace{2mm}\noindent
\textbf{ACTION.} \textrm{As reviewers did not suggest which part of the paper is to be removed or shortened, we therefore added an appendix to meet the reviewer’s request.}

\vspace{2mm}\noindent
\textbf{Q5.} \texttt{Requirement analysis isn't very convincing and visual analytics suggested aren't particularly innovative. (R3)}

\vspace{2mm}\noindent
\textbf{A5.} We strongly disagree with this comment by R3. R3 seems to consider that the requirement analysis process in Section 3 is not correct or not true (i.e., not convincing) without any evidence or explanation. If R3 wishes to see more evidence, we could provide evidence such as PPT slides for various meetings, etc. To suggest that what was written in Section 3 is not true is a serious accusation.

R3 suggested that \texttt{data imputation} [as an alternative] could be a solution for addressing limited data in this application. Data augmentation and data imputation are both defined as using statistics of existing data to generate new data. The term ``data imputation'' is mainly used in the context of filling in missing data, while the term ``data augmentation'' is mainly used in the context of providing additional
training data in ML. Mathematically, these two terms more or less mean the same thing. We did not use either term in the original version of the paper as we thought data synthesis and synthetic data is easier to understand. If we were to use one term, ``data augmentation'' is the better option.

Using LLMs to generate synthetic data is a data augmentation method. It can
also be interpreted as a data imputation method if the meaning of ``missing data'' is re-interpreted. That is, in addition to ``missing data as part of an existing table, time series, image, video, etc.'' one may consider additional training data needed is missing data. R3 might consider using statistics to generate new data is different from using LLMs. It is necessary to note that fundamentally, LLMs are using statistics
to generate new data. Similarly, all ML models are using statistics to generate predictions or decisions. The perception that they are different is because LLMs and ML models are perceived as ``black boxes'', while traditional statistical methods use ``white box'' formulae. Actually, LLMs and ML models are also using explicit statistical formulae, except they are applied to a huge number of calculation units (e.g., neurons and synapses) in many iterations. So they are perceived as ``black boxes''
because we cannot physically follow that many calculation units and iterations.

R3 might consider that the traditional ``white box'' approach would be better. This implies [texts] -> [features] -> [statistics] -> [imputation] -> [synthetic features] -> [synthetic text]. Mathematically, text messages entail a gigantic information space (e.g., just considering how many POSSIBLE text messages that have 50 or less words). Whatever data that a ticketing system has collected, it is only a tiny portion of this information space. Hence the local statistics computed from a dataset of a
ticketing system will unlikely be of a good, reliable quality. Meanwhile, locally defined features and locally-implemented text reconstruction processes will also be challenging. Hence using LLMs provides a better solution as there will be better statistics, more features, and more advanced text reconstructions. In other words, not only using LLMs to generate synthetic data is a form of data imputation, and it is
a cost-effective solution, especially from an industrial perspective.

In a related ML-focused project (Oxford and Inetum, inspired by the VA work
in this paper), tests were carried out to compare traditional data augmentation and LLM-based data augmentation. The latter was much more effective. A paper for that project is currently under review.

R3 also commented that ``\texttt{the paper mentioned that visualization helps model developers tackle training data problems and reduce the search space. it doesn't specify which aspects of text data it addresses.}'' We are not sure if this comment refers to the Related Work section only or about the whole paper. If it were the former, this would be a small clarification request. If it were about the latter, surely Sections 3, 4, 5, and 6 provided sufficient details about ``which aspects'' of text data.
In particular, the application of ``ticketing systems'' should provide an adequate context about business texts, relatively short messages, usually reporting problems and making requests, and messages to be processed in a categorized way.

R3 also commented that ``\texttt{the visual analytics techniques suggested aren’t particularly innovative, mainly relying on established methods like t-SNE and PCA.}'' We also strongly disagree with R3's dismissive comments. Firstly the main contributions of the paper are a VA workflow, involving a relatively new analytical component (i.e., LLMs, in addition to t-SNE and PCA), new visual components (i.e., RBF-based heatmap, hierarchical treemaps, in addition to scatter plots), and a new VA software design (i.e., the four-view design). The integrated uses of analytical
algorithms and visualization is the core of this paper, including:

\begin{enumerate}
    \item t-SNE dimensionality reduction and scatter plot (t-SNE was commonly mistaken as a type of plot),
    \item PCA and scatter plot,
    \item RBF (for predicting missing data) and heatmap,
    \item word statistics and tag-treemap, and
    \item LLMs for generating synthetic data that is visualized using the above VA techniques (2), (3) and (4).
\end{enumerate}

Meanwhile, the iGAiVA tool features a variety of interaction in all four views. It was designed to improve an industrial application in a significant way, which should be encouraged rather than discouraged. The level of novelty is consistent with the majority of VA papers published in IEEE VIS Conferences and TVCG.
\end{small}
\end{helveticasection}

\section{Reviews of PacificVIS 2025 Journal Track}
\label{apx:PVIS-TVCG-reviews}

\emph{The following email was received on November 7, 2024.}

\begin{tiny}
\begin{verbatim}
Dear <name of the first author>

We regret to inform you that we are unable to accept your paper
 5992 - iGAiVA: Integrated Generative AI and Visual Analytics in a Machine Learning
 Workflow for Text Classification
for publication to the IEEE PacificVis 2025 journal track. We hope that you will find
the reviews, copied below, useful in revising your paper for a future venue. One such
potential venue is coming up very soon; the deadline for the IEEE PacificVis 2025
conference paper track is November 13 (abstracts) and November 20 (full papers). See
below for details.
 
ACCEPTANCE RATE
This year, the IEEE PacificVis 2025 journal track received 123 submissions. We
conditionally accepted 15 of them, amounting to an acceptance rate of 12 percent.
To boost the acceptance rate while upholding the high standards of TVCG, we invited an
additional 15 papers as fast-track (major revisions with reviewer continuity) to IEEE
TVCG. This yields a tentative final acceptance rate of 24 percent.
 
REVIEW PROCESS
All papers were reviewed by at least two PC members (the primary and secondary) and two
external reviewers. Please find the "Summary Review" among the review texts, where the
primary reviewer has identified the most important issues raised by the reviewers.
There were several factors considered in evaluating the reviews, recommendations,
scores, and subsequent decision making. Although the numerical scores are important,
the paper chairs carefully read the reviews (shared with you below) to understand the
reviewers' reasons for the scores and also the online discussion (not shared with
authors). In a number of cases, the reason for not accepting a submission is that the
submission would require a major revision to be ready for publication. Unfortunately,
such major revisions are not possible within the time frame of the PacificVis review
process.

CONFERENCE PAPER TRACK
The IEEE PacificVis 2025 conference paper track deadline was chosen to give authors
sufficient time to quickly revise and resubmit papers rejected from the journal track.
The deadline is as follows:
* Abstracts - November 13, 2024
* Full papers - November 20, 2024
 
Conference track papers are submitted via PCS (same place where you submitted your
journal paper): https://new.precisionconference.com/vgtc
 
When submitting a revised paper to the IEEE PacificVis 2025 conference track, you are
given the option to maintain reviewer continuity and provide the submission ID from
this journal track. If you choose to maintain continuity, the conference paper track
chairs and PC members will do their best to invite the same reviewers who reviewed your
paper for the journal track. You will also be asked to provide a review response letter
with your paper.
 
We thank you for submitting your paper to the IEEE PacificVis 2025 journal track. We
wish you the best in your endeavors and hope to see you at the conference. Please see
the conference website for updates on this year's program.
 
Sincerely,
<names of co-chairs>
IEEE PacificVis TVCG Journal Track Paper Chairs

----------------------------------------------------------------

primary review (reviewer 4)
score 2.5/5

  Recommendation (Second Round)

    (blank)

  Final Review (Second Round)

    (blank)

  Recommendation (First Round)

    Probably reject

  Summary Review (First Round)

    All reviewers see some potential of the paper, but there are more issues in each
                part of the paper to be solved.


    + Pros
    1. Close collaboration between the author and industry, and the success of the
                system has been acknowledged by the partnering company. (R4)
    2. The paper details the implementation process and highlights how users derive
                hypotheses and actions from the visualizations. (R2, R3, R4)
    3. A timely and relevant topic (R1, R3)
    4. The potential of leveraging LLMs to synthesize training data is explored. (R2)


    - Cons
    1. The introduction is too brief and there are Conceptual confusion,
                inconsistencies and lacking motivations. (R1, R3, R4)
    2. Insufficient Comparison with Related Work and overlook how previous research
                has dealt with these issues, especially regarding data imputation.
                (R3, R4)
    3. There are problems in evaluation setup, result analysis (R1, R2, R3)
    4. The paper lacks a discussion section, which should address aspects like
                scalability, potential challenges with LLMs, lessons learned, and
                limitations. (R2, R4)
    5. Requirement analysis isn't very convincing and visual analytics suggested
                aren't particularly innovative. (R3)

  Paper Review

    This paper presents iGAiVA, a tool that combines visual analytics (VA) and
                generative AI to address data deficiencies in text classification
                model development.

    Strengths:

    1. (The application merits) This paper’s idea and implementation stem from close
                collaboration between the author and industry, and the success of the
                system has been acknowledged by the partnering company.
    2. (The system merits) The paper clearly details the implementation process and
                highlights how users derive hypotheses and actions from the
                visualizations.

    Weaknesses and Questions:

    1. The introduction is too brief. 
       Without sufficient background information, readers unfamiliar with vis4ml or
                LLM4data synthesis may struggle to understand the paper's goals.
                Additionally, the absence of related work or citations undermines the
                discussion of challenges and prior work. Core contributions, such as
                system design, are also insufficiently explained.

    2. Insufficient Comparison with Related Work:
       i. Given the approach used in this paper, such as leveraging LLMs in visual
                text analysis and addressing data deficiency in visual analytics, some
                relevant papers are missing. For example, Section 2.1 (VIS in Text
                Analysis) should discuss recent work using LLMs for text analysis
                (e.g., "Automatic Histograms," CHI 24 EA). In Section 2.2 or 2.3,
                vis4ml papers addressing training quality, like SliceTeller (VIS 22)
                and OoDAnalyzer (TVCG 2020), should be discussed.
       ii. Some references are misplaced, such as explorer being categorized under
                VIS4ML in Text Analysis rather than VIS4ML in General. Additionally,
                many references seem tangential to the paper’s focus and are listed
                without clear organization or meaningful discussion.
       iii. The lack of directly related references, combined with insufficient
                emphasis on how this work builds upon or differs from prior research,
                makes it hard to gauge its contribution. In particular, Section 2.1
                ends without comparing this research to the broader literature, and
                Sections 2.2 and 2.3 merely state, "we use LLMs/VIS to...," which
                doesn’t clarify how this paper advances the field. It’s important to
                explicitly state how prior works influenced this paper and what
                specific improvements were made.


    3. Lack of Generalizability Discussion:
       The generalizability of this work is unclear and needs to be addressed. While
                the paper is based on the internship company’s processes and
                experiences, which helps validate system usability, it limits broader
                applicability. For example, it's unclear whether the requirements in
                Section 3 apply to other text datasets, classification scenarios, or
                organizations, and whether the visual views, hypotheses, and actions
                in Section 4 are similarly applicable. The paper should include a
                discussion on generalizability or clarify that the tool is primarily
                intended for this internship setting.

    4. Writing issues
       The paper lacks an overview of the research questions, methodology, and other
                key elements. For instance, before Section 4.1, it would be helpful to
                add a paragraph summarizing the four visualizations and how they
                relate to each other, offering readers insight into the high-level
                design choices. Similarly, the design evaluation and requirements
                analysis in Section 7 overlap with Section 3, and Section 6 does not
                provide clear usage cases, particularly regarding multiple rounds of
                model tuning, which would be the best way to demonstrate system
                effectiveness. This should be illustrated with both text and images so
                readers don’t have to consult the supplementary materials.

  Overall Rating

    2.5 - Between Reject and Possible Accept.

  Expertise

    Expert

----------------------------------------------------------------

secondary review (reviewer 1)
score 2/5

  Paper Review

    This manuscript describes a visual analytics system that aims to improve text
                classification by augmenting the available training data with text
                that has been synthesized by a large language model. Even though the
                topic is timely, the paper falls short of my expectations for
                publication at a top-tier venue for multiple reasons:

     1. The evaluation setup is not valid. Proper evaluation would require a three-way
                data split into training, validation, and test data. The user of the
                visual analytics (VA) system can only be allowed access to the
                training and validation data, and must never see the test data on
                which the final benefit of the VA system is evaluated. In the proposed
                approach, incorrectly classified test data are highlighted, and users
                are encouraged to play with the system to reduce those errors. It is
                unsurprising that this can be achieved, but the tough question is
                whether this overfits to the specific test examples that the user has
                seen in the interface, or if it also generalizes to new (unseen)
                examples. In the worst case, the interventions via the system might
                reduce accuracy on future cases due to overfitting. There is no proper
                evaluation to check for this.

     2. Even though the main claimed benefit is the VA system, and the paper has been
                submitted to a VA venue, experiments are designed to evaluate the
                benefit of GPT-based data augmentation, as opposed to the benefit of
                taking a VA approach as such: The only comparison that is shown is
                between the original classifier (without data augmentation) and a
                classifier that has been trained with the data augmentation via the
                proposed interface. To judge whether the time and effort that has been
                invested by involving a human in fact pays off, a more meaningful
                comparison would be towards a fully automated data augmentation.
                Authors describe such an approach ("Without VIS techniques...") in
                Section 5, but do not report corresponding results. Therefore, it
                remains unclear whether the VA approach as such leads to a specific
                benefit.

     3. Even though it is plausible that a "classical" ML approach such as gradient-
                boosted decision trees can benefit from data augmentation via a GPT
                model, the paper does not justify why that approach should be
                preferred over using the GPT model for the classification task
                directly, either by using it to compute a text embedding, or by fine-
                tuning it. I would expect that such an approach might produce even
                stronger results than those that are reported here, without requiring
                human feedback.

  Overall Rating

    2 - Reject.<br>The paper is not ready for publication in PacificVis. The work may
                have some value but the paper requires major revisions or additional
                work that are beyond the scope of the conference review cycle to meet
                the quality standard. Without this I am not going to be able to return
                a score of '4 - Accept'.

  Expertise

    Knowledgeable

----------------------------------------------------------------

reviewer review (reviewer 2)
score 2/5

  Paper Review

    This paper presents an exploration of integrating visual analytics (VA) and large
                language models (LLMs) within a machine learning (ML) workflow for
                text classification. It focuses on addressing data distribution
                challenges by using VA to guide LLMs in generating synthetic data,
                which is aimed at improving ML model performance. And proposing a
                VIS4ML approach and developing a software tool, iGAiVA, which
                integrates VA and LLMs into the ML workflow.

    Strengths
    1. The methods and algorithms are clearly described in detail.
    2. The system’s functionality is well-developed and comprehensive.
    3. The potential of leveraging LLMs to synthesize training data is explored.

    Weaknesses
    1. Conceptual Confusion:
      • The distinction between visualization and visual analytics is blurred. The
                paper uses the term "VIS" to refer to both, which does not clearly
                define the specific research scope.
      • Terms like ML, generative AI, and LLMs are frequently used in a confusing
                manner. It's important to clarify which concept applies in each
                context.
    2. Inconsistencies:
      • The title and abstract focus only on visual analytics, while the rest of the
                paper includes broader topics, leading to inconsistency.
      • The paper begins discussing text analysis but later introduces a ticketing
                system without clearly explaining the link between the task and
                application.
    3. The motivation is vague. It does not clarify the current challenges in ML, the
                limitations of existing methods, or why visualization or visual
                analytics are necessary to address these problems in text analysis,
                especially with regard to LLM integration.
    4. The paper lacks a formative study. The requirements are drawn solely from the
                author’s intern, which does not represent a reliable user group.
                Involving domain experts and conducting a formative study with a
                broader participant base would ensure more valid and reliable
                requirements.
    5. Section 4 lacks logical coherence. The application of methods is presented
                without adequate reasoning. The rationale for using certain methods
                and not others is unclear. Some patterns are identified, but not
                discussed further. It is recommended to improve the logical flow and
                systematically summarize the problems identified.
    6. The user study should exclude stakeholders and recruit external participants. A
                detailed procedure and tasks focusing on system and workflow analysis
                should be developed for more credible results.
    7. The paper lacks a discussion section, which should address aspects like
                scalability, potential challenges with LLMs, lessons learned, and
                limitations.

    Minor issues:
      • Terminology: ML4ML workflow should be VIS4ML.
      • Several grammatical and spelling errors need correction.

    In summary, while this paper introduces interesting technical approaches and
                system functionalities, it does not meet the standards required for
                publication in PacificVis. It lacks the conceptual clarity,
                consistency, and rigorous empirical validation necessary for a
                research paper. It might serve as a useful technical report but is not
                sufficiently mature for academic publication in its current form.

  Overall Rating

    2 - Reject.<br>The paper is not ready for publication in PacificVis. The work may
                have some value but the paper requires major revisions or additional
                work that are beyond the scope of the conference review cycle to meet
                the quality standard. Without this I am not going to be able to return
                a score of '4 - Accept'.

  Expertise

    Knowledgeable

----------------------------------------------------------------

reviewer review (reviewer 3)
score 2/5

  Paper Review

    The paper discusses a visual analytics system designed to improve text
                classification by enhancing training data with text generated by large
                language models. While the topic is relevant, it doesn’t quite meet my
                expectations for a top-tier publication for a few reasons:

    1. The abstract throws out a bunch of concepts but doesn’t clearly connect
                them—like how large language models relate to generative AI, what the
                four ML task groups are, and the different types of data deficiencies.
                Also, "vis" should mean visualization, and "visual analytics" should
                be shortened to VA.

    2. The introduction is a bit confusing; the first paragraph talks about the link
                between vis and ML, but the second one suddenly jumps to generative AI
                without connecting it back to vis or explaining its importance for
                text analysis.

    3. The authors point out that data collection limits ML model performance and list
                three reasons but overlook how previous research has dealt with these
                issues, especially regarding data imputation.

    4. It’s mentioned that visualization helps model developers tackle training data
                problems and reduce the search space, but it doesn’t specify which
                aspects of text data it addresses. Section 2.1 reviews various
                visualization techniques for text analysis, but it’s unclear how they
                relate to the paper’s main problem.

    5. The requirement analysis isn’t very convincing; for instance, the first
                requirement talks about using more visualization techniques to spot
                potential errors but doesn’t discuss the current practices'
                limitations or how often they use visualization. The second
                requirement about using large language models to create synthetic data
                lacks a solid justification—other methods like federated learning
                could also solve data shortages. The authors need to provide more
                details on where these requirements come from, ideally based on expert
                interviews rather than just their own experiences.

    6. In Section 4, the visual analytics techniques suggested aren’t particularly
                innovative, mainly relying on established methods like t-SNE and PCA.

    7. Finally, the evaluation section doesn’t really look at the role of
                visualization itself, focusing only on the benefits of data
                augmentation with large models, which leaves it unclear if the system
                brings any extra advantages.

  Overall Rating

    2 - Reject.<br>The paper is not ready for publication in PacificVis. The work may
                have some value but the paper requires major revisions or additional
                work that are beyond the scope of the conference review cycle to meet
                the quality standard. Without this I am not going to be able to return
                a score of '4 - Accept'.

  Expertise

    Expert

----------------------------------------------------------------
\end{verbatim}
\end{tiny}

\end{document}